\documentclass{article}
\PassOptionsToPackage{numbers, compress}{natbib}

\newif\ifpreprint\preprintfalse
\newif\ifnotpreprint

\ifpreprint
    \notpreprintfalse
    \usepackage[preprint]{neurips_2024}
\else
    \notpreprinttrue
    \usepackage[final]{neurips_2024}
\fi
\newcommand{\ghfootnote}{Our code is available at \url{https://github.com/mllam/neural-lam/tree/prob_model_global} (global forecasting) and \url{https://github.com/mllam/neural-lam/tree/prob_model_lam} (\gls{LAM}).}

\usepackage[utf8]{inputenc} %
\usepackage[T1]{fontenc}    %
\usepackage[hypertexnames=false]{hyperref}       %
\usepackage{url}            %
\usepackage{booktabs}       %
\usepackage{amsfonts}       %
\usepackage{nicefrac}       %
\usepackage{microtype}      %
\usepackage{xcolor}         %

\usepackage{paralist} %
\usepackage{xspace}
\usepackage{graphicx}
\usepackage{mathtools}
\usepackage{subcaption}
\usepackage{algorithm}
\usepackage{algpseudocode}
\usepackage{wrapfig}
\usepackage{tikz}
\usepackage{cleveref}
\usepackage{multirow}
\usepackage{placeins}
\usepackage{titletoc}

\usepackage{multibib}
\newcites{app}{Appendix References}
\newcites{both}{In both reference lists} %

\usepackage{siunitx}
\definecolor{LiUblue}{RGB}{0,185,231}
\definecolor{LiUorange}{RGB}{255,100,66}
\definecolor{LiUgreen}{RGB}{0,207,181}
\definecolor{NodePurple}{RGB}{154,152,217}

\let\originalleft\left
\let\originalright\right
\renewcommand{\left}{\mathopen{}\mathclose\bgroup\originalleft}
\renewcommand{\right}{\aftergroup\egroup\originalright}

\usepackage{amssymb} %
\usepackage{amsmath}
\usepackage{bm} %

\newcommand{\eq}[2][my_equation]{\begin{equation}\label{eq:#1}#2\end{equation}}
\newcommand{\al}[2][my_equation]{\begin{align}\label{eq:#1}#2\end{align}}
\newcommand{\als}[2][my_equation]{\begin{align}\label{eq:#1}\begin{split}#2\end{split}\end{align}}

\newcommand{\qm}[1]{``#1''}

\newcommand{\loss}{\mathcal{L}}

\newcommand{\set}[1]{\left\{ #1 \right\}}
\newcommand{\setsize}[1]{\left| #1 \right|}
\newcommand{\R}{\mathbb{R}} %

\renewcommand{\b}[1]{\bm{#1}} %

\newcommand{\E}[2]{\mathbb{E}_{#1} \left[ #2 \right]} %
\newcommand{\normal}[2]{\mathcal{N}\left(#1, #2\right)} %
\newcommand{\normalpdf}[3]{\mathcal{N}\left(#1 \middle| #2, #3\right)} %

\newcommand{\divergence}[3]{D_{#1}\left(#2 \middle\| #3 \right)}
\newcommand{\kl}[2]{\divergence{\text{KL}}{#1}{#2}}

\newcommand{\alse}[2][my_eq]{\begin{subequations}\label{eq:#1}\begin{align}#2\end{align}\end{subequations}}

\newcommand{\varsubfig}[4]{%
\begin{subfigure}[b]{0.33\textwidth}%
    \centering
    \includegraphics[width=\textwidth]{graphics/\detokenize{#1}/#2/#2_\detokenize{#3}.pdf}%
    \caption{\wvar{\detokenize{#4}}}
\end{subfigure}}

\newcommand{\legendsubfig}[3]{%
\begin{subfigure}[b]{#3\textwidth}%
        \centering
        \includegraphics[width=\textwidth]{graphics/\detokenize{#1}/#2/#2_legend.pdf}%
\end{subfigure}}

\newcommand{\mainsubfig}[2]{%
\begin{subfigure}[b]{0.33\textwidth}%
        \includegraphics[width=\textwidth]{graphics/#1}
        \caption{\gls{#2}}
\end{subfigure}%
}

\newcommand{\glsfirstbold}[1]{%
\acrlong{#1}~(\textbf{\acrshort{#1}})%
\glsunset{#1}%
}

\newcommand{\varline}[4]{\parbox{0.35\textwidth}{#4} & \wvar{\detokenize{#1}} & #3 & \parbox{0.35\textwidth}{#2}\\}
\newcommand{\mepsbottomlevel}{Lvl65\textsuperscript{\textdagger}~(\texttt{65})}
\newcommand{\preslevel}[1]{#1~\si{\hecto\pascal}}
\newcommand{\tablesubsec}[1]{\multicolumn{4}{@{}l}{\textbf{#1}}\\ \midrule}

\newcommand{\wvar}[1]{\texttt{#1}}%

\newcommand{\wstate}{X}
\newcommand{\forcing}{F}
\newcommand{\latent}{Z}
\newcommand{\boundarystate}{B}
\newcommand{\pred}{\hat{\wstate}}
\newcommand{\predprime}{\check{\wstate}}
\newcommand{\ensmean}{\bar{\wstate}}

\newcommand{\initstates}{\wstate^{-1:0}}
\newcommand{\forecast}{\wstate^{1:\forecastlength}}
\newcommand{\allforcing}{\forcing^{1:\forecastlength}}

\newcommand{\modeldist}{p\left(\forecast | \initstates, \allforcing\right)}
\newcommand{\arcond}{\wstate^{\timei-2:\timei-1}, \forcing^{\timei}}
\newcommand{\ardist}{p\left(\wstate^{\timei} \middle| \arcond \right)}
\newcommand{\priordist}{p\left(\latent^\timei \middle| \arcond \right)}
\newcommand{\preddist}{p\left(\wstate^\timei | \latent^\timei, \arcond \right)}
\newcommand{\vardist}{q\left(\latent^\timei \middle| \wstate^{\timei-2:\timei-1}, \wstate^{\timei}, \forcing^{\timei} \right)}
\newcommand{\vardistname}{q}
\newcommand{\postdist}{p\left(\latent^\timei \middle| \wstate^{\timei-2:\timei-1}, \wstate^{\timei},\forcing^{\timei} \right)}

\newcommand{\firstrv}{x}
\newcommand{\secondrv}{x'}
\newcommand{\observation}{y}

\newcommand{\arfunc}{f}
\newcommand{\predictorfunc}{g}
\newcommand{\predictorfunctilde}{\tilde{g}}

\newcommand{\timei}{t}
\newcommand{\ensi}{k}
\newcommand{\samplei}{m}
\newcommand{\nodei}{\alpha}
\newcommand{\nodeii}{\beta}
\newcommand{\vari}{j}
\newcommand{\leveli}{l}

\newcommand{\samplesize}{S}
\newcommand{\ngridpoints}{N}
\newcommand{\nboundarynodes}{N_b}
\newcommand{\enssize}{K}
\newcommand{\statedim}{d_x}
\newcommand{\latentdim}{d_z}
\newcommand{\forecastlength}{T}
\newcommand{\himaxl}{L}

\newcommand{\gc}{GraphCast\xspace}
\newcommand{\gcour}{GraphCast*\xspace}
\newcommand{\gcswa}{GraphCast*+SWAG\xspace}

\newcommand{\himodeldet}{Graph-FM\xspace}
\newcommand{\himodelprob}{Graph-EFM\xspace}
\newcommand{\msmodelprob}{Graph-EFM~(ms)\xspace}

\newcommand{\lamsection}{Limited Area Modeling with MEPS Data}
\newcommand{\globalsection}{Global Forecasting with ERA5}
\newcommand{\lamsectionlower}{Limited area modeling with MEPS data}
\newcommand{\globalsectionlower}{Global forecasting with ERA5}

\newcommand{\graph}{\mathcal{G}}
\newcommand{\higraph}[1]{\graph_{#1}}
\newcommand{\hiintergraph}[2]{\graph_{#1,#2}}
\newcommand{\hiintragraph}[1]{\graph_{#1}}
\newcommand{\msgraph}{\graph_\multiscale}
\newcommand{\nodeset}{\mathcal{V}}
\newcommand{\edgeset}{\mathcal{E}}
\newcommand{\gridnodes}{\nodeset_\grid}
\newcommand{\meshnodes}{\nodeset_\mesh}
\newcommand{\meshedges}{\edgeset_\mesh}
\newcommand{\himeshnodes}[1]{\nodeset_{#1}}
\newcommand{\himeshedges}[1]{\edgeset_{#1}}
\newcommand{\hiinteredges}[2]{\edgeset_{#1,#2}}
\newcommand{\hiintraedges}[1]{\himeshedges{#1}} %

\newcommand{\neigh}[1]{\text{Ne}(#1)}
\newcommand{\multiscale}{\text{MS}}

\newcommand{\noderep}{H}
\newcommand{\noderepprime}{\tilde{H}}
\newcommand{\edgerep}{E}
\newcommand{\grid}{G}
\newcommand{\mesh}{M}
\newcommand{\sender}{S}
\newcommand{\receiver}{R}
\newcommand{\gtm}{\text{G2M}}
\newcommand{\mtm}{\text{M2M}}
\newcommand{\mtg}{\text{M2G}}
\newcommand{\meanrep}{Y}
\newcommand{\stdrep}{U}

\DeclareMathOperator*{\gnn}{GNN}
\DeclareMathOperator*{\mlp}{MLP}
\DeclareMathOperator*{\softplus}{Softplus}

\newcommand{\edgevec}[3]{\b{e}_{#2 \rightarrow #3}^{#1}}

\newcommand{\edgevecprime}[3]{\b{\tilde{e}}_{#2 \rightarrow #3}^{#1}}

\newcommand{\stateweight}{\omega}
\newcommand{\invdiffweight}{\lambda}
\newcommand{\areaweight}{w}
\newcommand{\klweight}{\lambda_{\text{KL}}}
\newcommand{\crpsweight}{\lambda_{\text{CRPS}}}
\newcommand{\elbo}{\tilde{\loss}_\text{Var}} %
\newcommand{\elboloss}{\loss_\text{Var}} %
\newcommand{\crpsloss}{\loss_{\text{CRPS}}}
\newcommand{\wmseloss}{\loss_{\text{WMSE}}}
\newcommand{\nllloss}{\loss_{\text{NLL}}}

\newcommand{\rmse}{\text{RMSE}}
\newcommand{\spread}{\text{Spread}}
\newcommand{\spskrtext}{Spread-Skill-Ratio\xspace}
\newcommand{\spskrabr}{SpSkR\xspace}
\newcommand{\spskr}{\text{\spskrabr}}
\newcommand{\crps}{\text{CRPS}}

\newcommand{\meanfunc}[1]{\mu_{#1}}
\newcommand{\predsigma}{\sigma}
\newcommand{\stdfunc}[1]{\predsigma_{#1}}

\newcommand{\lat}{\text{Latitude}}
\newcommand{\lon}{\text{Longitude}}

\usepackage[acronyms, shortcuts]{glossaries}
\glsdisablehyper
\newacronym{RNN}{RNN}{Recurrent Neural Network}
\newacronym{CNN}{CNN}{Convolutional Neural Network}
\newacronym{GNN}{GNN}{Graph Neural Network}
\newacronym{MLP}{MLP}{Multi-Layer Perceptron}
\newacronym{MPNN}{MPNN}{Message Passing Neural Network}
\newacronym{GRU}{GRU}{Gated Recurrent Unit}
\newacronym{ELBO}{ELBO}{Evidence Lower Bound}
\newacronym{VAE}{VAE}{Variational AutoEncoder}
\newacronym{CVAE}{CVAE}{Conditional Variational Auto-Encoder}
\newacronym{SWA}{SWA}{Stochastic Weight Averaging}
\newacronym{SWAG}{SWAG}{Stochastic Weight Averaging Gaussian}
\newacronym{SGD}{SGD}{Stochastic Gradient Descent}

\newacronym{RMSE}{RMSE}{Root Mean Squared Error}
\newacronym{MSE}{MSE}{Mean Squared Error}
\newacronym{MAE}{MAE}{Mean Absolute Error}
\newacronym{NLL}{NLL}{Negative Log-Likelihood}
\newacronym{CRPS}{CRPS}{Continuous Ranked Probability Score}
\newacronym{SPSKR}{\spskrabr{}}{\spskrtext{}}

\newacronym{ODE}{ODE}{Ordinary Differential Equation}
\newacronym{PDE}{PDE}{Partial Differential Equation}
\newacronym{GMRF}{GMRF}{Gaussian Markov Random Field}
\newacronym{DGMRF}{DGMRF}{Deep GMRF}

\newacronym{NWP}{NWP}{Numerical Weather Prediction}
\newacronym{NeurWP}{MLWP}{Machine-Learning-based Weather Prediction}
\newacronym{LAM}{LAM}{Limited Area Model}
\newacronym{MetCoOp}{MetCoOp}{Meterological Cooperation on Operational NWP}
\newacronym{SMHI}{SMHI}{Swedish Meteorological and Hydrological Institute}
\newacronym{ECMWF}{ECMWF}{European Centre for Medium-Range Weather Forecasts}
\newacronym{MEPS}{MEPS}{MetCoOp Ensemble Prediction System}

\title{Probabilistic Weather Forecasting with Hierarchical Graph Neural Networks}

\author{%
Joel Oskarsson\\
Link\"{o}ping University\\
\texttt{joel.oskarsson@liu.se}
\And
Tomas Landelius\\
Swedish Meteorological and\\
Hydrological Institute\\
\texttt{tomas.landelius@smhi.se}
\AND
Marc Peter Deisenroth\\
University College London\\
\texttt{m.deisenroth@ucl.ac.uk}
\And
Fredrik Lindsten\\
Link\"{o}ping University\\
\texttt{fredrik.lindsten@liu.se}
}
\begin{document}
\maketitle

\begin{abstract}
In recent years, machine learning has established itself as a powerful tool for high-resolution weather forecasting.
While most current machine learning models focus on deterministic forecasts, accurately capturing the uncertainty in the chaotic weather system calls for probabilistic modeling.
We propose a probabilistic weather forecasting model called \himodelprob, combining a flexible latent-variable formulation with the successful graph-based forecasting framework.
The use of a hierarchical graph construction allows for efficient sampling of spatially coherent forecasts.
Requiring only a single forward pass per time step, \himodelprob allows for fast generation of arbitrarily large ensembles.
We experiment with the model on both global and limited area forecasting.
Ensemble forecasts from \himodelprob achieve equivalent or lower errors than comparable deterministic models, with the added benefit of accurately capturing forecast uncertainty.

\end{abstract}

\section{Introduction}
\label{sec:introduction}
Forecasting the dynamics of Earth's atmosphere is a scientific problem of utmost importance.
Society is dependent on fast and informative weather forecasts for planning in areas such as transportation and agriculture and for balancing the energy system \cite{quiet_revolution_nwp}. %
Especially important is the use of forecasts to issue warnings for extreme weather events \cite{extreme_weather_definitions}.
Recent advances in \gls{NeurWP} have enabled models that produce accurate forecasts in a fraction of the time of traditional physics-based systems \cite{fourcastnet, panguweather, graphcast}.
So far these developments have largely been focused on deterministic modeling.
However, forecasting only one likely weather scenario ignores the many uncertainties in predicting future weather.

Weather is a chaotic system, resulting in high forecast uncertainty \cite{challenges_of_conv_scale_nwp}.
This uncertainty comes from both imperfect representations of initial states and inaccurate descriptions of the function mapping from one time step to the next \cite{ens_forecasting_overview}.
Accurately modeling this uncertainty significantly increases the value of weather forecasts.
Such uncertainty can be communicated to end-users to improve decision making or be used in downstream products, for example to compute a distribution over solar power generation.
Capturing the full forecast uncertainty requires us to predict not just a single likely state trajectory, but a collection of possible future weather states.
Due to the complexity and dimensionality of the weather system the feasible way to achieve this is by generating samples from a modeled distribution.
Such \textit{ensemble forecasting} is today performed using physics-based methods, where a number of \textit{ensemble members} are simulated as samples from this distribution.
The computational cost of this is however massive, often limiting the spatial resolution or size of the ensemble \cite{quiet_revolution_nwp}.

\gls{NeurWP} is a promising approach for addressing this limitation and enabling large ensemble forecasts.
However, for the ensemble to add value the machine learning model needs to accurately represent the distribution.
Initial attempts at \gls{NeurWP} ensemble forecasting either rely on ad-hoc initial state perturbations \cite{fuxi, fourcastnet, panguweather} or have not been scaled to spatial resolutions of interest \cite{swinvrnn}.
Also diffusion models \cite{denoising_diffusion} have been applied to the problem, but sampling forecasts from these is computationally expensive and can be prohibitively slow  \cite{gencast}.
We propose a Graph-based Ensemble Forecasting Model (\himodelprob), enabling efficient sampling of ensemble members with only one forward-pass per time step.
The method builds on graph-based \gls{NeurWP} \cite{keisler, graphcast}, which is a flexible framework that can be adapted to different geometries and state grid representations \cite{aifs}.
By combining a latent-variable formulation with a hierarchical \gls{GNN} the distribution is modeled in a lower-dimensional space and sampled forecasts are spatially coherent.

\gls{NeurWP} models are typically trained for and evaluated on global weather forecasting \cite{weatherbench2, graphcast, panguweather}. 
Another common forecasting setup in practice is the use of \glspl{LAM} to produce high-resolution regional forecasts \cite{fundamentals_of_nwp}.
Such \glspl{LAM} are for example used by local weather services in order to provide forecasts tailored to the geographical properties and societal needs of the region \cite{lam_germany, lam_france, lam_uk, arome_metcoop}.
These high-resolution models are also invaluable to various industrial sectors, including energy forecasters, who rely on precise weather predictions to manage supply and demand.
This motivates research into also constructing \gls{NeurWP} \glspl{LAM}, which brings new challenges related to the high resolution and boundary conditions of the limited area.
In this work we experiment not just with global forecasting, but consider also how probabilistic \glspl{LAM} can be trained to produce forecasts for the Nordic region.

\textbf{Our main contributions are:}
\begin{inparaenum}[1)]
    \item We develop a hierarchical \gls{GNN} framework for both deterministic and probabilistic \gls{NeurWP}.
    The hierarchical construction encourages spatially coherent fields in forecasts.
    \item We use this framework to define the probabilistic weather forecasting model \hbox{\himodelprob}, capable of efficient sampling of arbitrarily large ensemble forecasts.
    \item We develop a training method targeting both forecast quality and ensemble calibration.
    \item We experiment with both global forecasting on 1.5\textdegree{} resolution and a novel limited-area modeling task at 10 \si{\kilo\metre} resolution.
\end{inparaenum}

\section{Related Work}
\paragraph{Deterministic \gls{NeurWP}}
\looseness=-1 Multiple machine learning methods have been successfully applied to large-scale weather forecasting.
These include graph-based models \citep{keisler, graphcast, aifs}, transformers \citep{panguweather, fengwu, fuxi, climax, atmorep, stormer} and neural operators \cite{fourcastnet, sphericalfno}.
While large neural network models learn weather dynamics purely from data, there are also parallel developments in building hybrid physics-\gls{NeurWP} models \citep{neural_gcm, climode}.

\paragraph{Ensembles from perturbations}
Most existing methods for \gls{NeurWP} ensemble forecasting follow closely the physics-based methods, where initial states and model parameters are \textit{perturbed} to create ensemble diversity.
A number of \gls{NeurWP} works create ensembles by ad-hoc perturbing initial states with random noise \cite{fuxi, fourcastnet, panguweather, calibration_of_large_neurwp, uq_for_data_models}.
More informed perturbations have been re-used from physics based ensembles \cite{gencast, uq_for_data_models} and created based on model-informed singular vectors \cite{neural_ensemble_svd}.
Others try to perturb the forecast model itself, rolling out ensemble members using different neural network parameters \cite{dlwp_ensemble, neural_ensemble_svd}.
Such multi-model approaches require training, or at least fine-tuning, a pre-defined number of \gls{NeurWP} models.
\citet{calibration_of_large_neurwp} use the SWAG method \cite{swag} to allow for constructing multi-model ensembles of arbitrary size.

\paragraph{Generative modeling}
Probabilistic machine learning approaches aim to directly learn generative models producing ensemble members.
Similar to our approach, the SwinVRNN model \cite{swinvrnn} uses a latent variable formulation, but combined with a Swin Transformer architecture \cite{swin_transformer}.
SwinVRNN is developed for global forecasting at 5\textdegree{} resolution and scales poorly to higher spatial resolutions.
Also building on the graph-based framework, \citet{gencast} train a diffusion model \cite{denoising_diffusion,diffusion_score_matching} to sample each time step.
Their Gencast model produces ensemble forecasts of 0.25\textdegree{} global data with 12 h time steps.
Diffusion models produce realistic-looking samples, but typically require solving an ordinary differential equation involving multiple passes through the neural network to sample each time step.
For GenCast, this results in a sampling time of \qty{8}{minutes} for a single 15 day forecast on a TPUv5 device \cite{gencast}.
Other works use diffusion models to increase the size of physics-based ensembles \cite{seeds} or stochastically downscale deterministic forecasts \cite{swinrdm,corrdiff}.

\paragraph{Hierarchical \glspl{GNN}}
Motivated by capturing multiple spatial scales, hierarchical \glspl{GNN} have been used for modeling general partial differential equations \cite{multiscalemgn, simulating_cont_dynamics}.
The overall hierarchical framework shares much of its structure with the popular U-Net architecture \cite{unet} for computer vision tasks, but extended to a general graph setting.

\section{Background}
\subsection{Problem Definition}
\looseness=-1 The weather forecasting problem can be summarized as mapping from a set of initial states $\initstates = (\wstate^{-1}, \wstate^{0})$ to the sequence of future states $\forecast = (\wstate^1, \dots, \wstate^\forecastlength)$.
A table of notation is provided in \cref{sec:table_of_notation}.
Each weather state $\wstate^t \in \R^{\ngridpoints \times \statedim}$ here contains $\statedim$ variables modeled at $\ngridpoints$ different locations.
Geospatial data is often represented as regular grids, in which case these locations correspond to the grid cells.
The $\statedim$ variables can include both atmospheric variables, modeled at multiple vertical levels, and surface variables.
As is common in \gls{NeurWP} we assume the initial states to consist of two time steps, which allows for capturing first-order state dynamics. 
To produce a forecast, a set of forcing inputs $\allforcing$ are also available.
These contain known quantities, such as the time of day.
There are also static features associated with the grid cells, such as the orography, which we here consider part of the forcing.

Many variables impact the chaotic weather system, all of which are not fully captured in initial states represented on finite grids.
This induces forecast uncertainty, which we view as a distribution $\modeldist$.
In deterministic forecasting we seek a model that minimizes the \gls{MSE} to the future weather states \citep{graphcast, climax, stormer}.
This is equivalent to modeling only the mean of the distribution.
In probabilistic forecasting we instead aim to model the full distribution.
Note that we here specifically model the \emph{conditional} distribution $\modeldist$, rather than $p\left(\forecast \middle| \allforcing \right)$.
Hence we do not marginalize over uncertainty in initial states.

\subsection{Graph-based Weather Forecasting}
Graph-based \gls{NeurWP} models use an autoregressive mapping $\pred^{\timei} = \arfunc(\wstate^{\timei - 2:\timei-1}, \forcing^\timei)$ consisting of a sequence of \glspl{GNN} \cite{keisler, graphcast, aifs}.
Starting from the initial states, this mapping can be iteratively applied to roll out a full forecast $\forecast$.
Central to the graph-based framework is the idea of mapping from the original $\ngridpoints$ grid locations to a \textit{mesh graph} $\graph_\mesh = (\meshnodes, \edgeset_\mesh)$.
In the graph-context we refer to the grid locations as a set $\gridnodes$ of \textit{grid nodes}.
By choosing $\setsize{\meshnodes} < \setsize{\gridnodes} = \ngridpoints$ it becomes efficient to perform the majority of computations on the mesh.
Such a mesh graph can also be tailored to the forecasting setting, for example to respect the spherical geometry in global forecasting \cite{keisler}.
The mapping $\arfunc$ realizes a single-step prediction by passing $\wstate^{\timei - 2:\timei-1}$ and $\forcing^\timei$ through a series of \gls{GNN} layers.
In sequence, these layers:
\begin{inparaenum}[1)]
    \item map grid inputs to representations on the mesh graph; 
    \item perform a number of processing steps on the mesh;
    \item map back to the grid to produce the prediction for $\wstate^{\timei}$.
\end{inparaenum}
Steps 1 and 3 use bipartite graphs
$\graph_\gtm = (\gridnodes \cup \meshnodes, \edgeset_\gtm)$
and
$\graph_\mtg = (\gridnodes \cup \meshnodes, \edgeset_\mtg)$
with edges connecting the grid and mesh nodes.
The \gls{GNN} layers in each step compute updates for node representations $\noderep \in \R^{\setsize{\nodeset} \times \latentdim}$ and edge representations $\edgerep \in \R^{\setsize{\edgeset} \times \latentdim}$ in the graphs.
For simplicity all representation vectors have dimensionality $\latentdim$.

\paragraph{Interaction Networks}
The specific \gls{GNN} layers used in previous works are \textit{Interaction Networks} \cite{interaction_nets, graphcast}.
The layers in these networks pass messages from a set of sender nodes along directed graph edges to a set of receiver nodes.
Based on these messages the edge and receiver node representations are then updated.
For a graph $\graph = (\nodeset, \edgeset)$ let $\edgevec{}{\nodei}{\nodeii} \in \R^{\latentdim}$ be the row of $\edgerep$ corresponding to the edge $(\nodei,\nodeii) \in \edgeset$.
Let $\noderep^\sender$ be the matrix with rows containing sender node representations and $\noderep^\receiver$ the corresponding matrix for receiver nodes.
Interaction Networks then implement the representation update
$\noderep^\receiver, \edgerep \gets \gnn(\graph, \noderep^\sender, \edgerep, \noderep^\receiver)$
as
\alse[interaction_net]{
    \edgevecprime{}{\nodei}{\nodeii} &\leftarrow \mlp\left(\edgevec{}{\nodei}{\nodeii}, \noderep^\sender_{\nodei}, \noderep^\receiver_{\nodeii}\right)\\
    \edgevec{}{\nodei}{\nodeii} &\leftarrow \edgevec{}{\nodei}{\nodeii} + \edgevecprime{}{\nodei}{\nodeii}
    \qquad
    \noderep^\receiver_{\nodeii} \leftarrow \noderep^\receiver_{\nodeii} + \mlp\left(\noderep^\receiver_{\nodeii}, {\textstyle \sum_{\nodei \in \neigh{\nodeii}}} \edgevecprime{}{\nodei}{\nodeii}\right)
}
\looseness=-1
where $\neigh{\nodeii} = \set{\nodei: (\nodei,\nodeii) \in \edgeset{}}$ are the incoming neighbors of node $\nodeii$.
Parameters in \glspl{MLP} are shared across nodes and edges in the graph, but not between \gls{GNN} layers.

\paragraph{Global mesh graphs}
\citet{keisler} proposed to construct a mesh graph for global \gls{NeurWP} as an icosahedral grid covering the globe.
This approach was extended in the \gc model \cite{graphcast} by introducing a multi-scale mesh graph with edges of varying length.
Such multi-scale edges are capable of propagating information and capturing statistical dependencies both locally and over long distances in the graph.
The multi-scale mesh graph is created by sequentially splitting the faces of an icosahedron into a sequence of graphs $\graph_\himaxl, \dots, \graph_1$ with node sets satisfying $\nodeset_\himaxl \subset \dots \subset \nodeset_1$ by construction.
The original icosahedron $\graph_\himaxl$ has the longest edges $\edgeset_\himaxl$, stretching far across the globe, whereas the final graph $\graph_1$ has short edges $\edgeset_1$ only connecting nodes locally.
The final multi-scale mesh graph is constructed as $\msgraph = (\nodeset_1, \edgeset_\himaxl \cup \dots \cup \edgeset_1)$, taking the nodes from the final graph but connecting these using edges of all different lengths \cite{graphcast}.

\section{Weather Forecasting with Hierarchical Graph Neural Networks}
\begin{figure}[t]
    \centering
    \includegraphics[width=\textwidth]{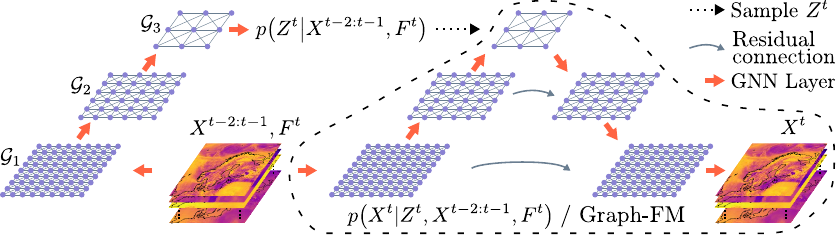}
    \caption{
    Overview of our \himodelprob model, with example data and graphs for a \acrlong{LAM}. 
    The corresponding overview for the global setting is given in \cref{fig:model_overview_global} in \cref{sec:model_details}.
    }
    \label{fig:model_overview_lam}
\end{figure}

Two great challenges in weather forecasting is to accurately capture processes unfolding over different spatial scales and modeling the uncertainty in the chaotic system \cite{challenges_of_conv_scale_nwp}.
To tackle these challenges, we propose to construct a hierarchical mesh graph, working with different length scales at each level in the hierarchy. 
We use a sequence $\graph_1, \dots, \graph_\himaxl$ of graphs as the different levels in the hierarchy, additionally adding connections between the nodes of adjacent levels.
This construction is also highly suitable as a basis for building probabilistic forecasting models, as discussed below.
\Cref{fig:model_overview_lam} shows an overview of the hierarchical mesh used in our model.
See \cref{fig:global_graph_plots,fig:lam_graph_plots} in the appendix for illustrations of how this differs from the multi-scale graph. 

\looseness=-1 There are multiple benefits to such a hierarchical mesh construction for \gls{NeurWP}.
By keeping the graphs at different levels separate, we can define \gls{GNN} layers with independent parametrizations at each level.
This adds flexibility by allowing the model to learn different representation updates for edges of different spatial scales.
A hierarchical mesh graph also offers a natural, spatially-aware dimensionality reduction, as the state in the grid is encoded into a few nodes at the top level.
Such a representation can capture the general structure of each weather state, with finer details added as this is propagated down through the hierarchy.
We leverage this property to construct a probabilistic model by imposing a distribution over these lower-dimensional representations at the top level.
This allows for efficiently drawing spatially coherent samples from the distribution of future weather states.

\subsection{Hierarchical Graph}
Our hierarchical mesh graph consists of $\himaxl$ graph levels $\hiintragraph{1}, \dots, \hiintragraph{\himaxl}$ with $\hiintragraph{\leveli} = (\himeshnodes{\leveli}, \himeshedges{\leveli})$.
Only level 1 of the hierarchy is connected to the grid, so we re-define $\graph_\gtm = (\gridnodes \cup \himeshnodes{1}, \edgeset_\gtm)$ and $\graph_\mtg = (\gridnodes \cup \himeshnodes{1}, \edgeset_\mtg)$.
The number of nodes $\setsize{\himeshnodes{\leveli}}$ decreases with the level $\leveli$.
The smallest set of nodes are found at the top level $\himaxl$.

\looseness=-1 To pass information between the levels of the hierarchy we introduce additional graphs connecting the different levels.
Let $\hiintergraph{\leveli}{\leveli+1} = (\himeshnodes{\leveli} \cup \himeshnodes{\leveli+1}, \hiinteredges{\leveli}{\leveli + 1})$ be a graph containing directed edges from mesh level $\leveli$ to level $\leveli + 1$.
We make use of a graph sequence $\hiintergraph{1}{2}, \dots, \hiintergraph{\himaxl - 1}{\himaxl}$ to propagate information up through the hierarchy and similarly a sequence $\hiintergraph{\himaxl}{\himaxl - 1}, \dots, \hiintergraph{2}{1}$ in the downward direction.
The exact layout of nodes and edges at and in-between levels are design choices that should be tailored to the specific forecasting setting.
Examples for global and limited-area forecasting are given in \cref{sec:experiments}. %

\subsection{\himodeldet: Deterministic Forecasting}
The hierarchical graph allows for defining \gls{GNN} layers both on and in-between the different levels.
By sequentially updating node and edge representations at different levels in the hierarchy, information can be propagated up from the grid to the different levels.
As these levels have edges of different lengths, the processing at each level happens on different spatial scales.
Note that this differs from the multi-scale graph approach, where information processing over all different spatial scales happen in the same \gls{GNN} layer \cite{graphcast}.
As a step towards our probabilistic model, we define an alternative deterministic Graph-based Forecasting Model \textit{\himodeldet}\footnote{The deterministic \himodeldet model was first proposed in a preliminary version of this work \cite{neural_lam}, but there only for the \gls{LAM} setting under the name \textit{Hi-LAM}.}, operating on the hierarchical graph.

In \himodeldet one processing step on the mesh graph is defined as a complete sweep through the hierarchy.
\glspl{GNN} are applied sequentially to the inter-level and intra-level graphs in the order $\hiintragraph{1}, \hiintergraph{1}{2}, \hiintragraph{2}, \dots, \hiintergraph{\himaxl-1}{\himaxl}, \hiintragraph{\himaxl}$, updating edge and node representations at the different levels.
Processing steps going up the hierarchy are alternated with similar steps going down from level $\himaxl$ to $1$.
The single step mapping $\arfunc$ consists of multiple such sweeps up and down (see \cref{sec:hi_det_details}).

\subsection{\himodelprob: Probabilistic Forecasting}
\begin{wrapfigure}[9]{r}{0.25\textwidth}%
    \centering%
    \vspace{-2.8em}%
    \includegraphics[width=0.25\textwidth]{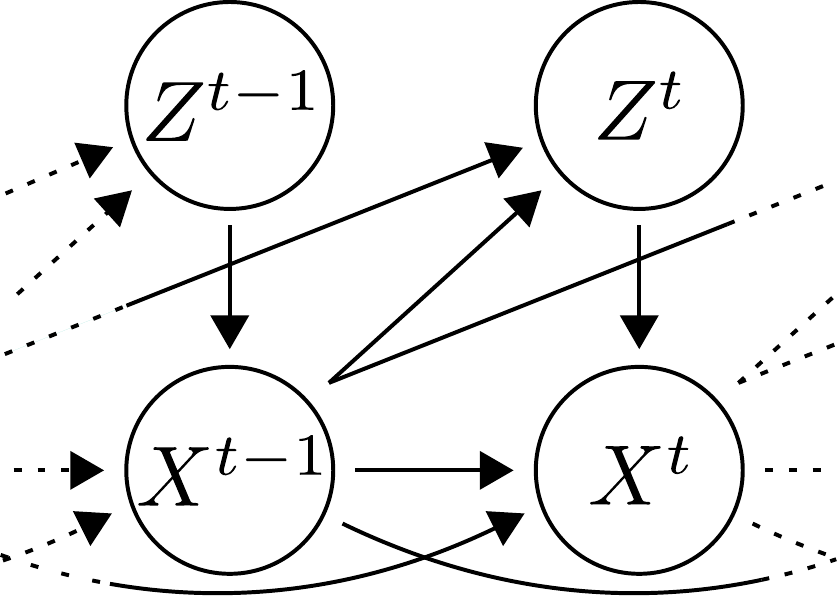}%
    \caption{Graphical model for \cref{eq:latent_var}.}
    \label{fig:graphical_model}
\end{wrapfigure}
To capture the uncertainty in the chaotic weather system we next aim to construct a probabilistic model from the ground up to capture the full distribution $\modeldist$.
We start by assuming the weather system to satisfy a second-order Markov assumption, decomposing
\eq[markov_decomp]{
    \modeldist = {\textstyle \prod_{\timei = 1}^{\forecastlength}} \ardist.
}
Factoring the distribution over time steps allows us to work with forecasts of varying length.
Specifying the model for single-step prediction avoids having to learn separate parameters for different lead times.
Next, we seek a flexible, but computationally efficient parametrization for the distribution $\ardist$.
This can be achieved by introducing a latent random variable $\latent^\timei$, and letting
\eq[latent_var]{
    \ardist = \int \preddist \priordist d\latent^{\timei}.
}
Here the stochasticity in $\latent^\timei$ should capture the uncertainty over $\wstate^\timei$ at each time step.
The corresponding graphical model is shown in \cref{fig:graphical_model}.
We impose a spatial structure over the latent variable by letting $\latent^\timei$ be $\setsize{\himeshnodes{\himaxl}} \times \latentdim$ matrix-valued, with each row a $\latentdim$-dimensional vector associated with one node in the top level $\hiintragraph{\himaxl}$ of the mesh graph.

The single-step model consists of two components, a latent map $\priordist$ and predictor $\preddist$.
The latent map is parametrized using \glspl{GNN}, mapping the conditioning variables to parameters of a Gaussian distribution.
We consider the predictor to be concentrated around its mean, and realize $\preddist$ as a deterministic mapping of a similar form as \himodeldet.
By sampling $\latent^\timei$ and passing this through the predictor we can draw a sample of $\wstate^\timei$ from \cref{eq:latent_var}.
This sample can then be conditioned on at the next time step, continuing this sampling process to roll out a forecast following \cref{eq:markov_decomp}.
This forecast constitutes one ensemble member, and the process can be repeated to sample an ensemble of arbitrary size.
We call our Graph-based Ensemble Forecasting Model \textit{\himodelprob}.
Full details about the model are given in \cref{sec:model_details}.

\paragraph{Latent map}
We let the latent map be an isotropic Gaussian
\eq[iso_prior]{
    \priordist = {\textstyle \prod_{\nodei \in \himeshnodes{\himaxl}}}
     \normalpdf{\latent^t_\alpha}{\meanfunc{\latent}\left(\arcond \right)_\nodei}{I}
}
with the mean as a function of the conditioning variables. 
The variance is fixed, imposing a fixed scale for the learned latent space.
The mean function $\meanfunc{\latent}$ consists of a sequence of \glspl{GNN}.
These take the inputs at the grid, propagate representations up through the hierarchical mesh graph, and finally predicts the mean of $\latent^{\timei}_{\nodei}$ at each node $\nodei$ at level $\himaxl$.
In \cref{sec:latent_map_exp} we verify empirically the importance of using the latent map over a static distribution for $\latent^\timei$.

\paragraph{Predictor}
The predictor is a deterministic mapping
\newcommand{\predictoreq}{\pred^\timei = \predictorfunc\left(\latent^\timei, \arcond\right) =\wstate^{\timei-1} + \predictorfunctilde\left(\latent^\timei, \arcond\right)}
\eq[predictor]{
    \predictoreq.
}
With the small time steps used in \gls{NeurWP}, $\wstate^\timei$ does not change dramatically in a single step.
We thus follow the common practice of including a skip connection to the previous state \cite{graphcast, sphericalfno, swinvrnn}.
The predictor takes both inputs $\arcond$ at the grid and $\latent^\timei$ at the top of the mesh graph.
To incorporate both we design $\predictorfunc$ similar to \himodeldet, performing sweeps up and down through the mesh hierarchy.
At the top of the hierarchy $\latent^\timei$ is added to node representations $\noderep^\himaxl$ through the residual connections in the \gls{GNN} layers.
A sampled value of $\latent^\timei$ then affects the prediction $\pred^\timei$ through the downward sweep.
While multiple such sweeps are possible, we found one to be sufficient in practice.

\paragraph{Spatial dependencies}
We want each sample of $\wstate^\timei$ to contain spatially coherent atmospheric fields.
One approach would be to impose spatial dependencies in the joint distribution over $\latent^t$.
However, learning and sampling from such a distribution typically comes with computational challenges~\cite{swinvrnn}.
Instead, we impose spatial dependencies by integrating the latent variable formulation with the hierarchical graph.
We argue that as the independent components of $\latent^\timei$ are propagated down through the mesh graph, gradually increasing the spatial resolution, spatial dependencies are introduced by the model in the \gls{GNN} layers.
The hierarchical graph is key to this property, as the stochasticity in $\latent^\timei$ is necessarily spread out over the forecast region, rather than only affecting the output locally.

\subsection{Training Objective}
Deterministic forecasting models can be straightforwardly trained by minimizing a weighted \gls{MSE}~\cite{graphcast} or \gls{NLL} loss \cite{fengwu} for rolled out forecasts.
To train \himodelprob we instead leverage the fact that the single-step model has a structure similar to a (conditional) \gls{VAE} \cite{vae, cvae}, allowing us to use a variational objective.
We introduce a variational approximation $\vardist$ at each time step, approximating the true posterior $\postdist$ over $\latent^\timei$.
This variational distribution is parametrized in a similar way as the latent map, with \gls{GNN} layers mapping to a Gaussian over $\latent^\timei$.
Note however that $\vardistname$ also depends on $\wstate^\timei$, since it approximates the posterior.
Using $\vardistname$, we can then define
\als[elbo_cvae]{
    \elboloss&\left(\wstate^{\timei-2:\timei-1}, \wstate^\timei, \forcing^{\timei} \right)
    =
    \klweight \kl{\vardist }{\priordist}
    \\
    -&\E{\vardist}{
        {\textstyle \sum_{\nodei \in \gridnodes}
        \sum_{\vari = 1}^{\statedim}}
        \log \normalpdf{
            \wstate^{\timei}_{\nodei, \vari}
        }{
            \predictorfunc\left(\latent^\timei, \arcond\right)_{\nodei, \vari}
        }{
            \predsigma^2_{\nodei, \vari}
        }
    }
}
which is equal to the (negative) \gls{ELBO} when the weighting is $\klweight = 1$.
While the predictor $\predictorfunc$ is a deterministic mapping, we introduce a Gaussian likelihood in  \cref{eq:elbo_cvae} to get a well-defined learning problem.
This setup corresponds to the common practice in \glspl{VAE} of assuming Gaussian observation noise, but not adding this to samples from the model \cite{effective_vae_training}.
The standard deviation $\predsigma_{\nodei, \vari}$ can either be a second output from the predictor or manually chosen (see \cref{sec:trainin_objectives_extra} for details).
As with deterministic models \cite{graphcast, keisler, fengwu, stormer}, we found it crucial to fine-tune on rolled out forecasts of multiple time steps. 
This improves stability and performance for longer lead times.
In the final fine-tuning we include also a \gls{CRPS} loss term $\crpsloss$ \cite{proper_scoring_rules,neural_gcm}. 
The full objective function is then $\loss = \elboloss + \crpsweight \crpsloss$, with $\crpsweight$ a weighting hyperparameter.
Including this \gls{CRPS} loss improves the calibration of ensemble forecasts.

\subsection{Improved GNN Layers: Propagation Networks}
In \himodelprob there is a large amount of information that needs to be propagated between the grid and $\latent^\timei$.
However, the Interaction Network \glspl{GNN} are biased towards keeping old representations of receiver nodes, rather than updating this with new information from incoming edges.
Note in \cref{eq:interaction_net} that if the \glspl{MLP} are initialized to give outputs close to 0, there will be no change to $\edgevec{}{\nodei}{\nodeii}$ and $\noderep^\receiver_\nodeii$.

In practice the model has a hard time learning to propagate useful information up from the grid to $\latent^\timei$.
Even when trained purely as an auto-encoder ($\klweight = 0$), $\latent^\timei$ easily ends up being ignored.
To remedy this we propose an alternative \gls{GNN} formulation that we call \textit{Propagation Network}, defined by
\alse[propagation_net2]{
    \edgevecprime{}{\nodei}{\nodeii} &\leftarrow \noderep^\sender_{\nodei} + \mlp\left(\edgevec{}{\nodei}{\nodeii}, \noderep^\sender_{\nodei}, \noderep^\receiver_{\nodeii}\right)&
    \edgevec{}{\nodei}{\nodeii} &\leftarrow \edgevec{}{\nodei}{\nodeii} + \edgevecprime{}{\nodei}{\nodeii}\\
    \noderepprime^\receiver_{\nodeii} &\leftarrow \frac{1}{\setsize{\neigh{\nodeii}}}{\textstyle \sum_{\nodei \in \neigh{\nodeii}}} \edgevecprime{}{\nodei}{\nodeii}&
    \noderep^\receiver_{\nodeii} &\leftarrow \noderepprime^\receiver_{\nodeii} + \mlp\left(\noderep^\receiver_{\nodeii}, \noderepprime^\receiver_{\nodeii} \right).
}
For \glspl{MLP} initialized with outputs close to $0$, Propagation Networks reduce to averaging the values of neighboring nodes.
This encourages the propagation of information from $\noderep^\sender$ to $\noderep^{\receiver}$ by construction.
Propagation Networks were found to perform better also in the deterministic model (see comparison in \cref{sec:propnet_exp}), so we employ these in both \himodeldet and \himodelprob.

\section{Experiments}
\label{sec:experiments}
To evaluate our models we conduct experiments on both global and limited area forecasting.
\looseness=-1 The models are implemented\footnote{\ghfootnote} in PyTorch and trained on 8 A100 \qty{80}{\giga\byte} GPUs in a data-parallel configuration.
Training takes 700--1400 total GPU-hours for the global models, and around half of that for the limited area models.
The computational demands prevent us from re-training multiple models for statistical analysis.
Once trained, sampling from \himodelprob is highly efficient. 
Using batched sampling on a single GPU, 80 ensemble members are produced in \qty{200}{\second} (\qty{2.5}{\second} per member) for global forecasting.

\paragraph{Metrics}
We measure the skill of deterministic models by \glsfirstbold{RMSE}.
For probabilistic models we compute the \gls{RMSE} for the ensemble mean.
Good skill in terms of \gls{RMSE} is however not enough for ensemble forecasts, where we want to capture the full distribution.
For these we also assess the ensemble calibration by computing the \glsfirstbold{SPSKR}.
Calibrated uncertainty corresponds to $\spskr \approx 1$ \citep{why_spskr}.
We additionally use \textbf{\gls{CRPS}} to measure how well the marginal distributions of the model matches the data.
For deterministic models the \gls{CRPS} reduces to \gls{MAE}.
Complete definitions of all metrics are given in \cref{sec:metric_defs}.

\paragraph{Models}
Achieving a fair comparisons of the actual machine learning methodology in \gls{NeurWP} is challenging due to models using different spatial resolution, variables and initial states.
We here train an illustrative set of models on the same data and with comparable training setups.
Our full \hbox{\textbf{\himodelprob}} model is compared to:
\begin{inparaenum}[1)]
    \item \textbf{\msmodelprob}, a version of \himodelprob using a multi-scale mesh graph instead of the hierarchical one.
    \item \textbf{\himodeldet}, our deterministic model using the hierarchical graph.
    \item \textbf{\gcour}, a reimplementation of GraphCast \cite{graphcast}, adapted and trained on our datasets.
    \item \textbf{\gcswa}, a multi-model ensemble created by applying \gls{SWAG} \cite{swag} to \gcour. Inspired by \citet{calibration_of_large_neurwp}, this represents a simple way to augment a deterministic model to perform ensemble forecasting.
\end{inparaenum}
Further details about the baseline models are given in \cref{sec:baseline_model_details}.
For ensemble models we sample 80 members for the global experiments and 100 members for limited area forecasting.
In \cref{sec:ens_size_experiment} we investigate the impact of ensemble size on the evaluation.
We find that improvements in metric values quickly saturate when increasing the ensemble size.
This shows that sampling even more members would have negligible impact on the results of our experiments.

\subsection{\globalsection}
\paragraph{Data and graphs}
\looseness=-1 We experiment on global weather forecasting up to \qty{10}{days} with \qty{6}{\hour} time steps.
The dataset used for training and evaluation is a 1.5\textdegree{} version of the global ERA5 reanalysis\footnote{Provided by the Copernicus Climate Change Service under the ECMWF Copernicus License.} \cite{era5}, provided through the WeatherBench 2 benchmark \cite{weatherbench2}.
The models forecast $\statedim = 83$ different variables in total, including both surface-level variables and atmospheric variables at 13 different pressure levels.
We use the years 1959--2017 for training, 2018--2019 for validation and 2020 as a test set.
Forecasts are always started from initial conditions taken directly from ERA5, both during training and evaluation.
For global forecasting we use the graph generation process from GraphCast \cite{graphcast}. The multi-scale graph $\msgraph$ is created by refining the icosahedron 4 times. The hierarchical graph contains 4 levels of such icosahedral grids.
More details on the global experiments are given in \cref{sec:global_details}.

\paragraph{Results}
\begin{table}[tbp]
\centering
\caption{
Selection of results for global forecasting, including geopotential at \qty{500}{\hecto\pascal} (\wvar{z500}) and \qty{2}{\meter} temperature (\wvar{2t}).
For \gls{RMSE} and \gls{CRPS} lower values are better, and \spskrabr should be close to $1$ for a calibrated ensemble.
The best metric values are marked with \textbf{bold} and second best \underline{underlined}.
}
\label{tab:global_res_table}
\begin{small}
\begin{tabular}{@{}llcccccc@{}}
\toprule
         &              & \multicolumn{3}{c}{Lead time 5 days}          & \multicolumn{3}{c}{Lead time 10 days}         \\ \cmidrule(lr){3-5} \cmidrule(lr){6-8} 
Variable & Model        & RMSE          & CRPS          & \spskrabr     & RMSE          & CRPS          & \spskrabr     \\ \midrule
\wvar{z500} & \gcour       & \underline{387}  & 236  & -    & 808  & 498  & -    \\
          & \himodeldet  & \textbf{363}  & 223  & -    & 825  & 510  & -    \\
          & \gcswa       & 437  & 269  & 0.07 & 960  & 590  & 0.12 \\
          & \msmodelprob & 472  & \underline{211}  & \underline{0.77} & \underline{756}  & \underline{333}  & \underline{0.83} \\
         & \himodelprob & 399  & \textbf{169}  & \textbf{1.18} & \textbf{695}  & \textbf{299}  & \textbf{1.15} \\ \midrule
\wvar{2t}   & \gcour       & 1.65 & 1.00 & -    & 2.82 & 1.69 & -    \\
          & \himodeldet  & \textbf{1.57} & 0.94 & -    & 2.82 & 1.66 & -    \\
          & \gcswa       & 2.03 & 1.20 & 0.06 & 3.58 & 2.04 & 0.13 \\
          & \msmodelprob & 1.76 & \underline{0.77} & \underline{0.75} & \underline{2.55} & \underline{1.09} & \underline{0.82} \\
         & \himodelprob & \underline{1.64} & \textbf{0.71} & \textbf{0.98} & \textbf{2.32} & \textbf{1.00} & \textbf{0.99} \\ \bottomrule
\end{tabular}
\end{small}
\end{table}

\begin{figure}[tb]
    \centering
    \begin{subfigure}[b]{\textwidth}%
        \centering
        \includegraphics[width=\textwidth]{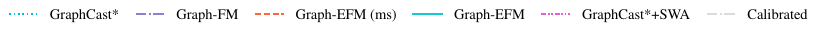}%
    \end{subfigure}
    \mainsubfig{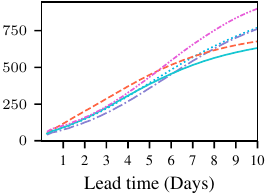}{RMSE}%
    \mainsubfig{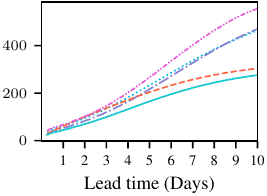}{CRPS}%
    \mainsubfig{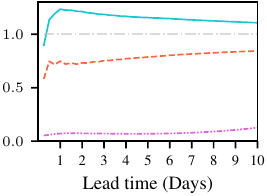}{SPSKR}%
    \caption{Results for global forecasting of mean sea level pressure (\wvar{msl}) at all lead times.}
    \label{fig:msl_main}
\end{figure}

\begin{figure}
    \centering
    \includegraphics[width=\textwidth]{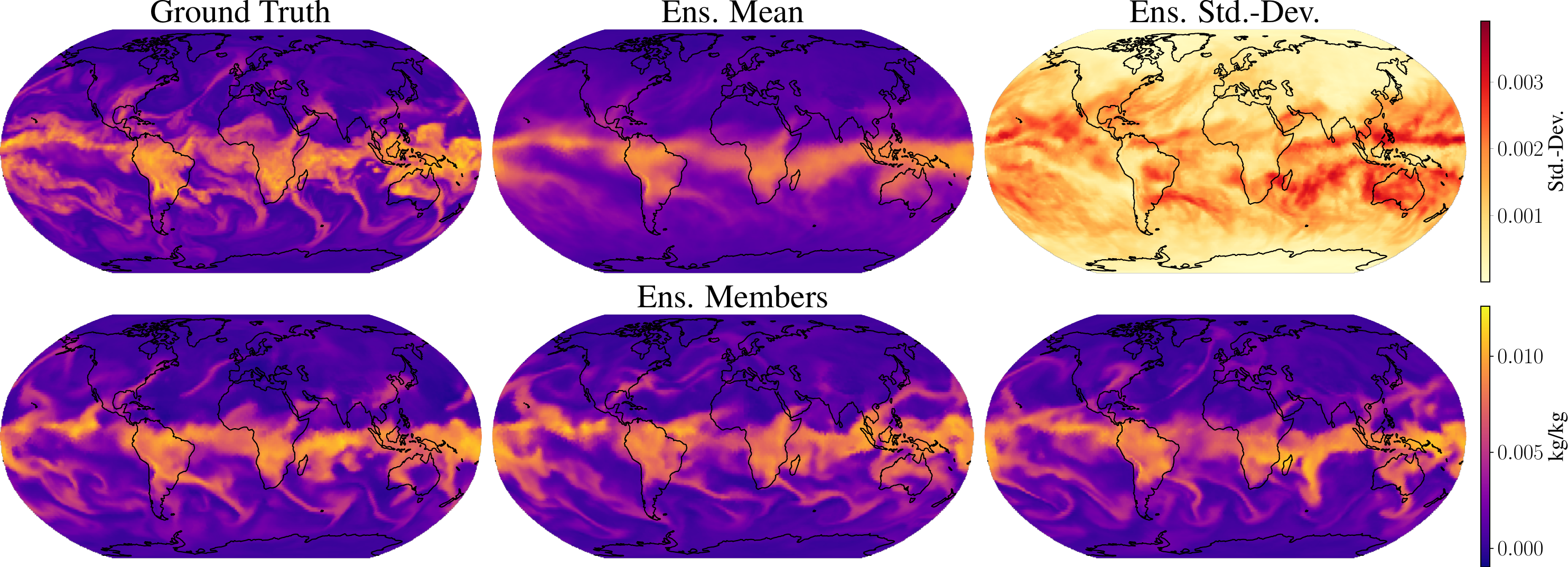}
    \caption{
    Example \himodelprob ensemble forecast for specific humidity at \qty{700}{\hecto\pascal} (\wvar{q700}), for lead time 10 days. 
    The bottom row shows 3 ensemble members, randomly chosen out of the 80.
    }
    \label{fig:global_example_fc_main}
\end{figure}

As the models forecast many different variables we present only a selection of results in the main paper.
Metric values for geopotential (\wvar{z500}) and 2 m temperature (\wvar{2t}) are listed in \cref{tab:global_res_table} and results for mean sea level pressure (\wvar{msl}) plotted in \cref{fig:msl_main}.
Line plots for all metrics and a large number of variables are given in \cref{sec:global_metrics_appendix}.
In the appendix we also show comparisons to additional models from the literature, trained on different data, as well as the physics-based IFS-ENS model \cite{ifs_ens}.
The ensemble mean from \himodelprob often shows improvements in \gls{RMSE} over the deterministic models, especially for longer lead times.
Across the ensemble models, \himodelprob achieves lower \gls{CRPS} values, better capturing the distribution of the weather data.
Without any perturbations to initial states \himodelprob reaches a \gls{SPSKR} close to 1.
We note that \gcswa does not produce useful ensemble forecasts, as these are poorly calibrated and in general do not lead to improved forecast errors.
\Cref{fig:global_example_fc_main} shows an example forecast from \himodelprob for specific humidity (\wvar{q700}) at 10 days lead time.
Examples for other variables are given in \cref{sec:global_forecasts_appendix}.

\paragraph{Extreme weather case study}
\looseness=-1 An important use case for ensemble forecasting is modeling extreme weather events. 
While higher resolutions than 1.5\textdegree{} are generally desirable for accurately capturing such extremes, we conduct one case study on using \himodelprob for forecasting hurricane Laura.
The full case study with visualized forecasts is available in \cref{sec:extreme_case_study}.
For this example we show that there exists ensemble members accurately predicting the landfall location of the hurricane at 7 days lead time, while the deterministic models still show no sign of the hurricane in the region.
Closer to the landfall event the ensemble forecast from \himodelprob indicates uncertainties associated with the landfall location and wind intensity.
This demonstrates the added value of a probabilistic forecasting model.

\subsection{\lamsection}
In \glspl{LAM} weather forecasts are produced for a bounded region of the globe.
\gls{LAM} forecasting allows for higher resolution modeling and regionally tailored model configurations~\citep{fundamentals_of_nwp}, properties that can be inherited by \gls{NeurWP} models by training on \gls{LAM} data.
To model weather over a limited domain, boundary conditions need to be taken into account.
In physics-based \glspl{LAM} these are typically given by a global forecast \cite{lam_germany, lam_france, lam_uk, arome_metcoop}.
We adapt a similar approach for \gls{NeurWP} \glspl{LAM}, by taking boundary conditions as additional forcing along the boundary of the forecast area.
The problem of \gls{LAM} forecasting is thus about simulating physics not just based on the initial state, but also consistent with these boundary inputs.
In the models we introduce $\nboundarynodes$ additional grid nodes along the area boundary, for the boundary forcing $\boundarystate^t \in \R^{\nboundarynodes \times \statedim}$.
Boundary forcing $\boundarystate^t$ is always fed together with $\wstate^t$ to the model.
Grid nodes on the boundary and within the area are treated identically by the \gls{GNN} layers.
We perform this adaptation to all models in our experiment.

\paragraph{Data and graphs}
We experiment with a dataset containing \num{6069} forecasts from the \gls{MEPS} \gls{LAM}.
Training on forecasts, the goal is here to learn a fast surrogate model for \gls{MEPS}.
We use forecasts started during April 2021 -- Jun 2022 for training and validation, and forecasts from July 2022 -- March 2023 as a test set.
The data is laid out in a $238 \times 268$ grid with spatial resolution 10 \si{\kilo\metre}, covering the Nordic region.
This dataset contains in total $\statedim = 17$ weather variables, some repeated on multiple vertical levels.
Forecasts are rolled out with \qty{3}{\hour} time steps up to lead time \qty{57}{\hour}.
In this experiment we take also the boundary forcing directly from the \gls{MEPS} dataset.
We define the boundary as the outermost 10 grid positions.
Using the same dataset for the area and boundary allows us to investigate the modeling choices in a controlled experimental setup.
In an operational scenario the boundary forcing would instead come from a re-gridded global forecast.
In the \gls{LAM} setting we define our graphs as regular quadrilateral meshes covering the MEPS forecasting area, but with far fewer nodes than the original grid.
The graph hierarchy $\graph_1, \dots, \graph_\himaxl$ is created by constructing such meshes at different resolutions.
By placing each node in $\graph_\leveli$ at the center of $3\times3$ nodes in $\graph_{\leveli-1}$, we can merge 4 such graph levels to create $\msgraph$.
In the hierarchical graph we instead introduce edges from each node in $\graph_\leveli$ to the $3\times3$ nodes in the level below.
More details about the \gls{MEPS} data and experiment can be found in \cref{sec:lam_details}.

\begin{table}[tbp]
\centering
\caption{
Selection of results for \gls{LAM} forecasting, including geopotential at \qty{500}{\hecto\pascal} (\wvar{z500}) and integrated column of water vapor (\wvar{wvint}).
}
\label{tab:lam_res_table}
\begin{small}
\begin{tabular}{@{}llcccccc@{}}
\toprule
             &              & \multicolumn{3}{c}{Lead time 24 h}            & \multicolumn{3}{c}{Lead time 57 h}            \\ \cmidrule(lr){3-5} \cmidrule(lr){6-8} 
Variable & Model        & RMSE & CRPS & \spskrabr  & RMSE & CRPS & \spskrabr  \\ \midrule
\wvar{z500}  & \gcour       & \textbf{153}  & \underline{108}     & -             & \textbf{201}  & \underline{138}     & -             \\
         & \himodeldet  & 230  & 162  & -          & 354  & 238  & -          \\
         & \gcswa       & 219  & 136  & 0.08       & 376  & 206  & 0.10       \\
         & \msmodelprob & 400  & 261  & \underline{0.22} & 711  & 470  & \underline{0.23} \\
             & \himodelprob & \underline{172}     & \textbf{91}   & \textbf{0.84} & \underline{219}     & \textbf{115}  & \textbf{0.75} \\ \midrule
\wvar{wvint} & \gcour       & \textbf{1.51} & \underline{1.01}    & -             & \textbf{2.06} & \underline{1.32}    & -             \\
         & \himodeldet  & 1.64 & 1.08 & -          & 2.48 & 1.58 & -          \\
         & \gcswa       & 1.78 & 1.17 & 0.05       & 2.34 & 1.50 & 0.05       \\
         & \msmodelprob & 2.39 & 1.43 & \underline{0.16} & 3.51 & 2.12 & \underline{0.13} \\
             & \himodelprob & \underline{1.61}    & \textbf{0.79} & \textbf{0.57} & \underline{2.08}    & \textbf{1.00} & \textbf{0.53} \\ \bottomrule
\end{tabular}
\end{small}
\end{table}
\begin{figure}
    \centering
    \includegraphics[width=\textwidth]{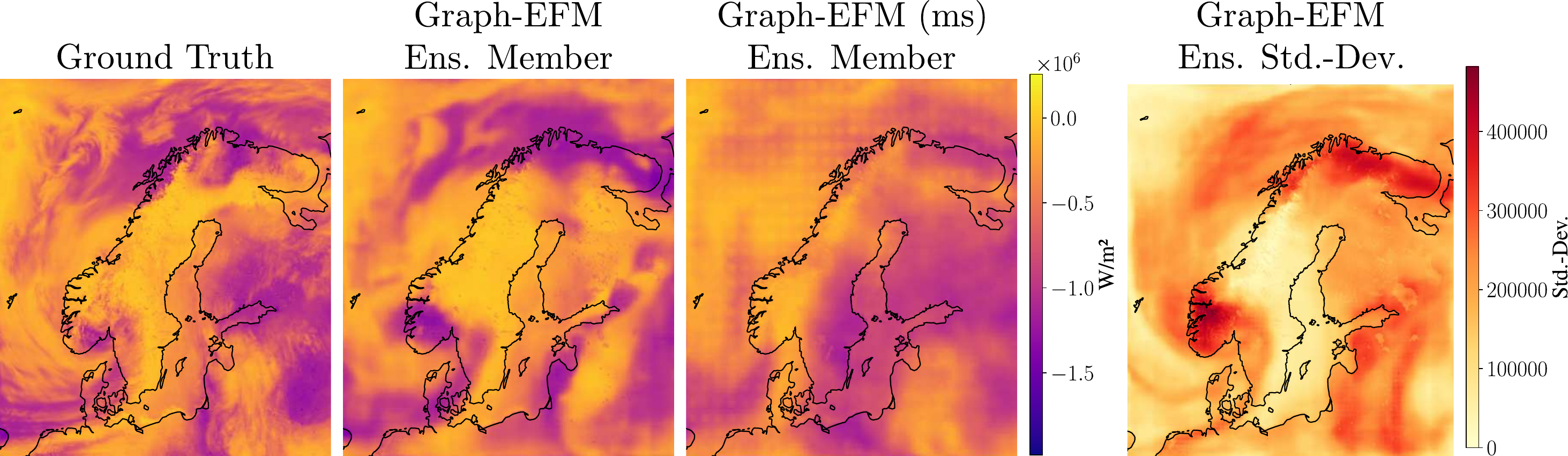}
    \caption{
    Example forecasts for net solar longwave radiation (\wvar{nlwrs}) at lead time \qty{57}{\hour}. 
    }
    \label{fig:lam_example_fc_main}
\end{figure}

\paragraph{Results}
A selection of metrics are shown in \cref{tab:lam_res_table} and full results given in \cref{sec:lam_extra_res}.
At these shorter lead times there is no clear benefit of probabilistic modeling in terms of \gls{RMSE}.
Still, as exemplified by the standard-deviation plotted in \cref{fig:lam_example_fc_main}, probabilistic modeling provides useful information about the forecast uncertainty.
Comparing the ensemble members in \cref{fig:lam_example_fc_main} highlights the improved spatial coherency of the hierarchical graph in \himodelprob.
In contrast, the \msmodelprob forecast looks patchy and lacks physically intuitive features. 
There are also clear visual artifacts, that can be traced to the multi-scale graph structure.
We discuss this more in-depth in \cref{sec:spatial_coherency}.
In the \gls{LAM} setting all models are under-dispersed, with $\spskr < 1$.
One explanation for this is that the boundary forcing constrains the space of plausible forecasts, hindering the ensemble spread.

\section{Discussion}
\label{sec:discussion}
In this paper we have explored \gls{NeurWP} ensemble weather forecasting using graph-based latent variable models.
Our \himodelprob model is capable of efficiently producing accurate ensemble forecasts.
This paves the way for large-scale \gls{NeurWP} ensemble forecasting both in operational use and research settings.
In \cref{sec:societal_impact} we further discuss the societal impact of this research.
With this work we hope to emphasize that \gls{NeurWP} models are not just deterministic mappings, but parametrize distributions of weather states.
It follows that ensemble forecasting should not be achieved by perturbing models, but by directly modeling the distribution of interest.

\paragraph{Limitations}
The training process comes with some complications in terms of choosing a training schedule and hyperparameters $\klweight$ and $\crpsweight$.
While the \gls{CRPS} fine-tuning is an important training step, we have found that choosing a too high $\crpsweight$ can introduce visual artifacts, especially for the \msmodelprob model (see \cref{sec:spatial_coherency}).
While \himodelprob produces diverse and physically plausible ensemble members, the forecasts still suffer from some of the blurriness common to deterministic models \cite{graphcast, weatherbench2}.
We here trade off some of the visual fidelity achieved for example by diffusion models \cite{gencast} for more efficient sampling of ensemble members.

\paragraph{Future work}
Interesting avenues for future work include learning probabilistic weather models based on other types of autoencoders \cite{wae, vq_vae}, or by directly optimizing scoring rules \cite{gen_scoring_rules, neural_gcm}.
Another approach for achieving efficient ensemble forecasting is to explore techniques for speeding up diffusion model sampling \cite{consistency_models}.

\begin{ack}
    This research is financially supported by the Swedish Research Council via the project
\emph{Handling Uncertainty in Machine Learning Systems} (contract number: 2020-04122),
the Wallenberg AI, Autonomous Systems and Software Program (WASP) funded by the Knut and Alice Wallenberg Foundation,
the Excellence Center at Linköping--Lund in Information Technology (ELLIIT),
and
the project \emph{OWGRE}, funded by partners of the ERA-Net Smart Energy Systems and Mission Innovation through the Joint Call 2020. As such, this project has received funding from the
European Union’s Horizon 2020 research and innovation programme under grant agreement no. 883973.
Our computations were enabled by the Berzelius resource at the National Supercomputer Centre, provided by the Knut and Alice Wallenberg Foundation.
\end{ack}

\bibliography{references}
\newpage
\appendix
\startcontents[sections]
\section*{Appendix Table of Contents}
\printcontents[sections]{l}{1}{\setcounter{tocdepth}{1}}
\FloatBarrier

\section{Table of Notation}
\label{sec:table_of_notation}
Notation used throughout the paper is listed in \cref{tab:notation}.
\newcommand{\tonrow}[2]{$#1$ & #2 \\}
\newcommand{\tonseparator}{\midrule}
\begin{table}[tbp]
\centering
\caption{
Table of notation 
}
\label{tab:notation}
\renewcommand{\arraystretch}{1.1}
\begin{tabular}{@{}cl@{}}
\toprule
\tonrow{\wstate^\timei}{Weather state at time step $\timei$}
\tonrow{\pred^\timei}{Predicted weather state at time step $\timei$}
\tonrow{\wstate^{\timei, (\samplei)}}{Ground truth weather state at time step $\timei$ in sample $\samplei$ from dataset}
\tonrow{\pred^{\timei, (\samplei)}}{Predicted weather state at time step $\timei$ in sample $\samplei$ from dataset}
\tonrow{\pred^{\timei, (\samplei), (\ensi)}}{Prediction of ensemble member $\ensi$ at time step $\timei$ in sample $\samplei$ from dataset}
\tonrow{\forcing^\timei}{Forcing inputs at time step $\timei$}
\tonrow{\latent^\timei}{Latent random variable for time step $\timei$}
\tonrow{\boundarystate^\timei}{Boundary forcing input at time step $\timei$, for limited area modeling}
\tonrow{\noderep}{Matrix with node representation vectors as rows}
\tonrow{\edgerep}{Matrix with edge representation vectors as rows}
\tonrow{\edgevec{}{\nodei}{\nodeii}}{Representation vector for the edge going from node $\nodei$ to node $\nodeii$}
\tonseparator

\tonrow{\arfunc}{Autoregressive mapping predicting the next state in deterministic models}
\tonrow{\predictorfunc}{Predictor part of \himodelprob, predicting the next state conditioned on a sample of $\latent^\timei$}
\tonrow{\predictorfunctilde}{Predictor output before skip connection}
\tonrow{\meanfunc{\latent}}{Mean function of latent map, mapping to the mean of $\latent^\timei$}
\tonseparator

\tonrow{\forecastlength}{Length of one forecast, in discrete time steps}
\tonrow{\ngridpoints}{Number of grid nodes (grid cells)}
\tonrow{\nboundarynodes}{Number of grid nodes in the boundary region, for limited area modeling}
\tonrow{\himaxl}{Number of levels in hierarchical graph}
\tonrow{\samplesize}{Number of samples (forecasts) in dataset}
\tonrow{\enssize}{Ensemble size, number of members in ensemble forecast}
\tonrow{\statedim}{Weather state dimensionality (total number of variables) in each grid node}
\tonrow{\latentdim}{Latent dimensionality, dimensionality of each node or edge representation vector}
\tonseparator

\tonrow{\graph_\mesh}{Mesh graph}
\tonrow{\meshnodes}{Nodes of mesh graph}
\tonrow{\meshedges}{Edges between the nodes of the mesh graph}
\tonrow{\gridnodes}{Grid nodes, corresponding to grid cells in the data}
\tonrow{\graph_\gtm}{Graph connecting grid nodes to the mesh}
\tonrow{\edgeset_\gtm}{Edges of $\graph_\gtm$, going from grid nodes to mesh nodes}
\tonrow{\graph_\mtg}{Graph connecting the mesh to grid nodes}
\tonrow{\edgeset_\mtg}{Edges of $\graph_\mtg$, going from mesh nodes to grid nodes}
\tonrow{\msgraph}{Multi-scale graph}
\tonrow{\hiintragraph{\leveli}}{Intra-level graph at level $\leveli$ in hierarchical mesh graph}
\tonrow{\himeshnodes{\leveli}}{Nodes at level $\leveli$ of hierarchical mesh graph}
\tonrow{\hiintraedges{\leveli}}{Edges of $\hiintragraph{\leveli}$, intra-level edges at level $\leveli$ in hierarchical mesh graph}
\tonrow{\hiintergraph{\leveli}{\leveli+1}}{Inter-level graph with edges from level $\leveli$ to level $\leveli+1$ in hierarchical mesh graph}
\tonrow{\hiinteredges{\leveli}{\leveli+1}}{Edges of $\hiintergraph{\leveli}{\leveli+1}$, from level $\leveli$ to level $\leveli+1$ in hierarchical mesh graph}
\tonrow{\neigh{\nodeii}}{Incoming neighbors of node $\nodeii$, $\set{\nodei: (\nodei,\nodeii) \in \edgeset{}}$}
\tonseparator

\tonrow{\areaweight_\nodei}{Area weighting for node $\nodei$, proportional to the area of corresponding grid cell}
\tonrow{\klweight}{Weighting for KL-term in variational training objective}
\tonrow{\crpsweight}{Weighting for \gls{CRPS} term in fine-tuning objective of \himodelprob}
\tonrow{\predsigma}{Standard deviation used in \gls{NLL} or variational training objective}
\bottomrule
\end{tabular}
\end{table}
\section{Societal Impact}
\label{sec:societal_impact}

\paragraph{Extreme weather}
Due to climate change the prevalence and severity of extreme weather events is expected to increase substantially, endangering both property and human life \citeapp{ipcc_2023}. 
These events are also getting harder to predict and the cost of damages per event has increased nearly 77\% over the past five decades \citeapp{world_economic_forum}.
In order to detect extreme events, there is a need for ensemble forecasting to better capture the full distribution of possible weather states.
While ensemble forecasting historically has been limited by computational costs \cite{quiet_revolution_nwp}, efficient \gls{NeurWP} ensemble models have the potential to vastly improve our ability to model extreme weather.
In \cref{sec:extreme_case_study} we present a case study showing how our \himodelprob could be used for forecasting Hurricane Laura.
More extensive evaluation of the abilities of \gls{NeurWP} models to capture extreme events is a complex, but important issue \citeapp{rise_of_neurwp,storm_ciaran_case_study}.
We view these capabilities as one of the main motivations for further developments of \gls{NeurWP} ensemble forecasting models.

\paragraph{Forecast failures}
There will always be cases where weather forecasting systems fail, and produce inaccurate predictions for future weather.
Depending on the weather event, such forecast errors can have disastrous consequences.
In these cases the lack of interpretability of black-box \gls{NeurWP} systems can be a problem, making it hard to understand why the model was wrong.
Traditional physics-based systems can also be tough to interpret due to their complexity, but at their core are physical equations understood by researchers.
With the rapid progress of \gls{NeurWP}, it is likely that we will soon see a landscape where the physical models take the back seat to a plethora of skillful \gls{NeurWP} forecasting systems.
In such a scenario the question of how to investigate forecast failures becomes pressing.
It seems desirable to be able to fall back on physical models for understanding impactful events poorly forecast by \gls{NeurWP}.
However, to allow this we can not do away with all infrastructure and expertise related to physical modeling.
These are important considerations as weather forecasting moves into an era of operational \gls{NeurWP}.

\paragraph{Renewable energy}
The production of renewable energy, such as solar and wind power, can be highly volatile \citeapp{re_integration}.
This creates challenges for including these sources in the larger energy system.
Accurate weather forecasts, translated into forecasts of energy production, fill an important role as enablers of these energy sources by making their output predictable.
Detailed probability estimates from ensemble forecasting can additionally allow for improved cost-loss decision making in these systems.

\paragraph{The energy footprint of weather forecasting }
Traditional weather forecasting systems utilize massive computing clusters \cite{fundamentals_of_nwp}, resulting in a substantial energy footprint of the forecasting process.
In comparison, \gls{NeurWP} models are highly energy efficient, even when taking into account their initial training process. 
The total energy required to train the large global \gls{NeurWP} model FourCastNet is comparable to running a 10 day forecast with 50 ensemble members using a traditional forecasting system \cite{fourcastnet}. 
Producing a forecast using the same model uses four orders of magnitude less energy than a physics-based model.
A similar reduction in energy footprint is to be expected for our models, if applied at the same scale.
For ensemble forecasting the total energy saving is even greater, as it is multiplied by the number of ensemble members.
However, one has to beware of rebound effects, where efficiency improvements result in more extensive use of resources.
If ensemble forecasting becomes $1000$ times more energy efficient and we run $1000$ times as many ensemble members there is no energy saved in the end.
\section{Model Details}
\label{sec:model_details}
We here give more details about the different models discussed in the main paper.

\subsection{Deterministic graph-based models \texorpdfstring{\cite{graphcast, keisler}}{}}
\label{sec:gc_details}
Deterministic graph-based \gls{NeurWP} models represent the single-step prediction function \hbox{$\pred^{\timei} = \arfunc(\wstate^{\timei - 2:\timei-1}, \forcing^\timei)$} as a sequence of \glspl{MLP} and \gls{GNN} layers.
\Cref{alg:graph_forecasting} describes the full prediction process.
Note that for some of the \gls{GNN} layers the edge representations are not updated, as these edges are not used in later steps of the process.
We denote this by $\noderep, \cdot \gets \gnn(\dots)$ in \cref{alg:graph_forecasting}.
\glspl{MLP} always act on the last dimension of its concatenated inputs.
Mesh node and edge representations are initialized by embedding related static features using additional \glspl{MLP}.
These static features contain information about node and edge positions (see \cref{sec:global_graph_construction,sec:lam_graph_construction}).
The number of processing steps to run on the mesh is a hyperparameter.
Each such step contains one \gls{GNN} layer, and these \glspl{GNN} do not share parameters.
The mapping $\arfunc$ can optionally output also standard deviations $\predsigma$, which we use when training with \gls{NLL} loss (see \cref{sec:training_objectives_det}).
We restrict $\predsigma$ to be positive by applying the Softplus function.

\begin{algorithm}[tbp]
\caption{Single-step prediction $\arfunc$ for graph-based \gls{NeurWP}}
\label{alg:graph_forecasting}
\begin{algorithmic}[1]
\State $\noderep^\grid \gets \mlp(\wstate^{\timei-2:\timei-1}, \forcing^\timei)$ \Comment Embedd grid inputs to $\latentdim$-dimensional vectors
\State $\noderep^\mesh, \cdot \gets \gnn(\graph_\gtm, \noderep^\grid, \edgerep^\gtm, \noderep^\mesh)$ \Comment Map grid representation to mesh
\ForAll{mesh processing steps}
    \State $\noderep^\mesh, \edgerep^\mtm \gets \gnn(\graph_\mesh, \noderep^\mesh, \edgerep^\mtm, \noderep^\mesh)$ \Comment Update mesh representations
\EndFor
\State $\noderep^\grid \gets \noderep^\grid + \mlp(\noderep^\grid)$ 
\State $\noderep^\grid, \cdot \gets \gnn(\graph_\mtg, \noderep^\mesh, \edgerep^\mtg, \noderep^\grid)$ \Comment Map mesh representation back to grid
\If{outputs $\predsigma$}
    \State $\left[\meanrep, \stdrep\right] \gets \mlp\left(\noderep^{\grid}\right)$
    \State \Return $\wstate^{\timei-1} + \meanrep, \softplus(\stdrep)$ \Comment Return prediction of $\wstate^{\timei}$ and $\predsigma$ for loss 
\Else
    \State \Return $\wstate^{\timei-1} + \mlp(\noderep^\grid)$ \Comment Return prediction of $\wstate^{\timei}$
\EndIf
\end{algorithmic}
\end{algorithm}

\subsection{\himodeldet}
\label{sec:hi_det_details}
In the hierarchical graph there are node representations associated with every level and edge representations associated with each subset of edges.
We let $\noderep^\leveli$ be the representations associated with nodes $\himeshnodes{\leveli}$ at level $\leveli$ in the mesh hierarchy.
Similarly, let $\edgerep^\leveli$ contain representations of intra-level edges $\himeshedges{\leveli}$, $\edgerep^{\leveli,\leveli+1}$ representations of upwards edges $\hiinteredges{\leveli}{\leveli + 1}$, and $\edgerep^{\leveli,\leveli-1}$ representations of downward edges $\hiinteredges{\leveli}{\leveli - 1}$.
As in the non-hierarchical case, all representations associated with the mesh graph are initialized by \glspl{MLP} applied to static features.
Independent \glspl{MLP} are used for the different levels of the hierarchy.

Forecasting using \himodeldet follows a similar structure as previous works, but replaces the mappings between the grid and mesh to encompass all levels and changes the processing steps into sweeps through the hierarchy.
Recall that for the hierarchical graph we re-define $\graph_\gtm = (\gridnodes \cup \himeshnodes{1}, \edgeset_\gtm)$ and $\graph_\mtg = (\gridnodes \cup \himeshnodes{1}, \edgeset_\mtg)$.
The full prediction mapping $\arfunc$ for \himodeldet is described in \cref{alg:det_hi_model}.
We define one processing step in \himodeldet as one update to all node and edge representations in the graph, keeping a similar interpretation as processing steps in non-hierarchical graphs.
This means that one sweep down and up (\crefrange{line:hi_processor_start}{line:hi_processor_end} in \cref{alg:det_hi_model}) is counted as two processing steps.
It follows that only even numbers of processing steps are reasonable in \himodeldet.
Note also here that \glspl{GNN} only share parameters across nodes and edges in the specific graph they operate on.
There is no parameter sharing across processing steps nor levels in the hierarchy.
We include Propagation Networks in \himodeldet, but not all parts of the model can be expected to benefit from this change.
For some steps of \cref{alg:det_hi_model} we want to keep the inductive bias of Interaction Networks to retain information.
Specifically, we use Propagation Networks for the mappings directly between the grid and lowest mesh level (\cref{line:hi_g2m,line:hi_m2g}) and for the upward processing steps (\cref{line:hi_up_gnn}).
\begin{algorithm}[tbp]
\caption{Single-step prediction $\arfunc$ in \himodeldet}
\label{alg:det_hi_model}
\begin{algorithmic}[1]
\State $\noderep^\grid \gets \mlp(\wstate^{\timei-2:\timei-1}, \forcing^\timei)$ \Comment Embedd grid inputs to $\latentdim$-dimensional vectors
\State $\noderep^1, \cdot \gets \gnn(\graph_\gtm, \noderep^\grid, \edgerep^\gtm, \noderep^1)$ \Comment Map from the grid \dots
\label{line:hi_g2m}
\For{$\leveli = 2, \dots, \himaxl$} \Comment \dots and up through the whole mesh hierarchy
    \State $\noderep^{\leveli}, \edgerep^{\leveli-1, \leveli} \gets \gnn(\hiintergraph{\leveli-1}{\leveli}, \noderep^{\leveli-1}, \edgerep^{\leveli-1, \leveli}, \noderep^{\leveli})$
\EndFor
\For{number of mesh processing steps / 2} \Comment Each iteration contains two steps, down and up
    \State $\noderep^{\himaxl}, \edgerep^{\himaxl} \gets \gnn(\hiintragraph{\himaxl}, \noderep^{\himaxl}, \edgerep^{\himaxl}, \noderep^{\himaxl})$
    \label{line:hi_processor_start}
    \For{$\leveli = (\himaxl-1), \dots, 1$} \Comment Downward pass of sweep
        \State $\noderep^{\leveli}, \edgerep^{\leveli+1, \leveli} \gets \gnn(\hiintergraph{\leveli+1}{\leveli}, \noderep^{\leveli+1}, \edgerep^{\leveli+1, \leveli}, \noderep^{\leveli})$
        \State $\noderep^{\leveli}, \edgerep^{\leveli} \gets \gnn(\hiintragraph{\leveli}, \noderep^{\leveli}, \edgerep^{\leveli}, \noderep^{\leveli})$
    \EndFor
    \State $\noderep^{1}, \edgerep^{1} \gets \gnn(\hiintragraph{1}, \noderep^{1}, \edgerep^{1}, \noderep^{1})$
    \For{$\leveli = 2, \dots, \himaxl$} \Comment Upward pass of sweep
        \State $\noderep^{\leveli}, \edgerep^{\leveli-1, \leveli} \gets \gnn(\hiintergraph{\leveli-1}{\leveli}, \noderep^{\leveli-1}, \edgerep^{\leveli-1, \leveli}, \noderep^{\leveli})$
        \label{line:hi_up_gnn}
        \State $\noderep^{\leveli}, \edgerep^{\leveli} \gets \gnn(\hiintragraph{\leveli}, \noderep^{\leveli}, \edgerep^{\leveli}, \noderep^{\leveli})$
    \EndFor
    \label{line:hi_processor_end}
\EndFor
\For{$\leveli = (\himaxl-1), \dots, 1$} 
    \State $\noderep^{\leveli}, \cdot \gets \gnn(\hiintergraph{\leveli+1}{\leveli}, \noderep^{\leveli+1}, \edgerep^{\leveli+1, \leveli}, \noderep^{\leveli})$ \Comment Map down through the mesh hierarchy \dots
\EndFor
\State $\noderep^\grid \gets \noderep^\grid + \mlp(\noderep^\grid)$ 
\State $\noderep^\grid, \cdot \gets \gnn(\graph_\mtg, \noderep^1, \edgerep^\mtg, \noderep^\grid)$ \Comment \dots and finally to the grid
\label{line:hi_m2g}
\If{outputs $\predsigma$}
    \State $\left[\meanrep, \stdrep\right] \gets \mlp\left(\noderep^{\grid}\right)$
    \State \Return $\wstate^{\timei-1} + \meanrep, \softplus(\stdrep)$ \Comment Return prediction of $\wstate^{\timei}$ and $\predsigma$ for loss 
\Else
    \State \Return $\wstate^{\timei-1} + \mlp(\noderep^\grid)$ \Comment Return prediction of $\wstate^{\timei}$
\EndIf
\end{algorithmic}
\end{algorithm}

\subsection{\himodelprob}
\label{sec:hi_prob_details}
\begin{figure}[t]
    \centering
    \includegraphics[width=\textwidth]{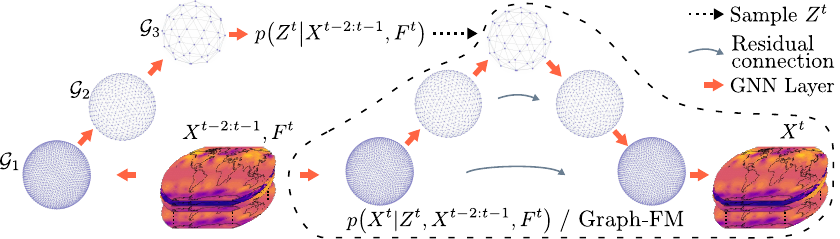}
    \caption{
    Overview of our \himodelprob model, with example data and graphs for global forecasting. 
    The corresponding overview for the \gls{LAM} setting is given in \cref{fig:model_overview_lam}.
    Note that the predictor part of the model has the same structure as one sweep through the hierarchy in \himodeldet.
    }
    \label{fig:model_overview_global}
\end{figure}

An overview of the \himodelprob model with global graphs is shown in \cref{fig:model_overview_global}.
At each tim step \himodelprob takes as input $\arcond$ and produces a sample of $\wstate^\timei$.
By feeding the sample from the model back at the next time step, an ensemble member can be rolled out.
This process is described in \cref{alg:ar_sampling}, where we let $\pred^\timei$ be an initial state for $\timei < 1$ and then assign it sampled predictions for the actual forecast time steps $1, \dots, \forecastlength$.
\begin{algorithm}[tbp]
\caption{Sampling of ensemble member from  $\modeldist$}
\label{alg:ar_sampling}
\begin{algorithmic}[1]
    \State $\pred^{-1:0} \gets \initstates$
    \For{$\timei = 1, \dots, \forecastlength$}
        \State $\latent^\timei \sim p\left(\latent^\timei \middle| \pred^{t-2:t-1}, \forcing^\timei \right)$ \Comment Sample $\latent^\timei$ from latent map
        \State $\pred^\timei \gets \pred^{\timei-1} + \predictorfunc\left(\latent^\timei, \pred^{t-2:t-1}, \forcing^\timei \right)$ \Comment Compute next state using predictor
    \EndFor
    \State \Return $\pred^{1:\forecastlength}$
\end{algorithmic}
\end{algorithm}

When creating an ensemble of size $\enssize$ from deterministic initial conditions some care has to be taken as to how many members to sample.
If we at each time $\timei$ would sample $\enssize$ values of $\wstate^\timei$ for each realization of $\wstate^{\timei-2:\timei-1}$ we would end up with $\enssize^\forecastlength$ members at time $\forecastlength$.
Instead, we sample $\enssize$ realizations of $\wstate^{1}$ given the initial conditions and after that only one realization of $\wstate^{\timei}$ for each of the members.
This is equivalent to doing $\enssize$ independent runs of \cref{alg:ar_sampling}.
Note that since we can always draw new samples of $\latent^\timei$ we are never restricted in the ensemble size.

\paragraph{Latent map}
The latent map is defined as 
\eq[latent_map_def_extra]{
    \priordist = \prod_{\nodei \in \himeshnodes{\himaxl}}
     \normalpdf{\latent^t_\alpha}{\meanfunc{\latent}\left(\arcond \right)_\nodei}{I}
}
where the mean function $\meanfunc{\latent}$ is realized as a sequence of \gls{GNN} layers mapping up through the hierarchy.
The exact process is described in \cref{alg:latent_map}.
Since only one upward pass is performed we do not update any of the edge representations.
The final \gls{MLP} at \cref{line:latent_map_mean_mlp} maps from node representations at mesh level $\himaxl$ to the mean of $\priordist$.
In the latent map we use exclusively Propagation Networks, encouraging the model to encode useful information from the grid into $\latent^\timei$.
\begin{algorithm}[tbp]
\caption{Latent map mean function $\meanfunc{\latent}$}
\label{alg:latent_map}
\begin{algorithmic}[1]
\State $\noderep^\grid \gets \mlp(\wstate^{\timei-2:\timei-1}, \forcing^\timei)$ \Comment Embedd grid inputs to $\latentdim$-dimensional vectors
\label{line:latent_map_grid_emb}
\State $\noderep^1, \cdot \gets \gnn(\graph_\gtm, \noderep^\grid, \edgerep^\gtm, \noderep^1)$
\State $\noderep^{1}, \cdot \gets \gnn(\hiintragraph{1}, \noderep^{1}, \edgerep^{1}, \noderep^{1})$
\For{$\leveli = 2, \dots, \himaxl$} \Comment Propagate up the hierarchy
    \State $\noderep^{\leveli}, \cdot \gets \gnn(\hiintergraph{\leveli-1}{\leveli}, \noderep^{\leveli-1}, \edgerep^{\leveli-1, \leveli}, \noderep^{\leveli})$
    \State $\noderep^{\leveli}, \cdot \gets \gnn(\hiintragraph{\leveli}, \noderep^{\leveli}, \edgerep^{\leveli}, \noderep^{\leveli})$
\EndFor
\State \Return $\mlp\left(\noderep^{\himaxl}\right)$ \Comment Return mean of $\priordist$
\label{line:latent_map_mean_mlp}
\end{algorithmic}
\end{algorithm}

\paragraph{Predictor}
The predictor $\preddist$ is chosen as a dirac measure, with all probability mass concentrated in one point.
It thus takes the form of a deterministic mapping
\eq[predictor_extra]{
    \predictoreq.
}
This choice emphasizes that all randomness in $\ardist$ should come from the latent variable $\latent^\timei$.
An alternative would be to parametrize the predictor as a diagonal Gaussian.
We found this inferior in practice, as that would require also sampling from this Gaussian when rolling out forecasts.
This sampling would simply entail adding independent Gaussian noise to the forecasts, which only degrades their quality.
We still re-introduce the Gaussian form in the variational objective (\cref{eq:elbo_cvae}), as some choice of likelihood is necessary for a well-defined learning problem.
The interpretation of this is that the forecast is rolled out as a latent process with all randomness coming from $\latent^\timei$.
Observation noise is then assumed added to this forecast independently at each time step.
We use the latent, noise-free forecast as our prediction.
Connecting back to the \gls{VAE} analogue of our model, this setup corresponds to the common practice in the \gls{VAE} literature of using a Gaussian likelihood but treating the mean as the generated sample \cite{effective_vae_training}.

The exact form of the predictor function is described in \cref{alg:predictor}.
It has a similar form as \himodeldet, doing sweeps up and down the mesh hierarchy.
We found it sufficient to perform one sweep up and down in the predictor, but it can easily be generalized to multiple sweeps as in \himodeldet.
The predictor function optionally returns also the $\predsigma$ to be used in our variational objective.
In the predictor we use Interaction Networks in the upward direction and Propagation Networks in the downward direction.
This is to guarantee that randomness and useful information encoded in $\latent^\timei$ at the top level reaches all the way down to the grid.
Note that on \cref{line:latent_combine} $\latent^\timei$ is combined with $\noderep^{\himaxl-1}$ through the Interaction Network layer mapping from level $\himaxl-1$ to $\himaxl$.
For the details of how this combination happens, see \cref{eq:interaction_net}, where $\latent^\timei$ takes the role of the receiver node representations $\noderep^\receiver$.
Note that edge and node representations updated on \cref{line:predictor_intra_reps} in the upward sweep then re-appear in the downward pass.
This constitutes residual connections between the upward and downward passes, as illustrated in \cref{fig:model_overview_global}.
\begin{algorithm}[tbp]
\caption{Predictor function $\predictorfunc$}
\label{alg:predictor}
\begin{algorithmic}[1]
\State $\noderep^\grid \gets \mlp(\wstate^{\timei-2:\timei-1}, \forcing^\timei)$ \Comment Embedd grid inputs to $\latentdim$-dimensional vectors
\label{line:predictor_grid_emb}
\State $\noderep^1, \cdot \gets \gnn(\graph_\gtm, \noderep^\grid, \edgerep^\gtm, \noderep^1)$
\For{$\leveli = 1, \dots, \himaxl-1$} \Comment Upward pass of sweep
    \State $\noderep^{\leveli}, \edgerep^{\leveli} \gets \gnn(\hiintragraph{\leveli}, \noderep^{\leveli}, \edgerep^{\leveli}, \noderep^{\leveli})$
    \label{line:predictor_intra_reps}
    \If{$\leveli = \himaxl-1$}
        \State $\noderep^{\himaxl}, \cdot \gets \gnn(\hiintergraph{\himaxl-1}{\himaxl}, \noderep^{\himaxl-1}, \edgerep^{\himaxl-1, \himaxl}, \latent^\timei)$ \Comment Incorporate $\latent^\timei$ through \gls{GNN}
        \label{line:latent_combine}
    \Else
        \State $\noderep^{\leveli+1}, \cdot \gets \gnn(\hiintergraph{\leveli}{\leveli+1}, \noderep^{\leveli}, \edgerep^{\leveli, \leveli+1}, \noderep^{\leveli+1})$
    \EndIf
\EndFor
\State $\noderep^{\himaxl}, \cdot \gets \gnn(\hiintragraph{\himaxl}, \noderep^{\himaxl}, \edgerep^{\himaxl}, \noderep^{\himaxl})$
\For{$\leveli = (\himaxl-1), \dots, 1$} \Comment Downward pass of sweep
    \State $\noderep^{\leveli}, \cdot \gets \gnn(\hiintergraph{\leveli+1}{\leveli}, \noderep^{\leveli+1}, \edgerep^{\leveli+1, \leveli}, \noderep^{\leveli})$
    \State $\noderep^{\leveli}, \cdot \gets \gnn(\hiintragraph{\leveli}, \noderep^{\leveli}, \edgerep^{\leveli}, \noderep^{\leveli})$
\EndFor
\State $\noderep^\grid \gets \noderep^\grid + \mlp(\noderep^\grid)$ 
\State $\noderep^\grid, \cdot \gets \gnn(\graph_\mtg, \noderep^1, \edgerep^\mtg, \noderep^\grid)$
\If{outputs $\predsigma$}
    \State $\left[\meanrep, \stdrep\right] \gets \mlp\left(\noderep^{\grid}\right)$
    \State \Return $\wstate^{\timei-1} + \meanrep, \softplus(\stdrep)$ \Comment Return prediction of $\wstate^{\timei}$ and $\predsigma$ for loss 
\Else
    \State \Return $\wstate^{\timei-1} + \mlp(\noderep^\grid)$ \Comment Return prediction of $\wstate^{\timei}$
\EndIf
\end{algorithmic}
\end{algorithm}

\paragraph{Variational approximation}
Defining the \gls{ELBO} for the model requires a variational approximation $\vardist$ at each time step, approximating the true posterior $\postdist$ over $\latent^\timei$.
Due to the autoregressive structure of the model, it is sufficient to condition $\vardistname$ on states at time steps $\timei-2$, $\timei-1$ and $\timei$.
This conditioning removes all dependence between $\latent^t$ and other variables (see \cref{fig:graphical_model}), simplifying the parametrization and training of $\vardistname$.
We choose the variational approximation to have a similar Gaussian form as the latent map
\als[var_dist]{
    \vardist& =
    \\
    \prod_{\nodei \in \himeshnodes{\himaxl}}
    \prod_{\vari = 1}^{\latentdim}
    &\normalpdf{
        \latent^t_{\alpha, \vari}
    }{
        \meanfunc{\vardistname}\left(\wstate^{\timei-2:\timei-1}, \wstate^{\timei}, \forcing^\timei \right)_{\nodei, \vari}
    }{
        \stdfunc{\vardistname}\left(\wstate^{\timei-2:\timei-1}, \wstate^{\timei}, \forcing^\timei \right)_{\nodei, \vari}^2
    }.
}
For added flexibility we also model the variance of $\vardistname$.
Here both $\meanfunc{\vardistname}$ and $\stdfunc{\vardistname}$ are functions mapping from the grid inputs to $\setsize{\himeshnodes{\himaxl}} \times \latentdim$ matrices.
These functions are implemented jointly as described in \cref{alg:var_approx}.
This mapping is similar to the latent map, with the differences being that $\wstate^\timei$ is taken as an input and both mean and standard deviation returned.
Similarly to the latent map, all \glspl{GNN} used in the variational approximation are Propagation Networks.
\begin{algorithm}[tbp]
\caption{Parameter mappings $\meanfunc{\vardistname}, \stdfunc{\vardistname}$ of variational approximation}
\label{alg:var_approx}
\begin{algorithmic}[1]
\State $\noderep^\grid \gets \mlp(\wstate^{\timei-2:\timei-1}, \wstate^\timei, \forcing^\timei)$ \Comment Embedd grid inputs (including $\wstate^\timei$) to $\latentdim$-dim. vectors
\State $\noderep^1, \cdot \gets \gnn(\graph_\gtm, \noderep^\grid, \edgerep^\gtm, \noderep^1)$
\State $\noderep^{1}, \cdot \gets \gnn(\hiintragraph{1}, \noderep^{1}, \edgerep^{1}, \noderep^{1})$
\For{$\leveli = 2, \dots, \himaxl$} \Comment Propagate up the hierarchy
    \State $\noderep^{\leveli}, \cdot \gets \gnn(\hiintergraph{\leveli-1}{\leveli}, \noderep^{\leveli-1}, \edgerep^{\leveli-1, \leveli}, \noderep^{\leveli})$
    \State $\noderep^{\leveli}, \cdot \gets \gnn(\hiintragraph{\leveli}, \noderep^{\leveli}, \edgerep^{\leveli}, \noderep^{\leveli})$
\EndFor
\State $\left[\meanrep, \stdrep\right] \gets \mlp\left(\noderep^{\himaxl}\right)$
\State \Return $\meanrep, \softplus(\stdrep)$ \Comment Return mean and standard deviation of $\vardist$
\end{algorithmic}
\end{algorithm}

\subsection{\texorpdfstring{\gls{MLP}}{MLP} parametrization}
Following \citet{graphcast}, all \glspl{MLP} have one hidden layer with Swish activation functions \citeapp{swish}.
With the exception of \glspl{MLP} outputting predictions or distribution parameters, we apply LayerNorm \citeapp{layernorm} to all other \gls{MLP} outputs.

While \glspl{MLP} in different \gls{GNN} layers never share parameters, we use some parameter sharing across the embedding \glspl{MLP} in the different components of \himodelprob.
Specifically, the latent map and predictor share parameters for the \gls{MLP} embedding the grid input (\cref{line:latent_map_grid_emb} in \cref{alg:latent_map} and \cref{line:predictor_grid_emb} in \cref{alg:predictor}).
The \glspl{MLP} that embedd static graph features, for initializing graph representation vectors, are also shared across all components of the \himodelprob model.
    
\subsection{Baseline models}
\label{sec:baseline_model_details}
\paragraph{\gcour}
\gcour is a reimplementation of the \gc model \cite{graphcast} in our codebase.
This was done to allow for a fair comparison by using the same data and multi-scale graphs as other models in our experiments.

\paragraph{\msmodelprob}
The \msmodelprob model is a version of \himodelprob, but using a multi-scale mesh graph $\msgraph$ \cite{graphcast}, instead of the hierarchical one.
We replace the sweeps up and down the hierarchy with multiple processing steps performed on the same mesh graph, similar to the deterministic case described in \cref{alg:graph_forecasting}.
In the predictor of this model, the latent $\latent^\timei$ is integrated already in the mapping from the grid to the mesh as 
\eq[msmodelprob_latent_combine]{
\noderep^\mesh, \cdot \gets \gnn(\graph_\gtm, \noderep^\grid, \edgerep^\gtm, \latent^\timei).
}
Note that as $\latent^\timei$ here is associated with the nodes of $\msgraph$, it has shape $\setsize{\nodeset_1} \times \latentdim$ as opposed to $\setsize{\himeshnodes{\himaxl}} \times \latentdim$ for \himodelprob.
As $\nodeset_1$ has many more nodes, the total dimensionality of $\latent^\timei$ is higher for \msmodelprob.
This is a consequence of the multi-scale graph offering no further dimensionality reduction beyond mapping from the grid to the mesh.

\paragraph{\gcswa}
One simple way to create an \gls{NeurWP} ensemble is by re-training multiple deterministic models from random initializations.
Training many large weather models from scratch is however infeasible in practice.
It also limits the number of ensemble members to the number of models trained.
Inspired by \citet{calibration_of_large_neurwp}, we create a multi-model ensemble baseline using \gls{SWAG} \cite{swag}.
\gls{SWAG} is a technique for approximate Bayesian inference based on running \gls{SGD} training with a constant high learning rate.
At regular intervals throughout this optimization process the model parameters are saved.
The full \gls{SWAG} process includes estimating a high-dimensional Gaussian over these parameters and drawing samples to create the ensemble.
This gets around the limitation of needing to decide the maximum ensemble size when training the model.
As we never require more ensemble members than saved during the \gls{SGD} training we skip estimating this Gaussian and resampling.
Instead, we directly use a subset of the saved model parameters to create our ensemble members.
\citet{calibration_of_large_neurwp} combine \gls{SWAG} with initial state perturbations, but as we model the conditional distribution $\modeldist$ we do not include any such perturbations to the initial states.

To create our \gcswa ensemble we start from the trained \gcour model and run \gls{SGD} training for $\forecastlength = 8$ unrolled steps and with a fixed learning rate ($10^{-3}$ for the global experiment and $5\times10^{-5}$ for \gls{MEPS}).
Even higher learning rates led to numerical issues or the model training diverging.
Model parameters are then saved every 100 gradient descent steps.
During evaluation these parameters are loaded one after another and used to produce the forecast for each ensemble member.
The error and spread of the \gcswa ensemble is highly dependent on the correlation of consecutively saved model parameters.
Given the poor spread of \gcswa we tested also saving parameters with 1000 steps in-between, but this led to similar results.

\section{Training Objectives}
\label{sec:trainin_objectives_extra}
We here give detailed definitions of the objectives used to train our models.

\subsection{Deterministic models}
\label{sec:training_objectives_det}
\paragraph{Weighted \gls{MSE}}
Deterministic forecasting models can be trained by minimizing a weighted \gls{MSE} \cite{graphcast} of rolled-out forecasts
\eq[wmse_loss]{
    \wmseloss = 
    \frac{1}{\forecastlength\ngridpoints}
    \sum_{\timei = 1}^{\forecastlength}
    \sum_{\nodei \in \gridnodes}
    \sum_{\vari = 1}^{\statedim}
    \areaweight_\nodei \stateweight_\vari \invdiffweight_\vari
    \left(
        \pred^{\timei}_{\nodei, \vari} - \wstate^{\timei}_{\nodei, \vari}
    \right)^2
}
where we recall that $\ngridpoints = \setsize{\gridnodes}$ and:
\begin{itemize}
    \item $\areaweight_\nodei$ is the normalized area weighting for grid cell $\nodei$. These are computed as described by \citet{weatherbench2}.
    \item $\invdiffweight_\vari$ is the inverse variance of time differences for variable $\vari$. These can be computed directly from iterating over the data.
    \item $\stateweight_\vari$ is a weight associated with the vertical level of variable $\vari$. For global experiments we use the same weights $\stateweight_\vari$ as \citet{graphcast}.
\end{itemize}

\paragraph{\acrlong{NLL} loss}
For the \gls{MEPS} data however, choosing weights $\stateweight_\vari$ is challenging due to the many different variables and their irregular vertical levels.
In such cases, an alternative approach is to use an \gls{NLL} loss, where the model itself determines the trade-offs between different variables \cite{fengwu}.
As the \gls{MEPS} area has grid cells of approximately equal size we can also simply set $\areaweight_\nodei = 1$.
For the \gls{MEPS} experiments we thus use
\eq[nll_loss]{
    \nllloss = 
    -\frac{1}{\forecastlength\ngridpoints}
    \sum_{\timei = 1}^{\forecastlength}
    \sum_{\nodei \in \gridnodes}
    \sum_{\vari = 1}^{\statedim}
    \areaweight_\nodei 
    \log \normalpdf{\wstate^{\timei}_{\nodei, \vari}}{\pred^{\timei}_{\nodei, \vari}}{\predsigma^2_{\nodei, \vari}}.
}
The standard deviation $\predsigma_{\nodei, \vari}$ is here a second output from the mapping $\arfunc$.
When unrolling forecasts we do not sample from this Gaussian, but feed the mean $\pred^\timei$ to the next time steps.
Note that \cref{eq:nll_loss} decomposes into weighted squared error terms and log-determinant terms preventing too large $\predsigma$ outputs.
The weighted \gls{MSE} in \cref{eq:wmse_loss} can therefore also be viewed as a special case of the \gls{NLL} loss, up to an additive constant.
This relationship corresponds to \cref{eq:nll_loss} with constant variances given by 
\eq[std_weighting_equivalence]{
\predsigma^2_{\nodei, \vari} = \frac{1}{2\stateweight_\vari \invdiffweight_\vari}.
}

\subsection{Probabilistic model}
For the probabilistic model we can not directly apply the training objectives above, as they only seek to match the first moments of the model distribution to the data.
To train the probabilistic model we instead leverage the fact that the single-step model has a structure similar to a (conditional) \gls{VAE} \cite{vae, cvae}.
We can therefore optimize a variational objective, the \gls{ELBO}, to match the distribution of the data.

\paragraph{Variational objective}
For a single time step, our variational objective is
\als[elbo_cvae_extra]{
    \elbo&\left(\wstate^{\timei-2:\timei-1}, \wstate^\timei, \forcing^{\timei} \right)
    =
    \klweight \kl{\vardist }{\priordist}
    \\
    -&\E{\vardist}{
        {\textstyle \sum_{\nodei \in \gridnodes}
        \sum_{\vari = 1}^{\statedim}}
        \areaweight_\nodei
        \log \normalpdf{
            \wstate^{\timei}_{\nodei, \vari}
        }{
            \predictorfunc\left(\latent^\timei, \arcond\right)_{\nodei, \vari}
        }{
            \predsigma^2_{\nodei, \vari}
        }
    }.
}
We include a weight $\klweight$ in front of the KL-regularizer to allow for some tuning between the two parts of the objective function \citepapp{how_to_train_vae}.
In practice we found this useful to make sure the model does not collapse to pure deterministic predictions.
Note that when $\klweight = 1$ this is exactly equal to the (negative) \gls{ELBO}.
The \gls{NLL} from \cref{eq:nll_loss} here shows up again in the second term.
Similarly to the deterministic models, $\predsigma_{\nodei, \vari}$ can either be output by the model (the predictor in this case) or chosen as a constant.
For global experiments we choose $\predsigma_{\nodei, \vari}$ as in \cref{eq:std_weighting_equivalence}, reducing the term to a weighted \gls{MSE} loss.
For the \gls{MEPS} data we again let each $\predsigma_{\nodei, \vari}$ be output by the model.
\Cref{eq:elbo_cvae_extra} can in practice be computed efficiently by using a single-sample Monte Carlo estimate for the expectation and the reparametrization trick \cite{vae}.
The KL-divergence is available in closed form as both distributions are Gaussian.
As with deterministic models \cite{graphcast, keisler, fengwu, stormer}, it is crucial to fine-tune on rolled out forecasts of multiple time steps. 
This induces stability and improves performance for longer lead times.
We thus define our fine-tuning objective as 
\eq[multi_step_loss]{
    \elboloss = {\textstyle \sum_{\timei=1}^{\forecastlength}}
    \elbo\left(\pred^{\timei-2:\timei-1}, \wstate^{\timei}, \forcing^{\timei} \right)
}
where $\pred^\timei$ is an initial state from the dataset for $\timei < 1$ and otherwise sampled using the variatonal approximation for $\latent^\timei$.
In \cref{eq:multi_step_loss} $\wstate^\timei$ is always from the training data.

\paragraph{\gls{CRPS} fine-tuning}
\label{sec:crps_fine_tuning}
In practice we found that fine-tuning the model with only $\elboloss$ tends to result in underdispersed ensemble forecasts, underestimating the variance of the distribution.
To remedy this, we include an additional fine-tuning objective based on the \gls{CRPS} \citeboth{proper_scoring_rules,crps_estimator,neural_gcm}.
The \gls{CRPS} measures how well a univariate distribution matches the observed data and is defined as
\eq[crps_def]{
    \crps\left(p(\firstrv), \observation\right) = \E{}{|\firstrv - \observation|} - \frac{1}{2} \E{}{|\firstrv - \secondrv|}
}
where $\firstrv, \secondrv$ are independent copies of a random variables with distribution $p(\firstrv)$ and $\observation$ the observed data.
The \gls{CRPS} is minimized only when the predicted distribution matches that of the data \cite{proper_scoring_rules}, making it suitable for calibrating our ensemble forecasting model.
We define our \gls{CRPS} fine-tuning loss as an unbiased two-sample estimator, summed over all dimensions of the predictive distribution
\eq[crps_loss]{
    \crpsloss = 
    \sum_{\timei = 1}^{\forecastlength}
    \sum_{\nodei \in \gridnodes}
    \sum_{\vari = 1}^{\statedim}
    \areaweight_\nodei
    \frac{1}{2} 
    \left(
        \left|\pred^{\timei}_{\nodei, \vari} - \wstate^{\timei}_{\nodei, \vari}\right|
        + \left|\predprime^{\timei}_{\nodei, \vari} - \wstate^{\timei}_{\nodei, \vari}\right|
        - \left|\pred^{\timei}_{\nodei, \vari} - \predprime^{\timei}_{\nodei, \vari}\right|
    \right)
}
with $\pred^\timei$ and $\predprime^\timei$ coming from two independent ensemble members sampled from the model.

Note that the \gls{CRPS} objective targets only the marginal distributions, while spatio-temporally coherent samples requires matching the joint distribution of the data.
To avoid degenerate solutions we weight $\crpsloss$ by $\crpsweight$ and combine it with the variational objective for the final fine-tuning loss $\loss = \elboloss + \crpsweight \crpsloss$.
Computing $\loss$ requires rolling out three forecasts,
one for $\elboloss$ (using samples from $\vardistname$) and two for $\crpsloss$ (using samples from the latent map).
This can have a substantial memory cost. 
It is however sufficient to include the \gls{CRPS} term only in the final steps of the training process, making this less of an issue in practice.

Empirically we found that including the \gls{CRPS} objective improves both ensemble calibration and forecast accuracy.
The accuracy improvement can be attributed to the \gls{MAE} terms in \cref{eq:crps_loss}.
Note that in $\elboloss$ forecasts are rolled out by sampling $\latent^\timei$ from the variational approximation, while in $\crpsloss$ these samples are from the latent map.
This matches how forecasting is carried out at test time.
As a consequence the \gls{CRPS} loss has an additional effect of bridging the gap between the variational training and the final forecasting.
\section{Metric Definitions}
\label{sec:metric_defs}
We here give full definitions for the metrics used to evaluate our models.

\subsection{Deterministic Metrics}
We define evaluation metrics based on set of $\samplesize$ forecasts, each a sequence $\pred^{1, (\samplei)}, \dots, \pred^{\forecastlength, (\samplei)}$ of predicted weather states.
Metrics are computed per lead-time and variable.
For deterministic forecasting we define the \gls{RMSE} of variable $\vari$ at lead time $\timei$ as 
\eq[rmse_def]{
    \rmse_{\timei, \vari} = \sqrt{
        \frac{1}{\samplesize \ngridpoints} \sum_{\samplei = 1}^{\samplesize}
        \sum_{\nodei \in \gridnodes}
        \areaweight_\nodei
        \left(\pred^{\timei, (\samplei)}_{\nodei, \vari} - \wstate^{\timei, (\samplei)}_{\nodei, \vari}\right)^2
    }.
}
In all metrics we include area weighting $\areaweight_\nodei$ to handle the fact that the grid cells for the global data has different area.
These weights are computed as described in \citet{weatherbench2}.
For the \gls{MEPS} data these are all set to $\areaweight_\nodei = 1$.
Note that the square root is applied after sample averaging, following standard convention and the WeatherBench 2 benchmark~\cite{weatherbench2}.

\subsection{Probabilistic Metrics}
In the probabilistic setting we evaluate an ensemble forecast of $\enssize$ members.
Let $\pred^{\timei, (\samplei), (\ensi)}$ be the prediction of ensemble member $k$.
For ensemble forecasts we compute the \gls{RMSE} of the ensemble mean as
\al[rmse_ens_def]{
    \rmse_{\timei, \vari} &= \sqrt{
        \frac{1}{\samplesize \ngridpoints} \sum_{\samplei = 1}^{\samplesize}
        \sum_{\nodei \in \gridnodes}
        \areaweight_\nodei
        \left(\ensmean^{\timei, (\samplei)}_{\nodei, \vari} - \wstate^{\timei, (\samplei)}_{\nodei, \vari}\right)^2
    }\\
    \ensmean^{\timei, (\samplei)} &= \frac{1}{\enssize} \sum_{\ensi = 1}^{\enssize} \pred^{\timei, (\samplei), (\ensi)}.
}
To measure the calibration of uncertainty expressed by the ensemble we use the (bias-corrected) \spskrtext
\alse[spskr_def]{
    \spskr_{\timei, \vari} &= \sqrt{\frac{\enssize + 1}{\enssize}} \frac{\spread_{\timei, \vari}}{\rmse_{\timei, \vari}}
    \\
    \spread_{\timei, \vari} &= \sqrt{
        \frac{1}{\samplesize\enssize\ngridpoints} \sum_{\samplei = 1}^{\samplesize}
        \sum_{\ensi = 1}^{\enssize}
        \sum_{\nodei \in \gridnodes}
        \areaweight_\nodei
        \left(\ensmean^{\timei, (\samplei)}_{\nodei, \vari} - \pred^{\timei, (\samplei), (\ensi)}_{\nodei, \vari}\right)^2
    }.
}
If the ensemble members represent realistic forecasts, and observe exchangeability with the ground truth, $\spskr_{\timei, \vari} \approx 1$ \citep{why_spskr}. 

As a probabilistic metric we also use the \gls{CRPS} \citep{proper_scoring_rules}, here computed as a finite sample estimate \citeapp{crps_estimator} over all the ensemble members 
\als[crps_metric_def]{
    \crps_{\timei, \vari} = 
        \frac{1}{\samplesize\ngridpoints} \sum_{\samplei = 1}^{\samplesize}
        \sum_{\nodei \in \gridnodes}
        \areaweight_\nodei
        \bigg(&
        \frac{1}{\enssize} \sum_{\ensi = 1}^{\enssize}
            \left|\pred^{\timei, (\samplei), (\ensi)}_{\nodei, \vari} - \wstate^{\timei, (\samplei)}_{\nodei, \vari}\right|
        \\
        &-
        \frac{1}{2 \enssize (\enssize - 1)}
        \sum_{\ensi = 1}^{\enssize}
        \sum_{\ensi' = 1}^{\enssize}
            \left|\pred^{\timei, (\samplei), (\ensi)}_{\nodei, \vari} - \pred^{\timei, (\samplei), (\ensi')}_{\nodei, \vari}\right|
        \bigg).
}
Note that this is a spatial average over marginal \gls{CRPS} values and does not take any covariance structure of the data or model distributions into account.
To avoid the quadratic sum in \cref{eq:crps_metric_def} we in practice use the idea of \citetapp{crps_estimator} to compute this with linear memory by sorting the ensemble members.
For deterministic models the \gls{CRPS} reduces to \gls{MAE}. 
See this for example from \cref{eq:crps_metric_def} by letting all ensemble members be the same predicted value.
\section{Extreme Weather Case Study: Hurricane Laura}
\label{sec:extreme_case_study}
One key motivations for probabilistic weather forecasting is to allow for better predictions of extreme events and improved estimates of uncertainties associated with these.
To exemplify this we include here a case study looking at forecasts for Hurricane Laura, using the global models. 

\begin{figure}
    \centering
    \includegraphics[width=\linewidth]{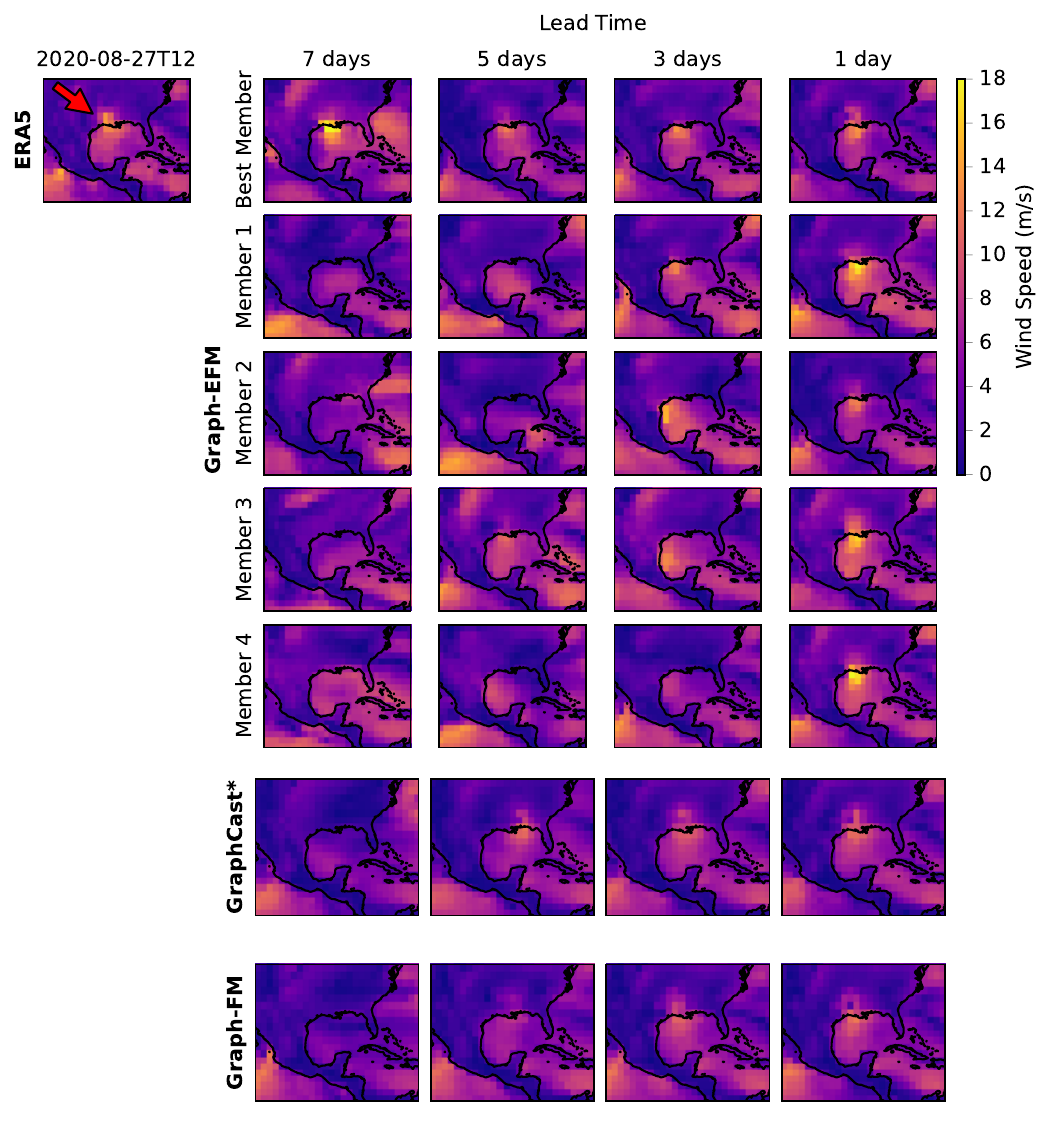}
    \caption{
        Example forecasts of 10 m wind speeds during Hurricane Laura, all for \texttt{2020-08-27T12} UTC.
        The first column shows the wind speeds in ERA5, with the red arrow indicating the location of the hurricane.
        Remaining columns show model forecasts for the specific time, initialized at different lead times ahead.
        For \himodelprob we plot both 4 randomly sampled ensemble members and one cherry-picked member that was deemed to best match ERA5.
    }
    \label{fig:laura_plot}
\end{figure}

In August of 2020 Laura developed in the Atlantic Ocean and reached hurricane levels in the Gulf of Mexico, eventually making landfall in Louisiana and causing major damages \citeapp{hurricane_laura}. 
We study here forecasts for \texttt{2020-08-27T12} UTC, which is 6 hours after the hurricane hit land. 
All forecasts are for this exact time point, but initialized a varying number of days before. 
We here run 50 member ensemble forecasts using \himodelprob and deterministic forecasts using the \gcour and \hbox{\himodeldet} models.
\Cref{fig:laura_plot} shows 10 m wind speeds in ERA5 and the forecasts.
For \himodelprob we plot both forecasts from randomly sampled ensemble members and a cherry-picked best member that was deemed to most closely match ERA5.
Note that the 1.5° resolution that we work with makes determining similarity or exact positions somewhat challenging.
The resolution also puts some limitations on how well the most extreme wind speeds can be captured.

We see in \cref{fig:laura_plot} that at 7 days lead time there is great uncertainty, and the deterministic models do not show the hurricane at all.
In the ensemble forecast from \himodelprob there exists however already members indicating the possibility of a hurricane making landfall a week ahead (see for example the cherry-picked \qm{Best member}).
While the ensemble includes many possible scenarios, a total of 7 members show the development of a hurricane in the area.
Having information of such possible scenarios a long time ahead allows for planning and readying disaster response efforts that might be needed.
Note that discovering these scenarios is only possible through an ensemble forecast, as the deterministic models do not indicate such an event.
At 5 days lead time all models are indicating the development of a hurricane.
While the deterministic models do a good job here, they are indicating a landfall location slightly too far eastward.
The ensemble members from \himodelprob however show a range of different positions for the hurricane, indicating the uncertainty in the landfall location.
At 3 days ahead 42 out of 50 ensemble members show the hurricane making landfall, but still with some uncertainty about the exact location.
At 1 day lead time all models give an accurate forecast of the position of the hurricane.
Apart from position, it is also interesting to consider how the models capture the intensity in terms of wind speeds.
At 1 day ahead the deterministic models are somewhat underestimating the wind speed.
The \himodelprob ensemble shows a range of possible values, indicating the uncertainty in the exact wind intensity.
Overall this study exemplifies how ensemble forecasts from a machine learning model can be used to discover possible extreme weather scenarios at long lead times and uncover the uncertainties associated with them.

\section{Spatial Coherency of Forecasts}
\label{sec:spatial_coherency}
Apart from giving a low forecast error, it is also important for forecasts to look realistic.
Specifically, we want forecasts to be \emph{physical}, containing features that are possible under laws of atmospheric physics.
This is necessary for forecasts to be interpretable by meteorologists and builds trust in the \gls{NeurWP} model.
One key property for realistic forecasts is that of spatial coherency. 
Forecasts should take realistic values not just locally, in each grid cell, but show larger-scale features that are consistent over the forecasting area.
This is important since the weather system contains spatial dependencies on multiple different scales.
Note that in the probabilistic setting this connects closely to capturing the correct joint distribution spatially, rather than just the marginal distributions in each grid cell.
As the metrics commonly used to evaluate \gls{NeurWP} models only consider the marginal distributions, it is important to also inspect sample forecasts to gauge the spatial coherency.

This modeling aspect is especially relevant for probabilistic models, since being able to visualize and interpret individual ensemble members is an important capability of ensemble forecasting.
Other uncertainty quantification techniques can be applied to estimate forecast uncertainty \cite{uq_for_data_models}, but these generally do not generate samples of possible future weather states.
Having samples to inspect can build trust and understanding of the model.
If one member is predicting an extreme event, it is possible to inspect this forecast and see what sequence of events in the atmosphere could lead to this outcome.

For the final \himodelprob model forecasts are generally spatially coherent and represent physically plausible scenarios.
Examples of this is shown in \cref{sec:global_forecasts_appendix,sec:lam_forecasts_appendix}.
In this appendix we contrast this with some of the shortcomings of baseline models.
We also discuss some of the challenges that we have encountered with achieving spatially coherent forecasts and ways to tackle these.

\subsection{\lamsection}
\newcommand{\sclamforecasthi}[2]{\begin{subfigure}[t]{0.24\textwidth}%
    \centering
    \includegraphics[width=\textwidth]{graphics/spatial_coherency/lam_prob_forecasts/hi_#1_57h.pdf}
    \captionsetup{justification=centering}
    \caption{\wvar{#2}\\\himodelprob}
    \label{fig:sclamforecasthi_#2}
\end{subfigure}}
\newcommand{\sclamforecastms}[2]{\begin{subfigure}[t]{0.24\textwidth}%
    \centering
    \includegraphics[width=\textwidth]{graphics/spatial_coherency/lam_prob_forecasts/ms_#1_57h.pdf}
    \captionsetup{justification=centering}
    \caption{\wvar{#2}\\\msmodelprob}
    \label{fig:sclamforecastms_#2}
\end{subfigure}}
\begin{figure}
    \centering
    \sclamforecasthi{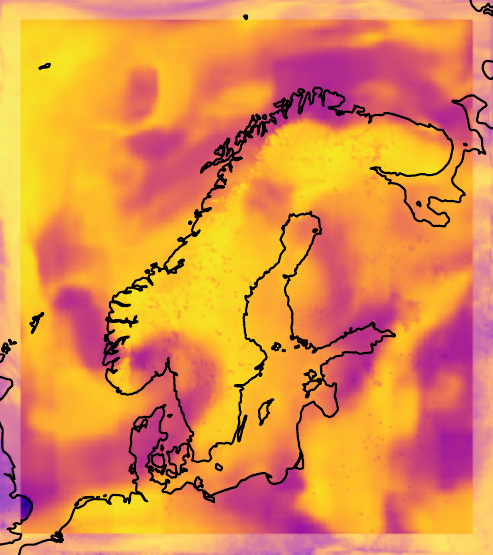}{nlwrs}%
    \sclamforecastms{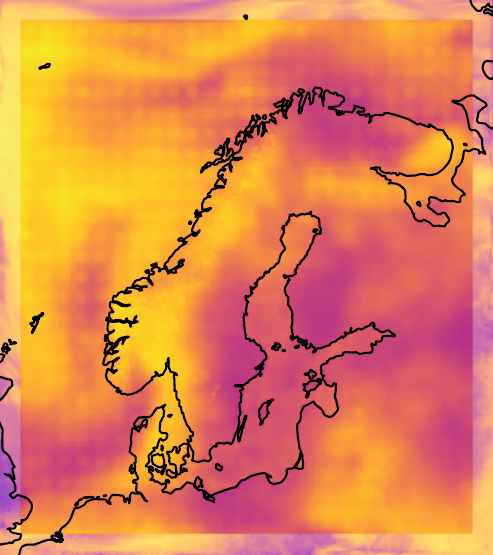}{nlwrs}\hspace{0.03\textwidth}%
    \sclamforecasthi{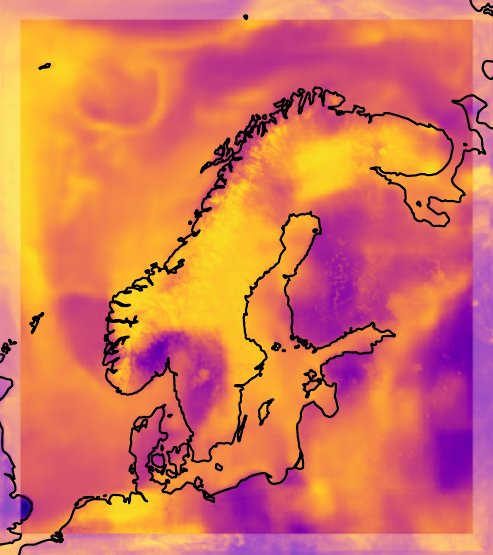}{2r}%
    \sclamforecastms{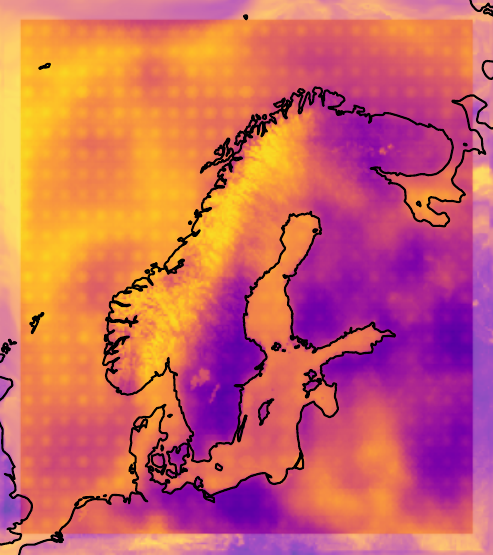}{2r}
    \sclamforecasthi{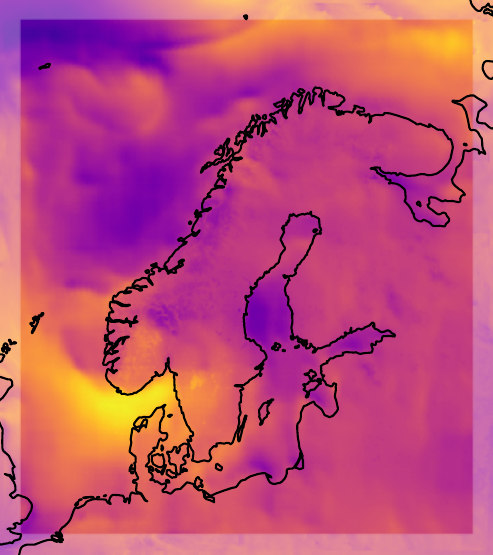}{u65}%
    \sclamforecastms{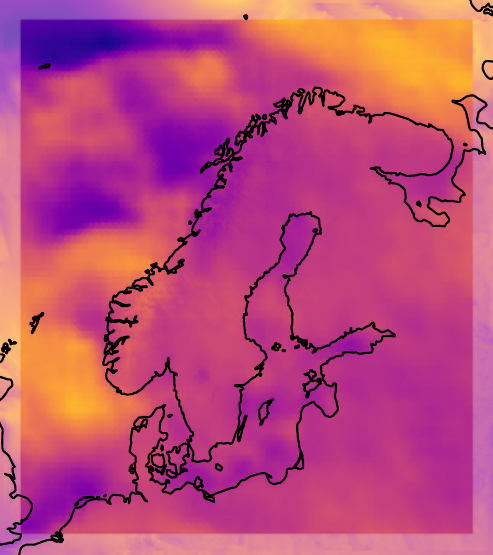}{u65}\hspace{0.03\textwidth}%
    \sclamforecasthi{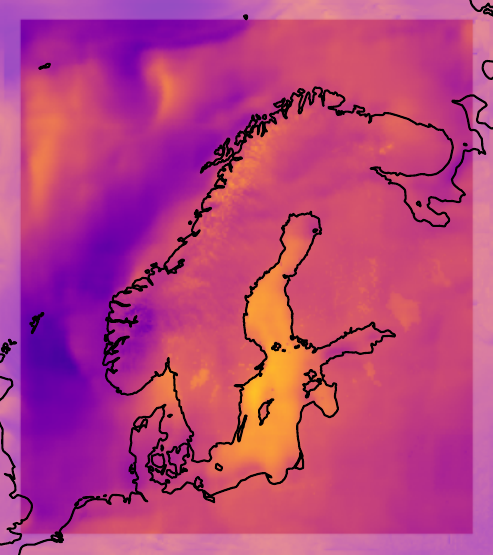}{v65}%
    \sclamforecastms{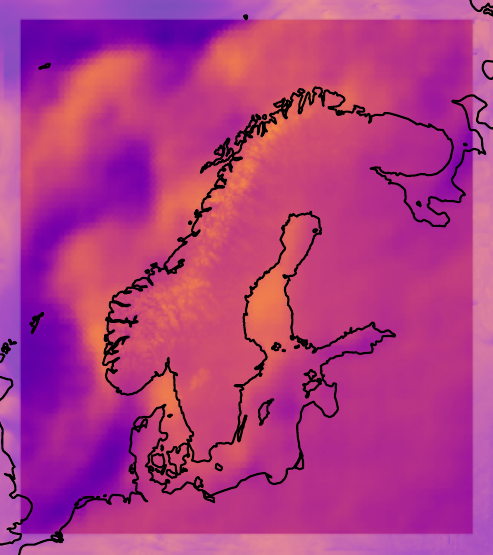}{v65}%
    \caption{Example forecasts from \himodelprob and \msmodelprob trained on the \gls{MEPS} data. All forecasts are from single ensemble members, plotted for lead time 57 h.}
    \label{fig:lam_sample_comparion}
\end{figure}
In the \gls{MEPS} experiment the benefits of the hierarchical graph structure become the most clear when visually inspecting forecasts.
\Cref{fig:lam_sample_comparion} shows a comparison between a few forecasts from \himodelprob and \msmodelprob.
In the \msmodelprob model samples tend to be more patchy and poorly reproduce spatial features.
We connect this to the fact that the randomness in $\latent^\timei$ is more local, as it is associated with the nodes in $\msgraph$ which directly connect to the grid.
While the randomness introduced from sampling $\latent^\timei$ can spread to the full forecast area using the multi-scale graph, this is something the model has to explicitly learn.
For matching marginal distributions in grid cells it can be enough for \msmodelprob to just have the randomness in each mesh node impact the connected grid nodes locally.
We can contrast this to \himodelprob, where $\latent^\timei$ is associated with nodes at the top of the hierarchical graph.
For this randomness to affect the prediction it must necessarily be propagated down the hierarchy, which disperses it spatially by construction.

We have observed that the \gls{CRPS} fine-tuning (see \cref{sec:crps_fine_tuning}) can be central to the performance of \himodelprob.
This aligns the model distribution with the distribution of the data.
However, since this objective only encourages matching of the marginal distributions it is not sufficient for making the model capture the full joint distribution.
This has to be achieved by constraints to the model or through additional parts of the training objective.
We have observed that the weighting $\crpsweight$ used for the \gls{CRPS} term can have a large impact on the spatial coherency of forecasts.
When $\crpsweight$ is chosen too high the models trade off the ability to generate meaningful large-scale patterns for capturing local variations.
This is especially severe for the \msmodelprob model, forcing us to use lower $\crpsweight$ values in order to still get physical-looking forecasts.
\Cref{fig:checkerboard_artefacts} shows an example of the type of local noise-patterns that can appear when training \msmodelprob with a too high $\crpsweight$.

\newcommand{\checkforecast}[2]{\begin{subfigure}[t]{0.30\textwidth}%
    \centering
    \includegraphics[width=\textwidth]{graphics/spatial_coherency/check_artifacts/memb1_#1_57h.pdf}
    \captionsetup{justification=centering}
    \caption{\wvar{#2}}
    \label{fig:checkforecast_#2}
\end{subfigure}}
\begin{figure}
    \centering
    \checkforecast{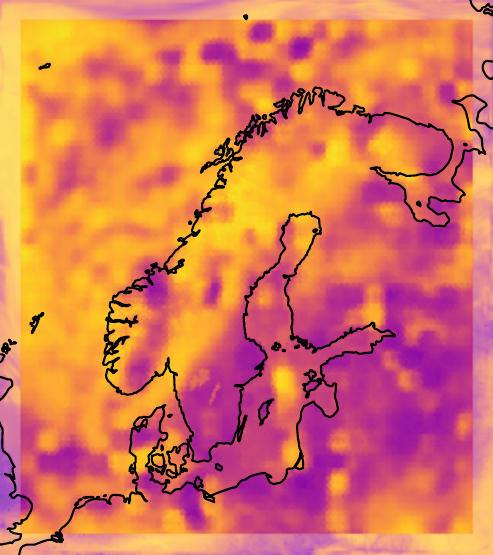}{nlwrs}%
    \hspace{0.02\textwidth}%
    \checkforecast{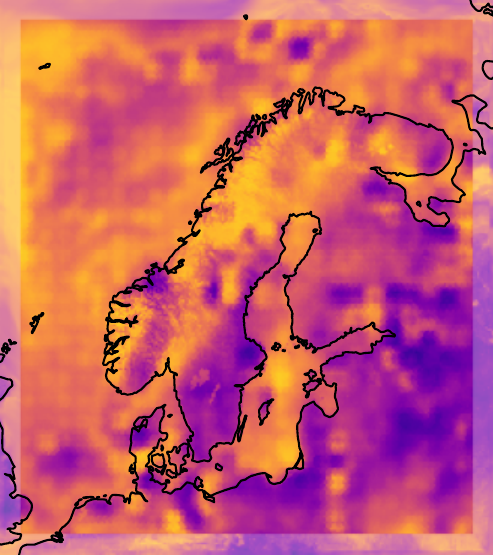}{2r}%
    \hspace{0.02\textwidth}%
    \checkforecast{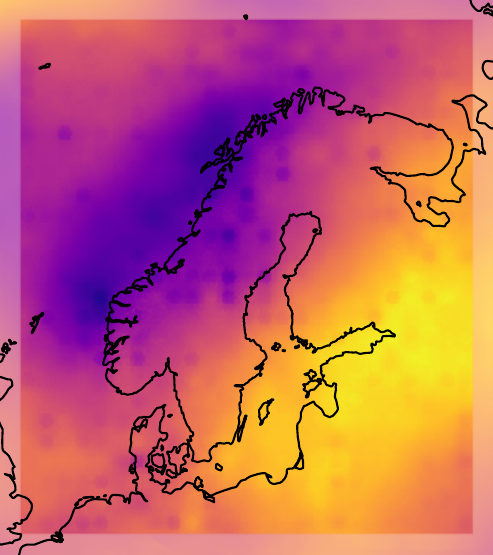}{z1000}%
    \caption{
    Examples of artifacts appearing in \msmodelprob forecasts when trained with too high $\crpsweight$ on the \gls{MEPS} data.
    The checkerboard-like pattern can be related to the layout of the multi-scale graph used.
    }
    \label{fig:checkerboard_artefacts}
\end{figure}

Another problem observed with the multi-scale graph $\msgraph$ in the \gls{LAM} setting is the appearance of circular artifacts over the forecasting area.
These appear in samples from \msmodelprob (see \cref{fig:sclamforecastms_nlwrs,fig:sclamforecastms_2r}), but can also be found in deterministic forecasts from \gcour.
These artifacts can be traced to the heterogeneous structure of the multi-scale graph.
Due to how $\msgraph$ is constructed, mesh nodes can have different number of neighbors.
The artifact positions match the positions of mesh nodes with many neighbors.
Note that the hierarchical graph does not have this issue, as $\graph_1$ (that connects to the grid) has a more uniform structure.
We noticed that these artifacts are even more prevalent for smaller models, as shown in \cref{fig:ms_artifacts_det}.
This points to the fact that with increased capacity, the \gcour model can to some extent learn to compensate for this problem.

\newcommand{\smallhifc}[2]{\begin{subfigure}[t]{0.24\textwidth}%
    \centering
    \includegraphics[width=\textwidth]{graphics/spatial_coherency/small_det_models/hi/#1_57h.pdf}
    \captionsetup{justification=centering}
    \caption{\wvar{#2}\\\himodeldet}
    \label{fig:smallhifc_#2}
\end{subfigure}}
\newcommand{\smallgcfc}[2]{\begin{subfigure}[t]{0.24\textwidth}%
    \centering
    \includegraphics[width=\textwidth]{graphics/spatial_coherency/small_det_models/gc/#1_57h.pdf}
    \captionsetup{justification=centering}
    \caption{\wvar{#2}\\\gcour}
    \label{fig:smallgcfc_#2}
\end{subfigure}}
\begin{figure}
    \centering
    \smallhifc{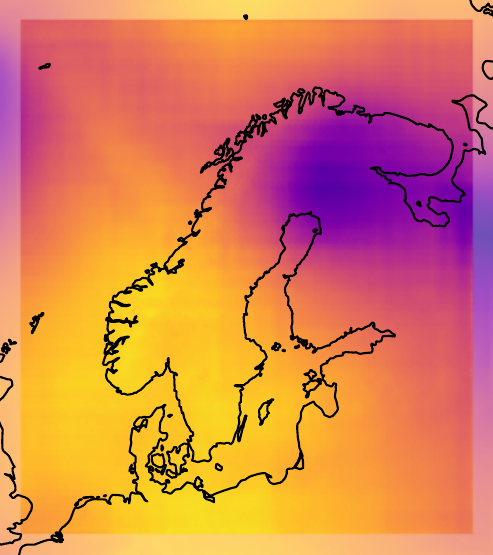}{pres0s}%
    \smallgcfc{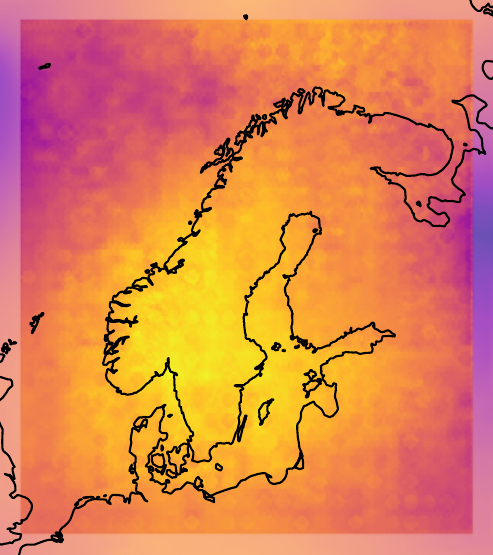}{pres0s}\hspace{0.03\textwidth}%
    \smallhifc{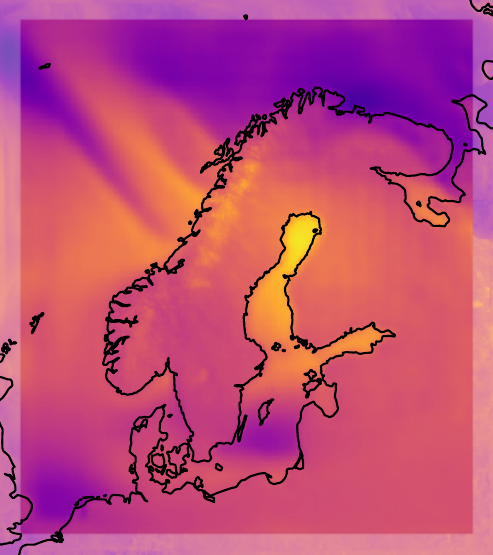}{u65}%
    \smallgcfc{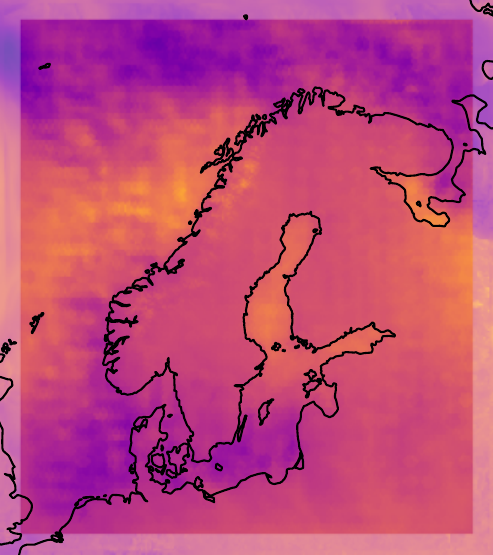}{2r}
    \caption{
    Examples of artifacts in smaller versions of the deterministic models.
    The circular artifacts in \gcour become much more prevalent when the model capacity is limited.
    These models use 4 processing steps, $\latentdim = 64$ and are trained using \gls{NLL} loss.
    }
    \label{fig:ms_artifacts_det}
\end{figure}

\subsection{\globalsection}
\label{sec:global_artifacts}
We have found it less challenging to achieve spatially coherent forecasts in the global setting.
Both \himodelprob and \msmodelprob generally produce samples with realistic physical features. 
Early in the training process we observe some hexagonal patterns for longer lead times (also noted by \citet{keisler}), but these disappear as we train the models on longer rollouts.

\newcommand{\globalarthi}[1]{\begin{subfigure}[t]{0.48\textwidth}%
    \centering
    \includegraphics[width=\textwidth]{graphics/spatial_coherency/global_prob_artifacts/hi/#1.pdf}
    \captionsetup{justification=centering}
    \caption{\wvar{#1}\\\himodelprob}
    \label{fig:smallhifc_#1}
\end{subfigure}}
\newcommand{\globalartms}[1]{\begin{subfigure}[t]{0.48\textwidth}%
    \centering
    \includegraphics[width=\textwidth]{graphics/spatial_coherency/global_prob_artifacts/ms/#1.pdf}
    \captionsetup{justification=centering}
    \caption{\wvar{#1}\\\msmodelprob}
    \label{fig:smallmsfc_#1}
\end{subfigure}}

\begin{figure}
    \centering
    \globalarthi{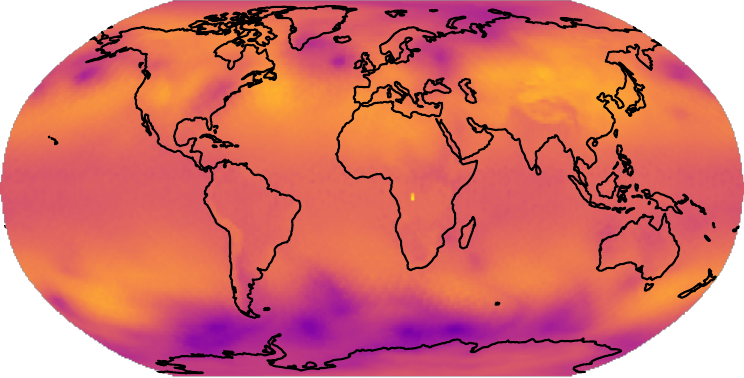}\hspace{0.03\textwidth}%
    \globalartms{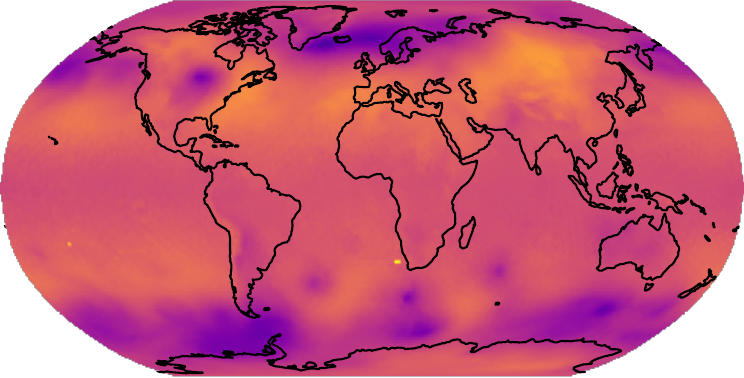}
    \globalarthi{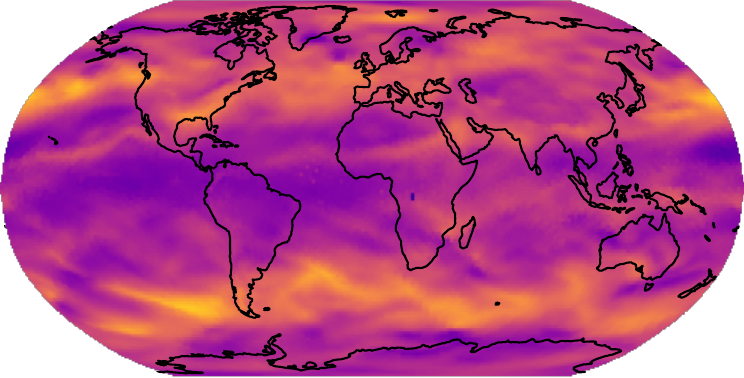}\hspace{0.03\textwidth}%
    \globalartms{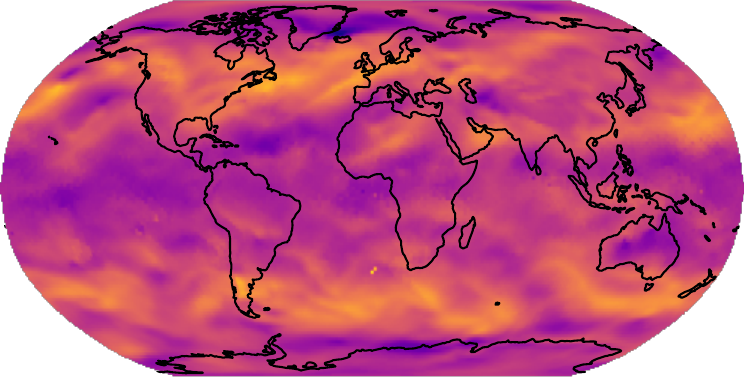}%
    \caption{
    Examples of artifacts in preliminary version of the global models unrolled to 10 days lead time. 
    Small patches of deviating values appear and impact multiple variables.
    While these were more prevalent in \msmodelprob, we did also observe this issue in \himodelprob.
    }
    \label{fig:global_artifacts}
\end{figure}

One challenge that we encountered with the global probabilistic models was the appearance of small areas of instabilities when rolling out the models to longer lead times. 
In \cref{fig:global_artifacts} we show some examples of this in forecasts from preliminary models.
These unphysical artifacts typically covered only a few grid-cells, but are of course undesirable in a forecasting model.
To remedy this for \himodelprob we found it sufficient to train on longer rollouts.
The stabilizing effect of unrolling up to $\forecastlength=12$ time steps during training did suppress these issues.
This was however not enough to solve the problem for \msmodelprob, for which we additionally needed to further lower $\crpsweight$.

\section{Experiment Details: \globalsection}
\label{sec:global_details}
In this appendix we give further detail about the graphs, data and experimental setup used in our global forecasting experiment.

\subsection{Graph Construction}
\label{sec:global_graph_construction}

\paragraph{Multi-scale graph}
The global multi-scale graph is created in the same way as in \gc \cite{graphcast}, by recursively splitting the faces of an icosahedron.
As we work with data on a coarser resolution, we perform only 4 steps of such splitting, resulting in the graphs $\graph_5, \dots, \graph_1$. 
These are all then merged to create $\msgraph$, used by \gcour and \msmodelprob.
By splitting 4 times we end up with a ratio of $\frac{N}{\setsize{\meshnodes}} \approx 11$ between grid nodes and mesh nodes.
This can be compared with the 6 splitting steps of \gc, resulting in a ratio $\frac{N}{\setsize{\meshnodes}} \approx 25$ for their 0.25\textdegree data.
We initially experimented also with splitting only 3 times (resulting in $\frac{N}{\setsize{\meshnodes}} \approx 45$), but models using these graphs showed inferior performance.

\paragraph{Hierarchical graph}
As the original icosahedron $\graph_5$ contains very few nodes, we do not use this for the hierarchical mesh graph, but rather construct the hierarchy using $\graph_4, \dots, \graph_1$.
Edges $\hiinteredges{\leveli}{\leveli+1}$ between levels are constructed by connecting each mesh node at level $\leveli$ with nodes at level $\leveli+1$ within a distance of $1.1$ times the edge length in $\graph_\leveli$.
This guarantees that each node at level $\leveli$ has 1 or 2 connection to the level above (see \cref{fig:global_inter_graph_example}).
Downward edges $\hiinteredges{\leveli+1}{\leveli}$ are created by simply flipping the edge directions of $\hiinteredges{\leveli}{\leveli+1}$.
We visualize the global mesh graphs in \cref{fig:global_graph_plots} and list
the exact number of nodes and edges in \cref{tab:global_graph_details}.

\newcommand{\globgraphplot}[1]{\begin{subfigure}[b]{0.24\textwidth}
    \centering
    \includegraphics[width=\textwidth]{graphics/graph_plots/global/g_#1.jpg}
    \caption{$\graph_#1$}
\end{subfigure}}
\begin{figure}[tbp]
    \centering
    \globgraphplot{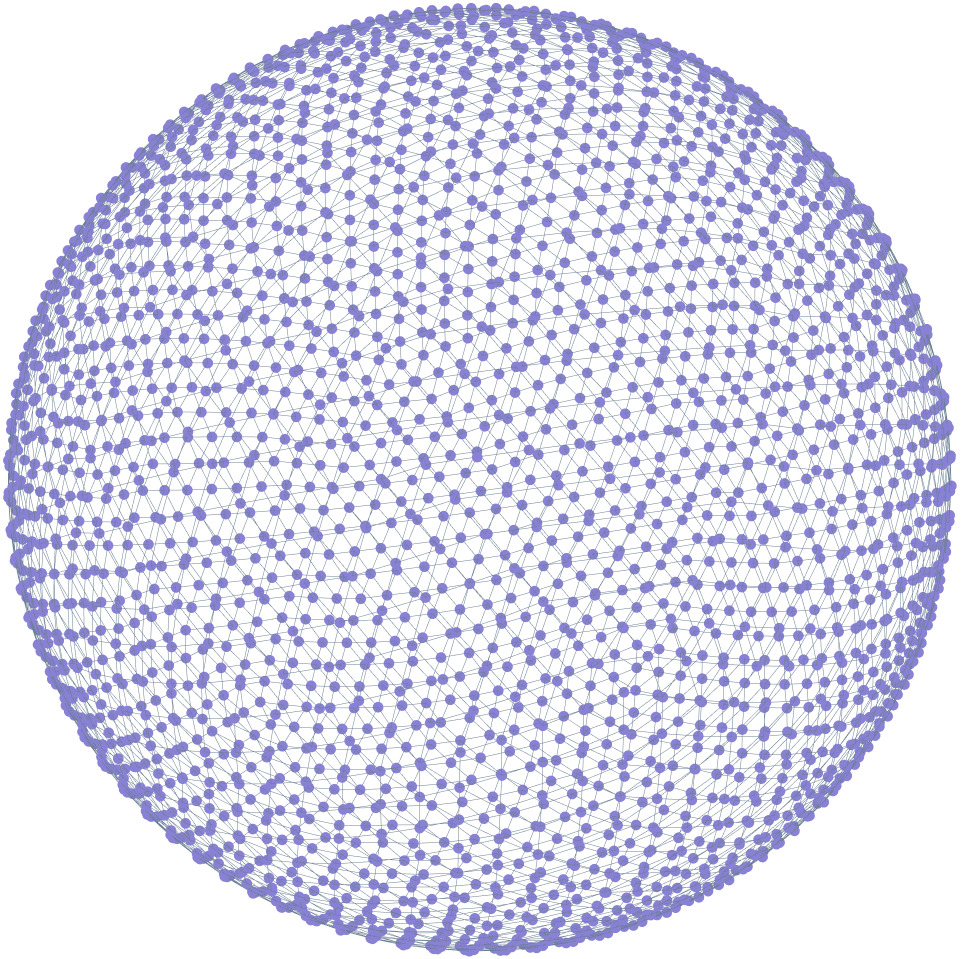}\hspace{0.01\textwidth}%
    \globgraphplot{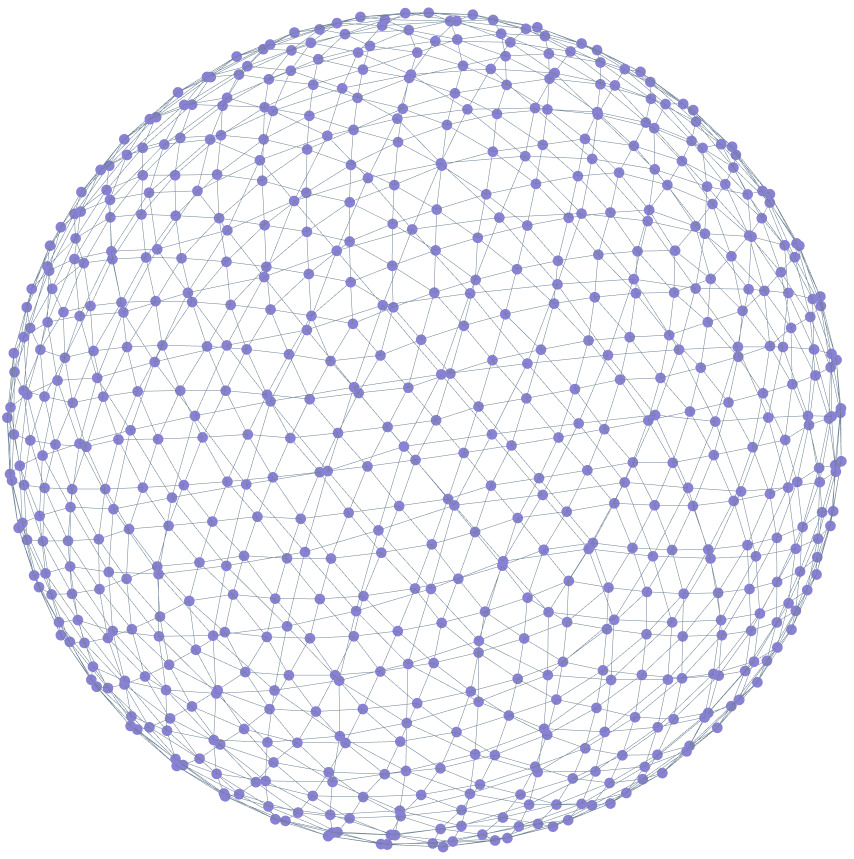}\hspace{0.01\textwidth}%
    \globgraphplot{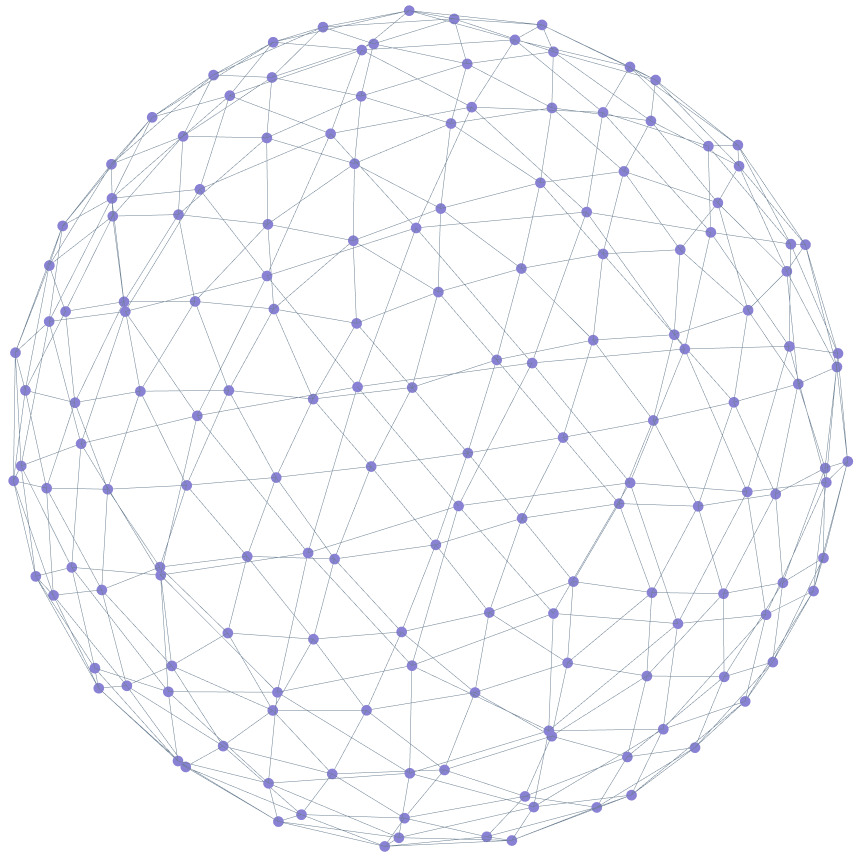}\hspace{0.01\textwidth}%
    \globgraphplot{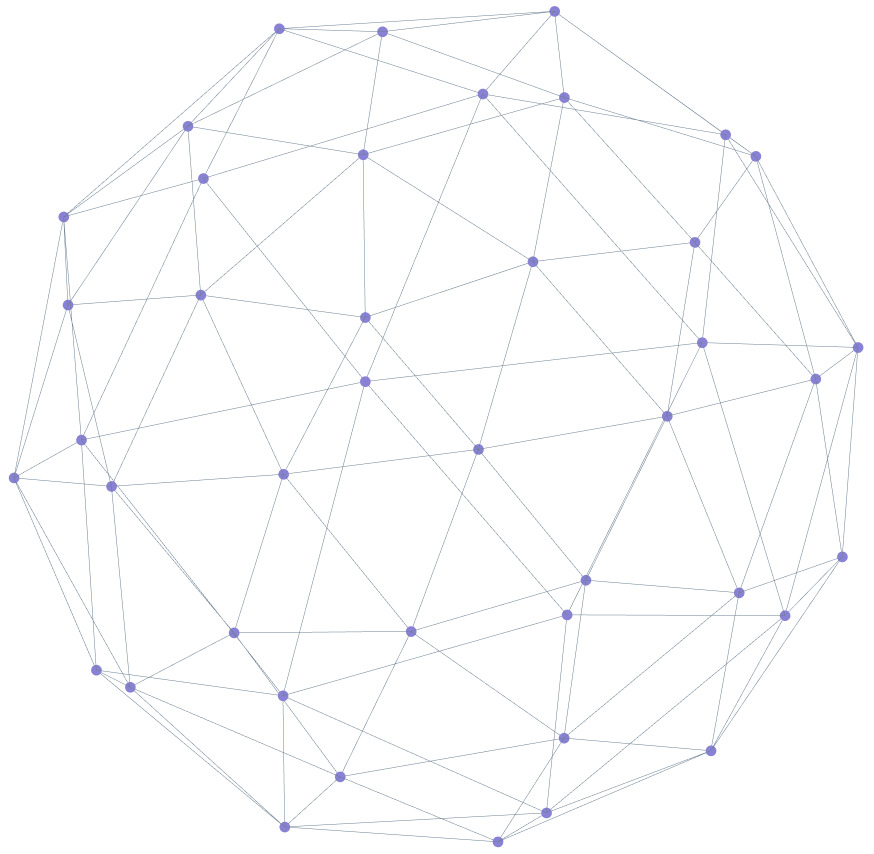}
    \begin{subfigure}[b]{0.5\textwidth}
        \centering
        \includegraphics[width=\textwidth]{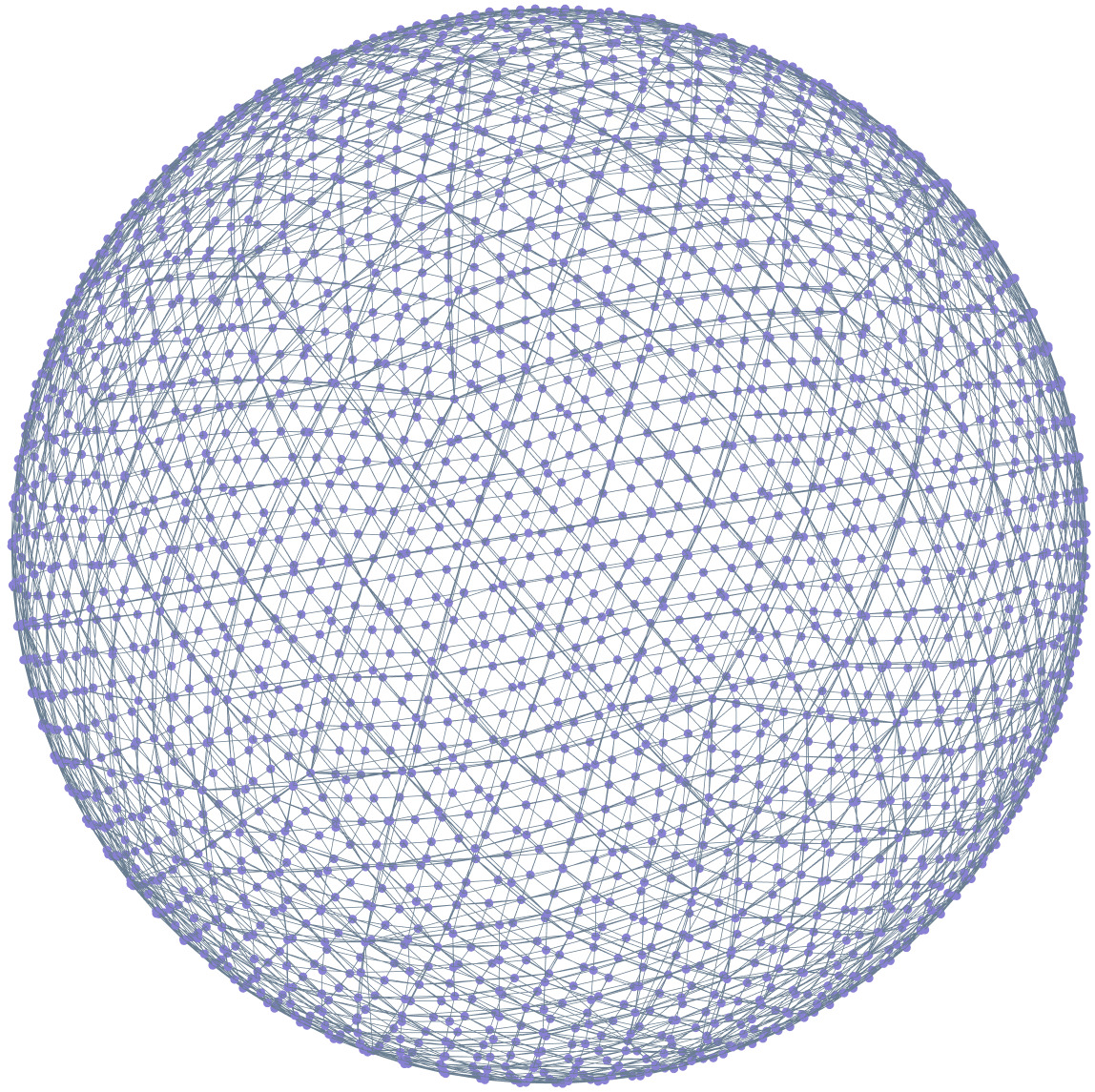}
        \caption{Multi-scale mesh graph $\msgraph$}
    \end{subfigure}
    \begin{subfigure}[b]{0.48\textwidth}
        \centering
        \includegraphics[width=\textwidth]{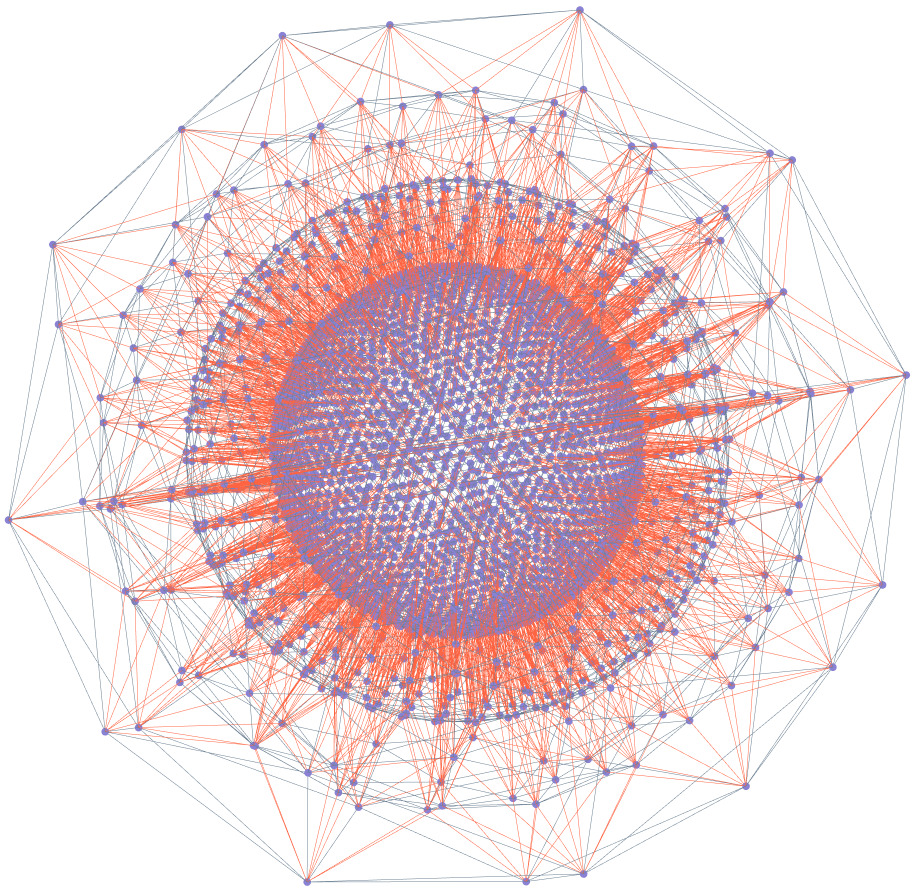}
        \caption{Hierarchical mesh graph}
    \end{subfigure}\hspace{0.04\textwidth}%
    \begin{subfigure}[b]{0.48\textwidth}
        \centering
        \includegraphics[width=\textwidth]{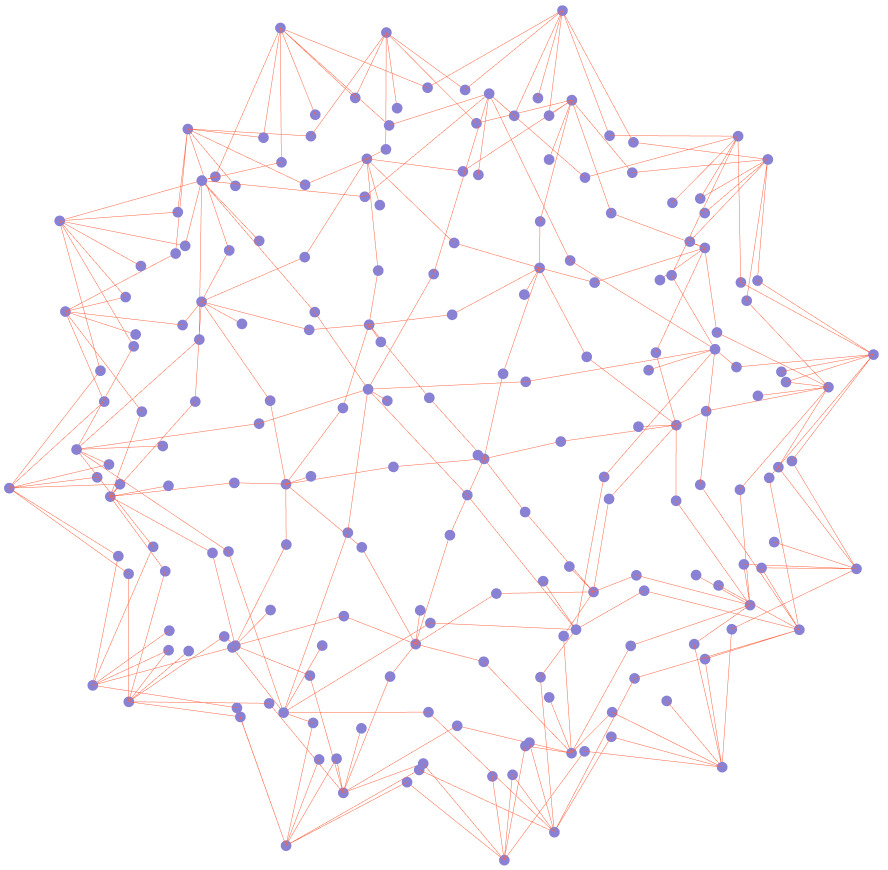}
        \caption{Inter-level graph $\hiintergraph{3}{4}$}
        \label{fig:global_inter_graph_example}
    \end{subfigure}
    \caption{Mesh graphs used in the global experiment. Note that the vertical positioning (away from earth's surface) is purely for visualization purposes.}
    \label{fig:global_graph_plots}
\end{figure}

\begin{table}[tbp]
\centering
\caption{Global graph statistics.}
\label{tab:global_graph_details}
\begin{tabular}{@{}llS[table-format=5.0,table-text-alignment=right]S[table-format=5.0,table-text-alignment=right]@{}}
\toprule
\textbf{Graph} & \textbf{}           & \textbf{Nodes} & \textbf{Edges} \\ \midrule
\multirow{8}{*}{Hierarchical} & $\higraph{1}$         & 2562         & 15360         \\
               & $\higraph{2}$         & 642         & 3840         \\
               & $\higraph{3}$         & 162         & 960         \\
               & $\higraph{4}$         & 42         & 240         \\
               & $\hiintergraph{1}{2}$/$\hiintergraph{2}{1}$ & {-}              & 4482         \\
               & $\hiintergraph{2}{3}$/$\hiintergraph{3}{2}$ & {-}              & 1122         \\
               & $\hiintergraph{3}{4}$/$\hiintergraph{4}{3}$ & {-}              & 282         \\ \cmidrule(l){2-4} 
               & Total               & 3408         & 32172         \\ \midrule
$\msgraph$       &                     & 2562         & 20460         \\ \midrule
$\graph_\gtm$    &                     & {-}              & 46158         \\
$\graph_\mtg$    &                     & {-}              & 87120         \\ \midrule
Grid    &                     & 29040              & {-}         \\ \bottomrule
\end{tabular}
\end{table}

\paragraph{Connecting the grid and mesh}
Edges connecting the grid and mesh graph are also constructed following \citet{graphcast}.
The edges $\edgeset_\gtm$ are created by connecting each grid node to mesh nodes within a distance of 0.6 times the edge length in $\graph_1$.
Edges $\edgeset_\mtg$ are constructed by for each grid node finding the triangle in $\graph_1$ containing it, and adding three edges from the corner mesh nodes of that triangle.
Note that the edge sets $\edgeset_\gtm$ and $\edgeset_\mtg$ are identical for the multi-scale and hierarchical mesh graphs, since $\msgraph$ and $\higraph{1}$ contain the same nodes.
Since the grid nodes are laid out in a latitude-longitude grid and the mesh is icosahedral all mesh nodes will not be connected to the same number of grid nodes.
Specifically, mesh nodes close to the poles will have many more grid connections than mesh nodes around the equator.

\paragraph{Static features}
All mesh nodes are associated with static features related to their position, specifically $\cos(\lat)$, $\sin(\lon)$ and $\cos(\lon)$.
Edges (in the mesh, $\edgeset_\gtm$ and $\edgeset_\mtg$) have static features containing the edge length and vector difference between its endpoints (see \citet{graphcast} for details).
Edge features are normalized by the length of the longest edge. 

\subsection{Dataset Details}
We use a version of ERA5 \cite{era5} re-gridded to 1.5\textdegree{} latitude-longitude gridding, provided through the WeatherBench 2 benchmark \cite{weatherbench2}.
Data from the period 1959-01-01T12 to 2017-12-31T12 is used for training, 2017-12-31T18 to 2019-12-31T12 for validation and 2019-12-31T18 to 2021-01-10T18 for testing.
These exact time stamps guarantee that there is no overlap between the subsets.
For evaluation we consider forecasts initialized at times 00 and 12 UTC each day of 2020, following WeatherBench 2.
We perform 10 day forecasts using 6 h time steps.
During training we also start forecasts at times 06 and 18 UTC.

The exact set of forecast variables, forcing and static fields is listed in \cref{tab:global_variables}.
We use the same inputs as \citet{graphcast}.
The forcing is windowed over three consecutive time steps, meaning that each $\forcing^\timei$ contains forcing from times $\timei-1$, $\timei$ and $\timei+1$ as well as the the static fields.
For training we rescale the values of each variable to zero mean and unit variance.
\newcommand{\pressurelevels}{\preslevel{50}, \preslevel{100}, \preslevel{150}, \preslevel{200}, \preslevel{250}, \preslevel{300}, \preslevel{400}, \preslevel{500}, \preslevel{600}, \preslevel{700}, \preslevel{850}, \preslevel{925}, \preslevel{1000}}
\begin{table}[tbp]
\centering
\caption{Variables, forcing and static fields from ERA5 used in our global forecasting experiments. \textsuperscript{\textdagger}~\si{\kilo\gram} of water vapour per \si{\kilo\gram} of air.}
\label{tab:global_variables}
\begin{tabular}{@{}llcl@{}}
\toprule
& Abbreviation & Unit & Vertical Levels \\ \midrule
\tablesubsec{Variables}
    \varline{z}{\multirow{6}{0.35\textwidth}{\pressurelevels}}{\si{\meter\squared\per\second\squared}}{Geopotential} 
    \varline{q}{}{\si{\kilo\gram\per\kilo\gram}\textsuperscript{\textdagger}}{Specific humidity}
    \varline{t}{}{\si{\kelvin}}{Temperature}
    \varline{u}{}{\si{\meter\per\second}}{$u$-component of wind}
    \varline{v}{}{\si{\meter\per\second}}{$v$-component of wind}
    \varline{w}{}{\si{\pascal\per\second}}{Vertical velocity}
    \midrule
    \varline{t}{2 m above surface}{\si{\kelvin}}{Temperature}
    \varline{u}{10 m above surface}{\si{\meter\per\second}}{$u$-component of wind}
    \varline{v}{10 m above surface}{\si{\meter\per\second}}{$v$-component of wind}
    \varline{msl}{Sea level}{\si{\pascal}}{Mean sea level pressure}
    \varline{tp}{Surface}{\si{\metre}}{Total precipitation (Acc. over \qty{6}{\hour})}
    \midrule
\tablesubsec{Forcing}
    \varline{toa}{Top of atmosphere}{\si{\watt\per\metre\squared}}{Top of atm. solar radiation flux}
    \varline{sin_tod}{-}{-}{Sine-encoded time of day}
    \varline{cos_tod}{-}{-}{Cosine-encoded time of day}
    \varline{sin_toy}{-}{-}{Sine-encoded time of year}
    \varline{cos_toy}{-}{-}{Cosine-encoded time of year}
    \midrule
\tablesubsec{Static Fields}
    \varline{water}{Surface}{$[0,1]$}{Land-sea mask}
    \varline{topography}{Surface}{\si{\meter\squared\per\second\squared}}{Surface geopotential (topography)}
    \varline{cos_lat}{-}{-}{$\cos(\lat)$}
    \varline{sin_lon}{-}{-}{$\sin(\lon)$}
    \varline{cos_lon}{-}{-}{$\cos(\lon)$}
\bottomrule
\end{tabular}
\end{table}

\subsection{Model and Training Configurations}
\label{sec:global_details_model_conf}
We train all models using the AdamW optimizer \citeapp{adamw} and utilize BFloat16 mixed precision to save GPU memory.
The training costs for the models makes extensive hyperparameter tuning unfeasible.
We choose hyperparameters based on initial experimentation with smaller models.
For \himodelprob the important weightings $\klweight$ and $\crpsweight$ in $\loss$ can be chosen based on monitoring the model behavior during training.
This was done by plotting example forecasts throughout the training process and monitoring metrics on the validation set.
The weight $\klweight$ was tuned to prevent the model from collapsing to deterministic predictions, and to still achieve useful predictions when $\latent^\timei$ was sampled from the latent map.
The \gls{CRPS} weight $\crpsweight$ was tuned to achieve a good ensemble spread while avoiding artifact issues, as discussed in \cref{sec:spatial_coherency}.

The exact model configurations and training times for our global experiments are listed in \cref{tab:global_model_details}. 
Parameter counts for probabilistic models include parameters in the variational approximation, although this component does not play a role during forecasting. 
We report training times as hours of active computations on a single GPU. 
In practice we use 8 80GB NVIDIA A100 GPUs in parallel, meaning that the wall-clock time of our training is given by dividing the numbers in \cref{tab:global_model_details} by 8.
For the \himodeldet model one processing step on the mesh graph constitutes a full pass through the hierarchy (either from the grid up or from level $\himaxl$ down).
In \msmodelprob multiple \gls{GNN} processing steps operate on $\msgraph$.
We list the number of such steps in \cref{tab:global_model_details} separately for the latent map ($\meanfunc{\latent}$), predictor ($\predictorfunc$) and variational approximation ($q$).

\begin{table}[tbp]
\centering
\caption{Details of model architectures and training times for global forecasting.}
\label{tab:global_model_details}
\begin{tabular}{@{}lccS[table-format=2.1e1]S[table-format=4.0]@{}}
\toprule
Model & \multicolumn{1}{c}{$\latentdim$} & \multicolumn{1}{c}{Processing steps} & {Parameters} & {Training time (GPU-hours)} \\ \midrule
\gcour      & 256 & 8 & 5.2e6 & 716 \ \\ 
\himodeldet & 256 & 8 & 30.9e6 & 1040 \\ 
\msmodelprob & 256 & $\meanfunc{\latent}$: 2, $\predictorfunc$: 4, $q$: 4 & 7.7e6 & 1372 \\ 
\himodelprob & 256 & - & 16.3e6 & 1264 \\ 
\bottomrule
\end{tabular}
\end{table}

We train all models by a sequence of training step, starting from single-step prediction and then unrolling predictions $\forecastlength$ steps.
For the probabilistic models we first train in a pure auto-encoder setup with $\klweight = 0$, encouraging $q$ to encode useful information in the distribution over $\latent^\timei$.
We find this useful for preventing the model from ignoring $\latent^\timei$ and collapsing to purely deterministic forecasting.
The probabilistic models additionally include the fine-tuning using \gls{CRPS} as a final training step.
The full training schedule for the deterministic models is given in \cref{tab:global_det_training} and for the probabilistic models in \cref{tab:global_prob_training}.
Note that we here define one epoch as initializing the model at each possible time in the training set (00, 06, 12 and 18 UTC in each day) such that all unrolled lead times are still within the training set period.
The reason for probabilistic models being unrolled to a longer lead time $\forecastlength$ is mainly to combat the artifacts discussed in \cref{sec:global_artifacts}.

\begin{table}[tbp]
\centering
\caption{Training schedule for deterministic models (\gcour and \himodeldet) on global data.}
\label{tab:global_det_training}
\begin{tabular}{@{}rcc@{}}
\toprule
Epochs & \multicolumn{1}{c}{Learning Rate} & \multicolumn{1}{c}{Unrolling $\forecastlength$} \\ \midrule
70    & $10^{-3}$ & 1      \\ 
20    & $10^{-4}$ & 4      \\ 
20    & $10^{-4}$ & 8      \\ \bottomrule
\end{tabular}
\end{table}
\begin{table}[tbp]
\centering
\caption{Training schedule for \himodelprob  on global data. For \msmodelprob we use the same schedule but with different constants ($\klweight = 1$, $\crpsweight = 10^3$).}
\label{tab:global_prob_training}
\begin{tabular}{@{}rcccc@{}}
\toprule
Epochs & \multicolumn{1}{c}{Learning Rate} & \multicolumn{1}{c}{Unrolling $\forecastlength$} & $\klweight$ & $\crpsweight$ \\ \midrule
50    & $10^{-3}$ & 1 & 0 & 0 \\ 
75    & $10^{-4}$ & 1 & 0.1 & 0 \\ 
20    & $10^{-4}$ & 4 & 0.1 & 0 \\ 
10    & $10^{-4}$ & 8 & 0.1 & 0 \\ 
2    & $10^{-4}$ & 12 & 0.1 & $10^4$ \\ \bottomrule
\end{tabular}
\end{table}
\section{Experiment Details: \lamsection}
\label{sec:lam_details}
In this appendix we give further detail about the graphs, data and experimental setup used in our experiments with the \gls{MEPS} data.

\subsection{Graph Construction}
\label{sec:lam_graph_construction}

\paragraph{Multi-scale graph}
In this limited-area setting we construct graphs as regular quadrilateral meshes covering the rectangular \gls{MEPS} forecasting area.
To create these we lay out mesh nodes in regular rows and columns over the area.
Each node is then connected with bidirectional edges to its neighbors horizontally, vertically and diagonally (see \cref{fig:graph_connections}).
This results in all nodes (except those at the edge of the area) having 8 neighbors.
The procedure is repeated at 4 different resolutions, tripling the distance between nodes at each resolution.
This means that a node at resolution level $l$ is positioned at the center of a group of $3 \times 3$ nodes at resolution level $l-1$, sharing its exact position with the center node of the group (illustrated in \cref{fig:graph_alignment}).
To create the multi-scale mesh graph $\msgraph$ we then merge the graphs at different resolutions, combining any nodes that sit at the same coordinates into one node.
Note that this is possible due to how nodes align across the resolution levels.

\newcommand{\tikzring}[1]{\tikz\draw[black,radius=#1,very thick] (0,0) circle ;}%
\newcommand{\tikzcircle}[2][LiUblue,fill=LiUblue]{\tikz\draw[#1,radius=#2] (0,0) circle ;}%
\begin{figure}[tbp]
    \centering
    \begin{subfigure}[t]{0.35\textwidth}
        \centering
        \includegraphics[width=\textwidth]{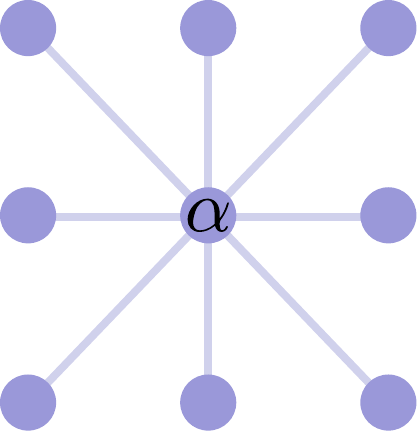}
        \caption{Each mesh node $\nodei$ is connected to its neighbors horizontally, vertically and diagonally.}
        \label{fig:graph_connections}
    \end{subfigure}%
    \hspace{0.1\textwidth}
    \begin{subfigure}[t]{0.45\textwidth}
        \centering
        \includegraphics[width=\textwidth]{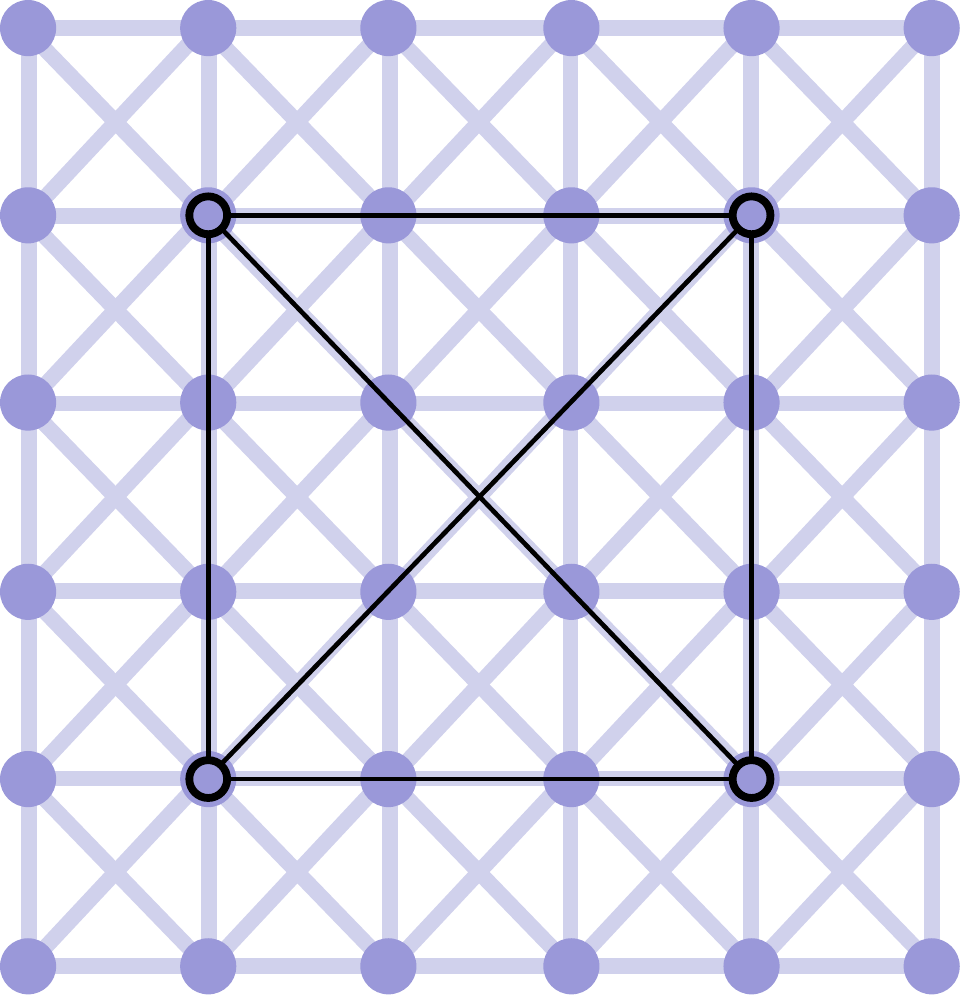}
        \caption{Alignment of mesh nodes \tikzring{2.5pt} in $\graph_\leveli$ with mesh nodes \tikzcircle[NodePurple,fill=NodePurple]{3pt} in $\graph_{\leveli-1}$.}
        \label{fig:graph_alignment}
    \end{subfigure}
    \caption{Illustration of mesh graph construction in the \gls{LAM} setting.}
    \label{fig:graph_construction}
\end{figure}

\begin{figure}[tbp]
        \centering
    \begin{subfigure}[b]{0.5\textwidth}
        \centering
        \includegraphics[width=\textwidth]{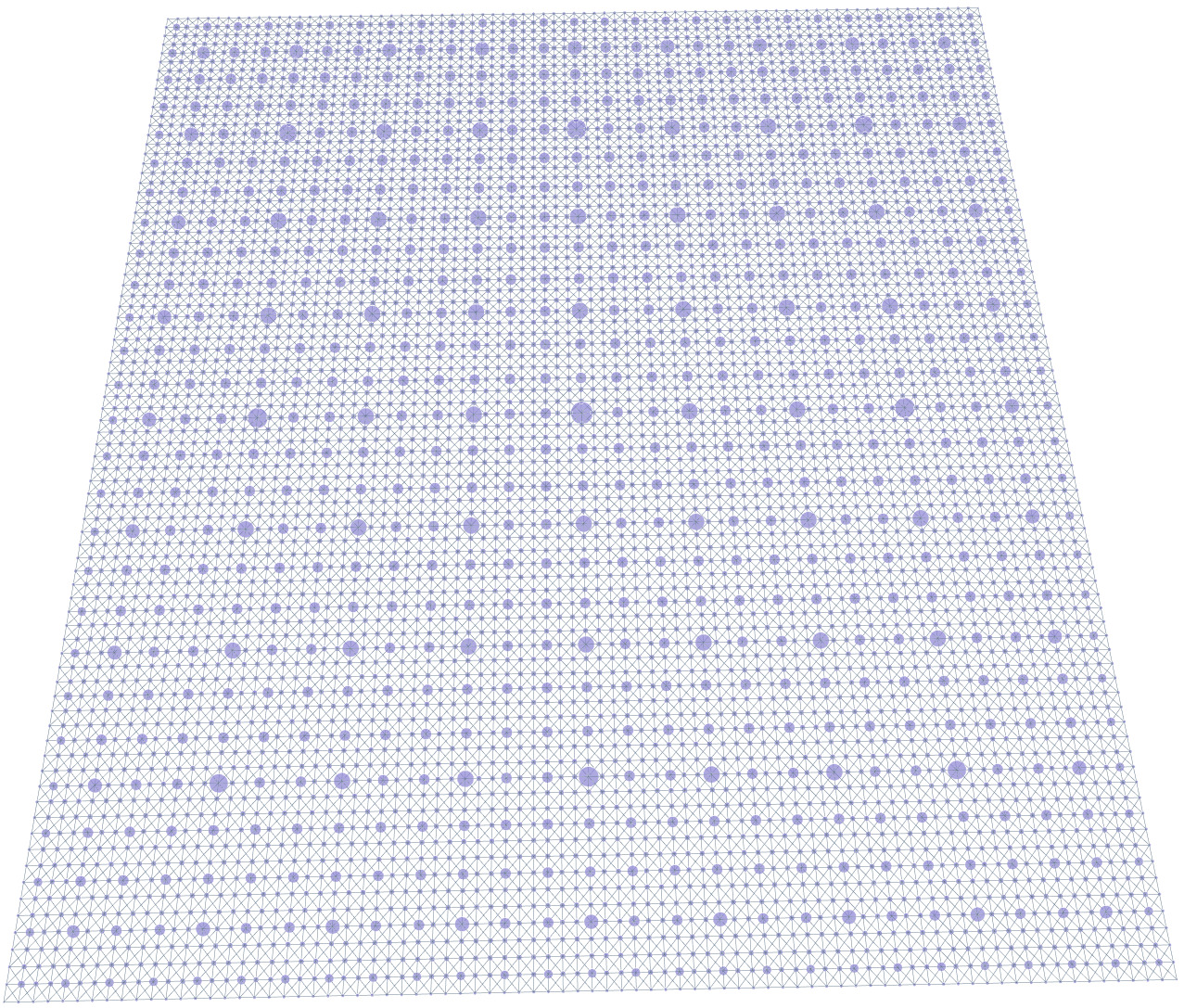}
        \caption{Multi-scale mesh graph $\msgraph$}
    \end{subfigure}\\
    \begin{subfigure}[b]{0.48\textwidth}
        \centering
        \includegraphics[width=\textwidth]{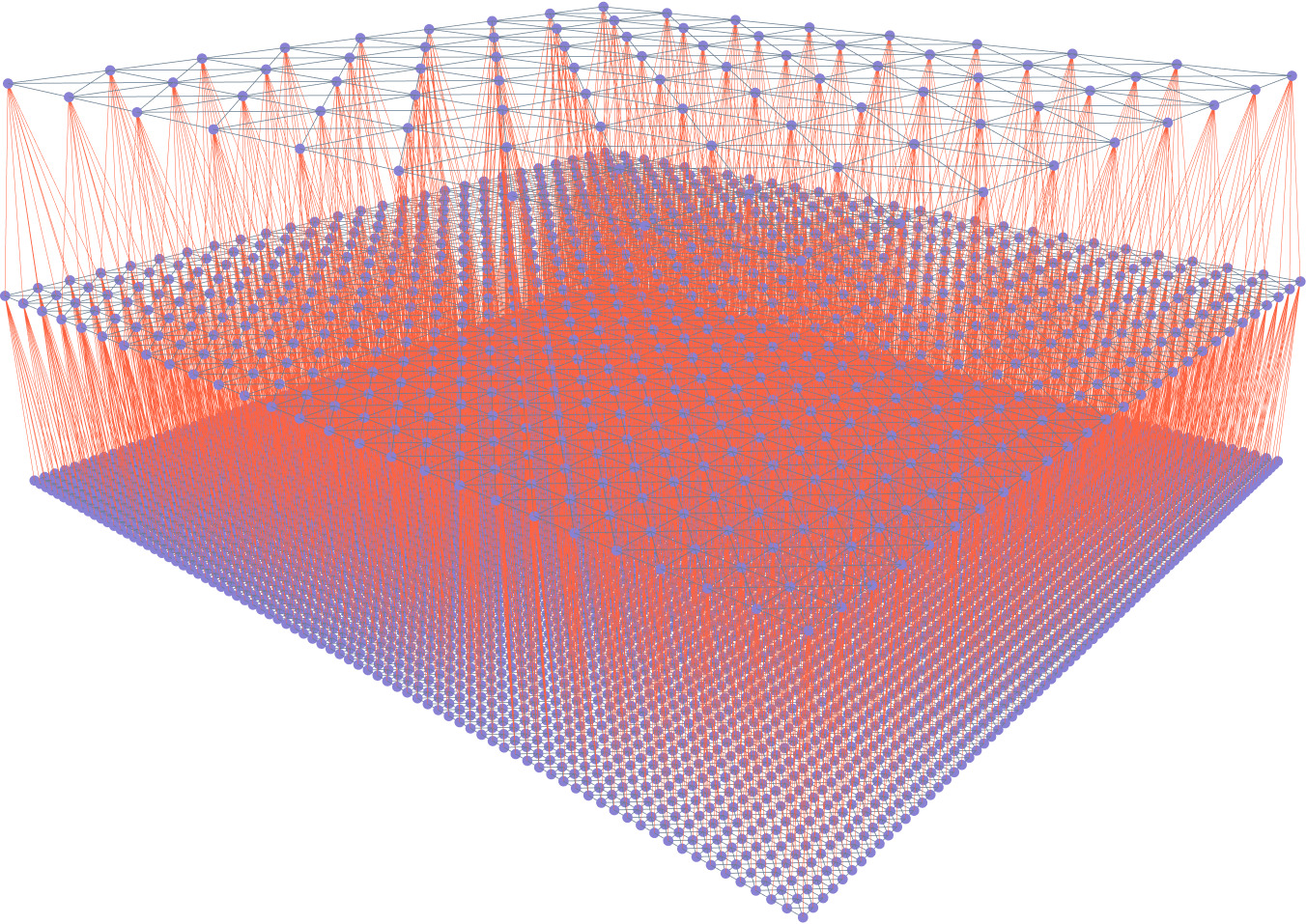}
        \caption{Hierarchical mesh graph}
    \end{subfigure}\hspace{0.04\textwidth}%
    \begin{subfigure}[b]{0.48\textwidth}
        \centering
        \includegraphics[width=\textwidth]{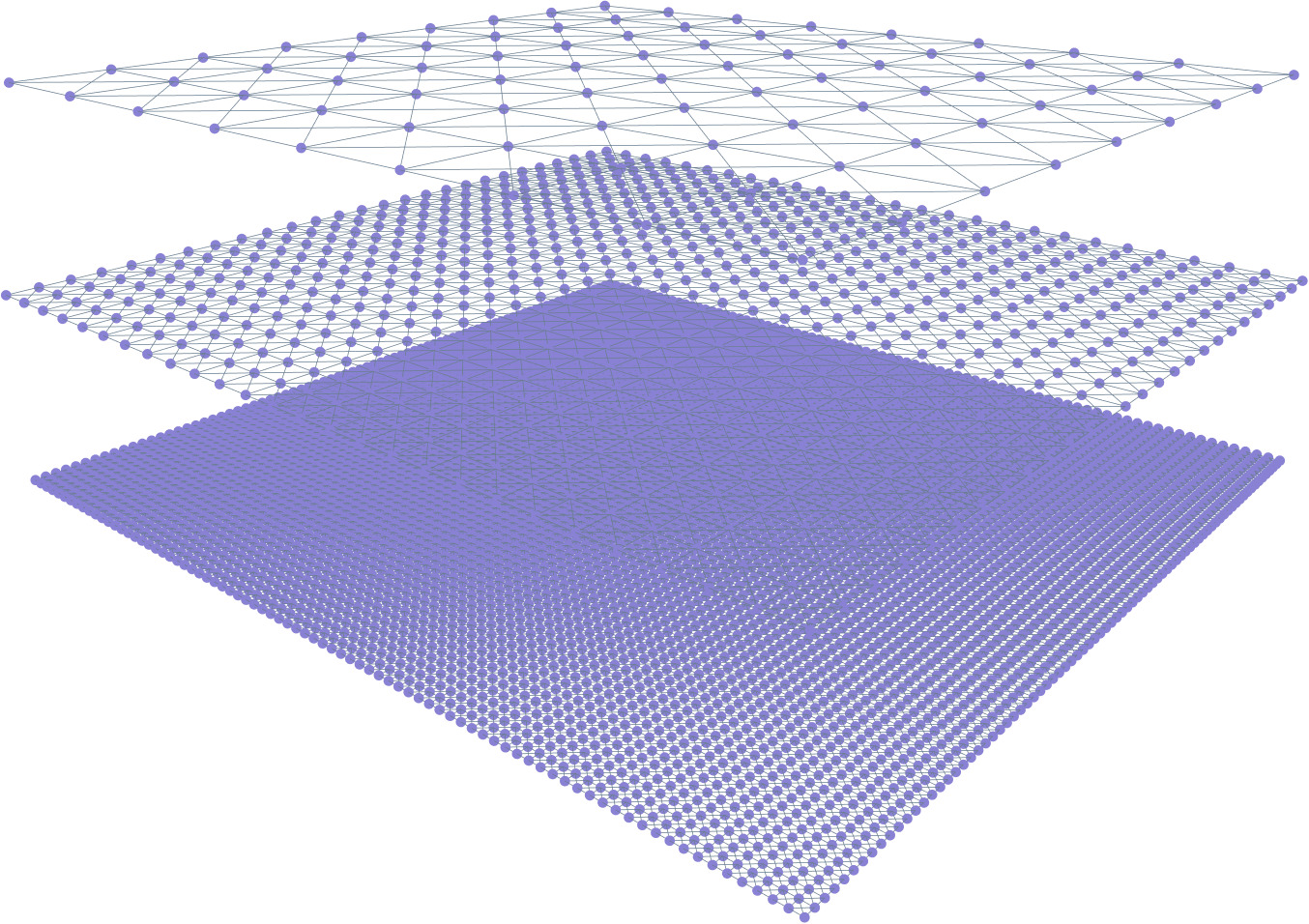}
        \caption{Mesh levels $\graph_1, \graph_2, \graph_3$}
    \end{subfigure}
    \caption{Mesh graphs used in the \gls{MEPS} experiment. Note that the vertical positioning and size of nodes is purely for visualization purposes.}
    \label{fig:lam_graph_plots}
\end{figure}

\paragraph{Hierarchical graph}
For the hierarchical model the graphs of different resolution are not merged, but used as the different levels of the hierarchy.
We include only the 3 finest meshes $\graph_1$, $\graph_2$ and $ \graph_3$, as $\graph_4$ contains only 9 nodes. 
Additional inter-level edge sets $\hiinteredges{\leveli}{\leveli+1}$ and $\hiinteredges{\leveli+1}{\leveli}$ are then created for all adjacent levels.
Each set $\hiinteredges{\leveli}{\leveli+1}$ of upwards edges is created by connecting each node on level $\leveli$ with the closest node on level $\leveli+1$.
This means that each node at levels $\leveli > 1$ will have 9 incoming edges from the level below.
The downward edges $\hiinteredges{\leveli+1}{\leveli}$ are a copy of $\hiinteredges{\leveli}{\leveli+1}$ with the direction of each edge flipped.
The mesh graphs used for the \gls{MEPS} experiment are visualized in \cref{fig:lam_graph_plots} and the number of nodes and edges in each graph listed in \cref{tab:lam_graph_details}.

\begin{table}[tbp]
\centering
\caption{\gls{MEPS} graph statistics.}
\label{tab:lam_graph_details}
\begin{tabular}{@{}llS[table-format=5.0,table-text-alignment=right]S[table-format=6.0,table-text-alignment=right]@{}}
\toprule
\textbf{Graph} & \textbf{}           & \textbf{Nodes} & \textbf{Edges} \\ \midrule
\multirow{6}{*}{Hierarchical} & $\higraph{1}$         & 6561         & 51520         \\
               & $\higraph{2}$         & 729         & 5512         \\
               & $\higraph{3}$         & 81         & 544         \\
               & $\hiintergraph{1}{2}$/$\hiintergraph{2}{1}$ & {-}              & 6561         \\
               & $\hiintergraph{2}{3}$/$\hiintergraph{3}{2}$ & {-}              & 729         \\\cmidrule(l){2-4}
               & Total               & 7371         & 72156         \\ \midrule
$\msgraph$       &                     & 6561         & 57616         \\ \midrule
$\graph_\gtm$    &                     & {-}              & 100656         \\
$\graph_\mtg$    &                     & {-}              & 255136         \\ \midrule
Grid    &                     & 63784              & {-}         \\ \bottomrule
\end{tabular}
\end{table}

\paragraph{Connecting the grid and mesh}
To form $\edgeset_\gtm$, each grid node is connected to mesh nodes closer than $0.67$ times the distance between nodes in $\graph_1$.
All distances are here 2-dimensional euclidean, computed in the \gls{MEPS} Lambert projection coordinates.
The set $\edgeset_\mtg$ is constructed by iterating over the grid nodes, and at each creating edges from the 4 closest mesh nodes to the node in the grid.

\paragraph{Static features}
The static features associated with mesh nodes are their 2-dimensional coordinates in the \gls{MEPS} Lambert projection, normalized by the maximum coordinate value.
All edges have static features including the length of the edge and the vector difference between the source and target nodes (using the projection coordinates).
The edge features are normalized by the length of the longest edge in the whole mesh graph.

\subsection{Dataset Details}
\label{sec:lam_dataset_details}
\begin{figure}
    \centering
    \includegraphics[width=\linewidth]{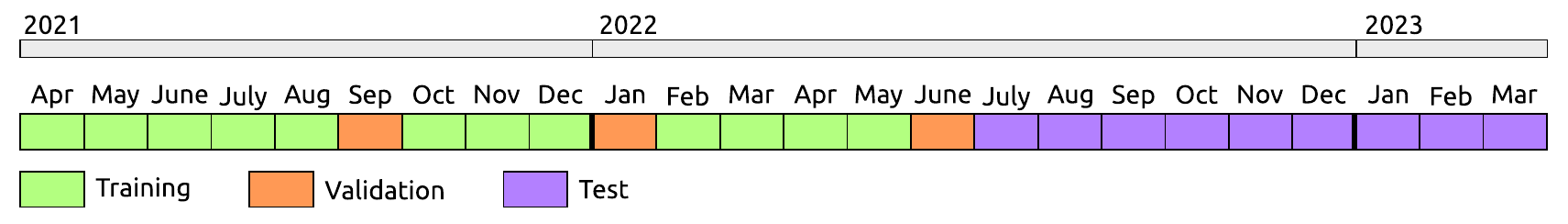}
    \caption{Overview of the period covered in the dataset and the training/validation/test split.}
    \label{fig:dataset_split}
\end{figure}

\paragraph{Dataset} 
The \gls{MEPS} dataset consists of archived forecasts from the operational \gls{MEPS} system during the period April~2021 -- March~2023.
This period was chosen due to the system configuration being reasonably stable, preventing distributional shifts within the dataset.
From the chosen period we extract the forecasts started at 00 and 12 UTC each day.
At each initialization time there are 5 ensemble forecasts, started from slightly different initial conditions.
This results in 10 forecasts per day (assuming no ensemble members are missing due to operational issues).
When retrieving the data we additionally downsample the spatial resolution from the original \qty{2.5}{\kilo\metre} to \qty{10}{\kilo\metre}.
This results in a dataset of 6069 forecasts of length \qty{66}{\hour} with \qty{1}{\hour} time steps.
We split the forecasts into training, validation and test sets according to \cref{fig:dataset_split}.
The specific validation months were chosen to reasonably cover the seasonal variations.
Note that the dataset contains around 46 years of individual time steps.
However, since there are obvious correlations between ensemble members and successively started forecasts the actual information content is far lower than in a 46 year reanalysis dataset.

\paragraph{Variables and forcing} 
\begin{table}[tbp]
\centering
\caption{Variables, forcing and static fields in the \gls{MEPS} dataset. \textsuperscript{\textdagger}In the \gls{MEPS} system 65 vertical model levels are defined from the ground to the top of the atmosphere \cite{arome_metcoop}. The lowest \gls{MEPS} level (Lvl65) sits at approximately 12.5 \si{\meter} above ground.}
\label{tab:lam_variables}
\begin{tabular}{@{}llcl@{}}
\toprule
& Abbreviation & Unit & Vertical Levels \\ \midrule
\tablesubsec{Variables}
    \varline{pres}{Ground level~(\texttt{0g}), Sea level~(\texttt{0s})}{\si{\pascal}}{Atmospheric pressure} 
    \varline{nlwrs}{Surface}{\si{\watt\per\metre\squared}}{Net longwave solar radiation flux} %
    \varline{nswrs}{Surface}{\si{\watt\per\metre\squared}}{Net shortwave solar radiation flux}
    \varline{r}{2 \si{\metre}, \mepsbottomlevel}{$[0,1]$}{Relative humidity}
    \varline{t}{2 \si{\metre}, \mepsbottomlevel, \preslevel{500}, \preslevel{850}}{\si{\kelvin}}{Temperature}
    \varline{u}{\mepsbottomlevel, \preslevel{850}}{\si{\meter\per\second}}{$u$-component of wind}
    \varline{v}{\mepsbottomlevel, \preslevel{850}}{\si{\meter\per\second}}{$v$-component of wind}
    \varline{wvint}{Full column}{\si{\kilo\gram\per\meter\squared}}{Water vapor, integrated column}
    \varline{z}{\preslevel{500}, \preslevel{1000}}{\si{\meter\squared\per\second\squared}}{Geopotential} \midrule
\tablesubsec{Forcing}
    \varline{toa}{Top of atmosphere}{\si{\watt\per\metre\squared}}{Top of atm. solar radiation flux}
    \varline{water}{Surface}{$[0,1]$}{Fraction of open water}
    \varline{sin_tod}{-}{-}{Sine-encoded time of day}
    \varline{cos_tod}{-}{-}{Cosine-encoded time of day}
    \varline{sin_toy}{-}{-}{Sine-encoded time of year}
    \varline{cos_toy}{-}{-}{Cosine-encoded time of year}\midrule
\tablesubsec{Static Fields}
    \varline{topography}{Surface}{\si{\meter\squared\per\second\squared}}{Surface geopotential (topography)}
    \varline{x_coord}{-}{-}{$x$-coordinate in \acrshort{MEPS} projection}
    \varline{y_coord}{-}{-}{$y$-coordinate in \acrshort{MEPS} projection}
    \varline{boundary}{-}{0/1}{On-boundary binary indicator}
\bottomrule
\end{tabular}
\end{table}
At each grid cell we model 17 weather variables, including a broad range of different quantities and different vertical levels in the atmosphere.
All variables, forcing and static fields are described in \cref{tab:lam_variables}.
The particular choice of variables was motivated by a combination of meteorological relevance, data availability and striving for a diverse set of variables to evaluate the model on. 
We use the same type of windowing for the forcing and standardization of variables as in the global experiment.
For solar radiation (\wvar{nlwrs} and \wvar{nswrs}) we consider the net flux at ground level, aggregated over the past \qty{3}{hours} (since the last time step).
Apart from the solar radiation all other variables are instantaneous.
For the \gls{MEPS} data we use the fraction of open water in the grid cell as a forcing input.
We assume this to be constant over the forecast period and take the value from the time of the initial state.
In the global experiment the land-sea mask is static, but treating this as forcing could be useful for taking into account seasonal fluctuations of the ice cover in the Nordic region.
The boundary forcing $\boundarystate^t$ consists of the same variables as listed in \cref{tab:lam_variables}.
We include a static binary indicator variable describing if a node is in the boundary or forecast area.

\paragraph{Forecast steps and length} 
The original data uses \qty{1}{\hour} time steps, but our \gls{NeurWP} models predict in \qty{3}{\hour} steps.
Because of this we can extract 3 training samples from each forecast in the dataset (i.e. original time steps $\set{1,4,7,\dots}$, $\set{2,5,8,\dots}$ and $\set{3,6,9,\dots}$).
As we train on only $\forecastlength \leq 8$ rollout steps, we also only need a subset of each such time series for each training iteration.
To make maximum use of the data during training we randomly sample which time step to use as initial state for unrolling the model.
The combination of 
\begin{inparaenum}[1)]
    \item the \qty{3}{\hour} time steps,
    \item using the last two weather states as model inputs,
    \item including forcing from multiple times as input,
\end{inparaenum}
means that we reduce the effective length of our ground truth forecasts in pre-processing.
This explains why we predict for \qty{57}{\hour} rather than the original \qty{66}{h}.
Note however that nothing prevents us from unrolling the models to create forecasts for \qty{66}{h} or beyond.
Although this is possible, we do not have ground truth data to compare against past \qty{57}{h} and therefore view such experiments to be of limited interest.
There is still no reason to expect the model performance to become drastically worse specifically past \qty{57}{h}, as all models are anyhow fine-tuned on shorter rollouts than this.

\subsection{Boundary Forcing}
\begin{figure}
    \centering
    \includegraphics[width=\linewidth]{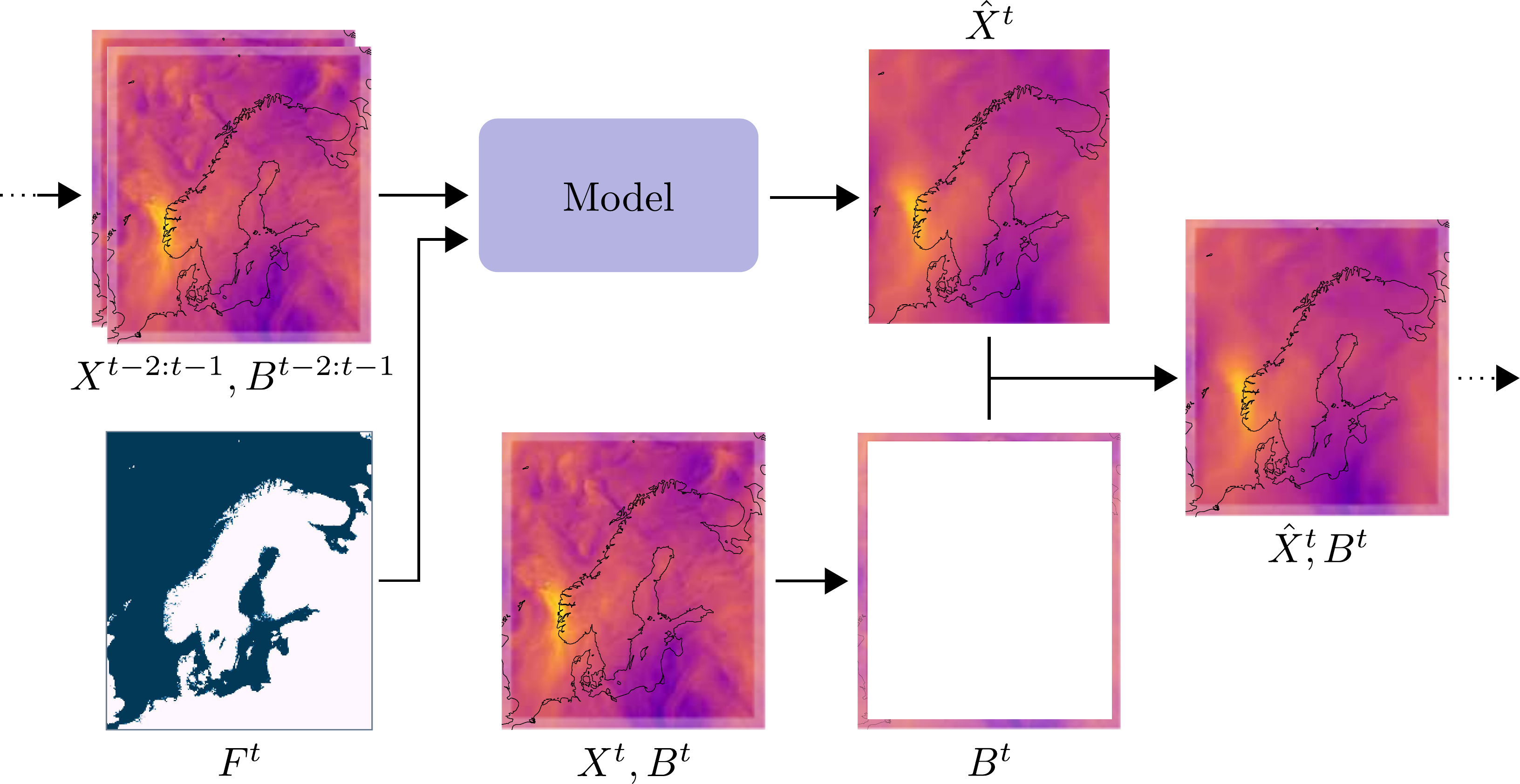}
    \caption{Schematic showing how we feed boundary forcing as input to the model in each autoregressive step.}
    \label{fig:boundary_forcing}
\end{figure}

In all models we use boundary forcing $\boundarystate^t$ in order to include information about the surrounding area.
At each time step boundary forcing $\boundarystate^{t-2:t-1}$ is passed to the model together with $\arcond$, as described in \cref{fig:boundary_forcing}.
There is a separate set of grid nodes used for the boundary input.
These nodes are treated identically as grid nodes within the forecasting area by the \gls{GNN} layers.
As the boundary forcing is only a change to the model inputs this does not require any substantial re-design and we can adapt all models in our experiments to this \gls{LAM} setting.
Note that we use a boundary area that lays \emph{inside} the original \gls{MEPS} area, allowing us to use parts of the ground truth forecasts as boundary forcing.
We specifically define the boundary as the 10 grid cells closest to the edge of the limited area.
In the operational system the boundary area is defined along the edge \emph{outside} the \gls{MEPS} area.
There is no major conceptual difference between these and one could easily re-define the different areas to match the operational \gls{MEPS} system.

\subsection{Model and Training Configurations}
For the \gls{MEPS} experiment we again use the AdamW optimizer \citeapp{adamw}, but not mixed precision computations.
Hyperparameter tuning follows a similar strategy as for the global experiment. 
The exact model configurations and training times for our \gls{MEPS} experiments are listed in \cref{tab:lam_model_details}. 
The training schedule for the deterministic models is given in \cref{tab:lam_det_training} and for the probabilistic models in \cref{tab:lam_prob_training}.
These tables follow the same formats as \cref{tab:global_model_details,tab:global_det_training,tab:global_prob_training}, and we refer to the global experiment details in \cref{sec:global_details_model_conf} for further explanations.
In the \gls{MEPS} case one epoch means training on one sub-sample (see description of sub-sampling in \cref{sec:lam_dataset_details}) of each forecast in the training set once.

\begin{table}[tbp]
\centering
\caption{Details of model architectures and training times for \gls{LAM} forecasting.}
\label{tab:lam_model_details}
\begin{tabular}{@{}lcccc@{}}
\toprule
Model & \multicolumn{1}{c}{$\latentdim$} & \multicolumn{1}{c}{Processing steps} & \multicolumn{1}{c}{Parameters} & Training time (GPU-hours) \\ \midrule
\gcour      & 128 & 6 & $1.1 \times 10^6$ & \num{336} \ \\ 
\himodeldet & 128 & 6 & $4.5\times 10^6$ & \num{432} \\ 
\msmodelprob & 128 & $\meanfunc{\latent}$: 2, $\predictorfunc$: 4, $q$: 2 & $1.7 \times 10^6$ & \num{456} \\ 
\himodelprob & 128 & - & $3.1 \times 10^6$ & \num{556} \\ 
\bottomrule
\end{tabular}
\end{table}
\begin{table}[tbp]
\centering
\caption{Training schedule for deterministic models (\gcour and \himodeldet) on \gls{MEPS} data.}
\label{tab:lam_det_training}
\begin{tabular}{@{}rcc@{}}
\toprule
Epochs & \multicolumn{1}{c}{Learning Rate} & \multicolumn{1}{c}{Unrolling $\forecastlength$} \\ \midrule
600    & $10^{-3}$ & 1      \\ 
300    & $10^{-4}$ & 4      \\ 
200    & $10^{-4}$ & 8      \\ \bottomrule
\end{tabular}
\end{table}
\begin{table}[tbp]
\centering
\caption{Training schedule for \himodelprob  on \gls{MEPS} data. For \msmodelprob we use the same schedule but with different constants ($\klweight = 100$, $\crpsweight = 10^4$, fine-tuning learning rate $10^{-4}$).}
\label{tab:lam_prob_training}
\begin{tabular}{@{}rcccc@{}}
\toprule
Epochs & \multicolumn{1}{c}{Learning Rate} & \multicolumn{1}{c}{Unrolling $\forecastlength$} & $\klweight$ & $\crpsweight$ \\ \midrule
300    & $10^{-3}$ & 1 & 0 & 0 \\ 
100    & $10^{-3}$ & 1 & 1 & 0 \\ 
200    & $5 \times 10^{-4}$ & 4 & 1 & 0 \\ 
50     & $5 \times 10^{-4}$ & 4 & 1 & $10^6$ \\ 
100    & $5 \times 10^{-4}$ & 6 & 1 & $10^6$ \\ \bottomrule
\end{tabular}
\end{table}
\section{Additional Results: \globalsection}
\label{sec:global_extra_res}
In this appendix we present additional results for global forecasting.

\subsection{Metrics}
\label{sec:global_metrics_appendix}
Comparisons between existing \gls{NeurWP} models are challenging, due to the many factors that impact forecast quality.
While most global models proposed in the literature are trained on ERA5, the exact choice of train/test split, spatial resolution and forecasted variables can vary (compare for example \citet{keisler}, \citet{fourcastnet} and \citet{swinvrnn}).
Models that operate on a higher spatial resolution and with more variables get more information in initial states, and should therefore be expected to produce better forecasts.
When it comes to \gls{NeurWP} ensemble forecasting, an even greater challenge is that different initial conditions are used \cite{gencast,uq_for_data_models,calibration_of_large_neurwp}.
Given this situation, it is hard to disentangle the proposed machine learning methods from their surrounding design choices and make fair comparisons of model architectures.

We approach these complications by re-training a smaller set of models on the same data and experimental setup.
Our focus is on graph-based, non-hybrid \gls{NeurWP} models, as this is the regime of \himodeldet and \himodelprob.
The baseline models are described in detail in \cref{sec:baseline_model_details}.
In this appendix we also include additional metrics taken directly from the WeatherBench 2 Benchmark \cite{weatherbench2}.
These are for the original \textbf{\gc} model \cite{graphcast}, \textbf{KeislerNet} \cite{keisler} and the \textbf{IFS-ENS} operational ensemble from the \gls{ECMWF} \cite{ifs_ens}.
The evaluation setup for these results match ours (evaluation against 2020 from ERA5, identical metrics computed on 1.5\textdegree{} data).
However, as discussed above there are differences in the exact training data, variables and operating resolutions compared to our models.

Metric values from our global experiments are presented in \crefrange{fig:global_metrics_rmse_extra}{fig:global_metrics_spskr_extra}.
We here showcase values for surface variables and atmospheric variables at pressure levels 500, 700 and 850 \si{\hecto\pascal}.
The results taken directly from WeatherBench 2 are shown in gray, to emphasize that any comparison with these comes with caveats.
Note that results for all variables are not available for all models.

\newcommand{\globalextramsec}[4]{%
\begin{figure}[tbp]%
    \centering
    \legendsubfig{#1}{#2}{#4}\\
    \varsubfig{#1}{#2}{2t}{2t}%
    \varsubfig{#1}{#2}{10u}{10u}%
    \varsubfig{#1}{#2}{10v}{10v}
    \varsubfig{#1}{#2}{msl}{msl}%
    \varsubfig{#1}{#2}{tp}{tp}%
    \varsubfig{#1}{#2}{z500}{z500}
    \varsubfig{#1}{#2}{t500}{t500}%
    \varsubfig{#1}{#2}{q500}{q500}%
    \varsubfig{#1}{#2}{u500}{u500}
    \varsubfig{#1}{#2}{v500}{v500}%
    \varsubfig{#1}{#2}{w500}{w500}%
    \varsubfig{#1}{#2}{z700}{z700}
    \varsubfig{#1}{#2}{t700}{t700}%
    \varsubfig{#1}{#2}{q700}{q700}%
    \varsubfig{#1}{#2}{u700}{u700}
\end{figure}
\begin{figure}[tbp]\ContinuedFloat%
    {\centering
    \legendsubfig{#1}{#2}{#4}\\}
    \varsubfig{#1}{#2}{v700}{v700}%
    \varsubfig{#1}{#2}{w700}{w700}%
    \varsubfig{#1}{#2}{z850}{z850}
    \varsubfig{#1}{#2}{t850}{t850}%
    \varsubfig{#1}{#2}{q850}{q850}%
    \varsubfig{#1}{#2}{u850}{u850}
    \varsubfig{#1}{#2}{v850}{v850}%
    \varsubfig{#1}{#2}{w850}{w850}%
    \caption{#3}
    \label{fig:#1_#2_extra}
\end{figure}
}

\globalextramsec{global_metrics}{rmse}{\gls{RMSE} results for global experiment.}{1}
\globalextramsec{global_metrics}{crps}{\gls{CRPS} results for global experiment.}{1}
\globalextramsec{global_metrics}{spskr}{\gls{SPSKR} results for global experiment.}{0.66}

\subsection{Example Forecasts}
\label{sec:global_forecasts_appendix}
In \cref{fig:global_baseline_fc} we showcase example forecasts from different models at lead time 10 days.
Note that forecasting this far into the future is challenging, and sampled ensemble members show only one possible scenario predicted by the models.

\Cref{fig:global_example_fc_extra} shows example ensemble forecasts from \himodelprob for surface variables, atmospheric variables at \qty{700}{\hecto\pascal}, \wvar{z500} and \wvar{t850}.
All forecasts are for 10 days lead time.
For practical reasons we use only 20 members to estimate the ensemble mean and standard deviation in these plots.

\newcommand{\globalblfig}[3]{%
    \begin{subfigure}[t]{#3}%
        \centering
        \includegraphics[width=\textwidth]{graphics/baseline_forecasts/global/gt/#1_2020-01-01T00_240h.pdf}%
        \captionsetup{justification=centering}
        \caption{\wvar{\detokenize{#2}}\\Ground Truth}
    \end{subfigure}
    \begin{subfigure}[t]{#3}%
        \centering
        \includegraphics[width=\textwidth]{graphics/baseline_forecasts/global/gc/#1_2020-01-01T00_240h.pdf}%
        \captionsetup{justification=centering}
        \caption{\wvar{\detokenize{#2}}\\\gcour}
    \end{subfigure}
    \begin{subfigure}[t]{#3}%
        \centering
        \includegraphics[width=\textwidth]{graphics/baseline_forecasts/global/fm/#1_2020-01-01T00_240h.pdf}%
        \captionsetup{justification=centering}
        \caption{\wvar{\detokenize{#2}}\\\himodeldet}
    \end{subfigure}
    \begin{subfigure}[t]{#3}%
        \centering
        \includegraphics[width=\textwidth]{graphics/baseline_forecasts/global/efm_ms/memb0_#1_2020-01-01T00_240h.pdf}%
        \captionsetup{justification=centering}
        \caption{\wvar{\detokenize{#2}}\\\msmodelprob, Ens. Member}
    \end{subfigure}
    \begin{subfigure}[t]{#3}%
        \centering
        \includegraphics[width=\textwidth]{graphics/baseline_forecasts/global/efm/memb0_#1_2020-01-01T00_240h.pdf}%
        \captionsetup{justification=centering}
        \caption{\wvar{\detokenize{#2}}\\\himodelprob, Ens. Member}
    \end{subfigure}
}
\begin{figure}[tbp]%
    \centering
    \globalblfig{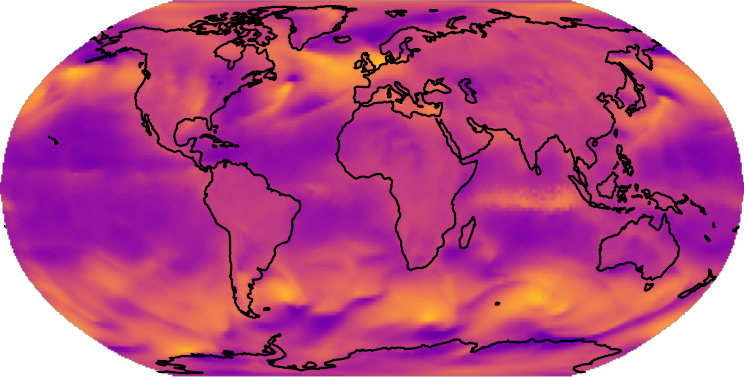}{10u}{0.32\textwidth}
    \\\vspace{1.5em}
    \globalblfig{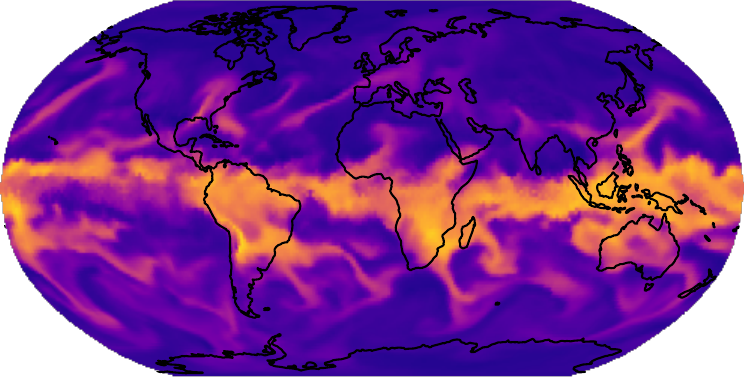}{q700}{0.32\textwidth}
    \\\vspace{1.5em}
    \globalblfig{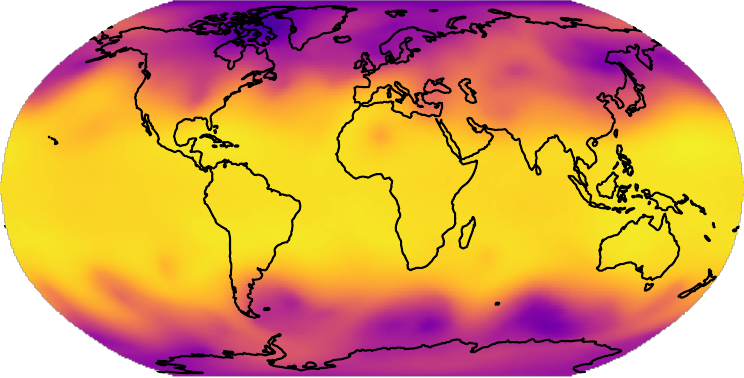}{z500}{0.32\textwidth}
    \caption{
    Comparison of global model forecasts at 10 days lead time for u-component of 10 m wind (\wvar{10u}), specific humidity at \qty{700}{\hecto\pascal} (\wvar{q700}) and  geopotential at \qty{500}{\hecto\pascal} (\wvar{z500}). 
    For probabilistic models we show sampled ensemble members.
    }
    \label{fig:global_baseline_fc}
\end{figure}
\newcommand{\globalfcfig}[2]{
    \begin{subfigure}[b]{\textwidth}%
        \centering
        \includegraphics[width=\textwidth]{graphics/example_forecasts/global/prob/hi/\detokenize{#1}_240h.pdf}%
        \caption{\wvar{\detokenize{#2}}}
    \end{subfigure}
    \\\vspace{4em}
}
\begin{figure}[tbp]%
    \centering
    \globalfcfig{2m_temperature}{2t}
    \globalfcfig{10m_u_component_of_wind}{10u}
    \globalfcfig{10m_v_component_of_wind}{10v}
\end{figure}
\begin{figure}[tbp]\ContinuedFloat%
    \centering
    \globalfcfig{mean_sea_level_pressure}{msl}
    \globalfcfig{total_precipitation_6hr}{tp}
    \globalfcfig{geopotential_500}{z500}
\end{figure}
\begin{figure}[tbp]\ContinuedFloat%
    \centering
    \globalfcfig{temperature_850}{t850}
    \globalfcfig{specific_humidity_700}{q700}
    \globalfcfig{geopotential_700}{z700}
\end{figure}
\begin{figure}[tbp]\ContinuedFloat%
    \centering
    \globalfcfig{temperature_700}{t700}
    \globalfcfig{u_component_of_wind_700}{u700}
    \globalfcfig{v_component_of_wind_700}{v700}
\end{figure}
\begin{figure}[tbp]\ContinuedFloat%
    \centering
    \globalfcfig{vertical_velocity_700}{w700}
    \caption{Example \himodelprob global ensemble forecasts at lead time 10 days.}
    \label{fig:global_example_fc_extra}
\end{figure}

\FloatBarrier
\section{Additional Results: \lamsection}
\label{sec:lam_extra_res}
In this appendix we show additional results from our experiment with the \gls{MEPS} data.

\subsection{Metrics}
\label{sec:lam_metrics_appendix}

\Crefrange{fig:lam_metrics_rmse_extra}{fig:lam_metrics_spskr_extra} show metric values for all variables and lead times in the \gls{MEPS} dataset.
\newcommand{\lamextramsecone}[4]{%
\begin{figure}[tbp]%
    \centering
    \legendsubfig{#1}{#2}{#4}\\
    \varsubfig{#1}{#2}{pres_0g}{pres0g}%
    \varsubfig{#1}{#2}{pres_0s}{pres0s}%
    \varsubfig{#1}{#2}{nlwrs_0}{nlwrs}
    \varsubfig{#1}{#2}{nswrs_0}{nswrs}%
    \varsubfig{#1}{#2}{r_2}{2r}%
    \varsubfig{#1}{#2}{r_65}{r65}
\end{figure}
\begin{figure}[tbp]\ContinuedFloat%
    {\centering
    \legendsubfig{#1}{#2}{#4}\\}
    \varsubfig{#1}{#2}{t_2}{2t}%
    \varsubfig{#1}{#2}{t_65}{t65}%
    \varsubfig{#1}{#2}{t_500}{t500}
    \varsubfig{#1}{#2}{t_850}{t850}%
    \varsubfig{#1}{#2}{u_65}{u65}%
    \varsubfig{#1}{#2}{v_65}{v65}
    \varsubfig{#1}{#2}{u_850}{u850}%
    \varsubfig{#1}{#2}{v_850}{v850}%
    \varsubfig{#1}{#2}{wvint_0}{wvint}
    \varsubfig{#1}{#2}{z_500}{z500}%
    \varsubfig{#1}{#2}{z_1000}{z1000}%
    \caption{#3}
    \label{fig:#1_#2_extra}
\end{figure}
}
\newcommand{\lamextramsectwo}[4]{%
\begin{figure}[tbp]%
    \centering
    \legendsubfig{#1}{#2}{#4}\\
    \varsubfig{#1}{#2}{pres_0g}{pres0g}%
    \varsubfig{#1}{#2}{pres_0s}{pres0s}%
    \varsubfig{#1}{#2}{nlwrs_0}{nlwrs}
    \varsubfig{#1}{#2}{nswrs_0}{nswrs}%
    \varsubfig{#1}{#2}{r_2}{2r}%
    \varsubfig{#1}{#2}{r_65}{r65}
    \varsubfig{#1}{#2}{t_2}{2t}%
    \varsubfig{#1}{#2}{t_65}{t65}%
    \varsubfig{#1}{#2}{t_500}{t500}
    \varsubfig{#1}{#2}{t_850}{t850}%
    \varsubfig{#1}{#2}{u_65}{u65}%
    \varsubfig{#1}{#2}{v_65}{v65}
    \varsubfig{#1}{#2}{u_850}{u850}%
    \varsubfig{#1}{#2}{v_850}{v850}%
    \varsubfig{#1}{#2}{wvint_0}{wvint}
\end{figure}
\begin{figure}[tbp]\ContinuedFloat%
    {\centering
    \legendsubfig{#1}{#2}{#4}\\}
    \varsubfig{#1}{#2}{z_500}{z500}%
    \varsubfig{#1}{#2}{z_1000}{z1000}%
    \caption{#3}
    \label{fig:#1_#2_extra}
\end{figure}
}

\lamextramsecone{lam_metrics}{rmse}{\gls{RMSE} for \gls{MEPS} experiment.}{0.83}
\lamextramsectwo{lam_metrics}{crps}{\gls{CRPS} for \gls{MEPS} experiment.}{0.83}
\lamextramsecone{lam_metrics}{spskr}{\gls{SPSKR} for \gls{MEPS} experiment.}{0.66} 

The poor performance of the \msmodelprob model is noteworthy, both in terms of forecast accuracy and ensemble calibration.
This can to a large extent be attributed to the exact training objective (see \cref{tab:lam_prob_training}), particularly a lower weighting $\crpsweight$.
Using a lower value for $\crpsweight$ in the multi-scale model was necessary to avoid the artifacts discussed in \cref{sec:spatial_coherency}.
The does however mean that \msmodelprob does not gain as much of the benefits that come with the \gls{CRPS} fine-tuning.

\subsection{Example Forecasts}
\label{sec:lam_forecasts_appendix}
A comparison between example forecasts from different models is given in \cref{fig:lam_baseline_fc}.
Note that ensemble members sampled from \himodelprob show more detailed features than deterministic forecasts.

In \cref{fig:lam_example_fc_extra} we plot example ensemble forecasts from the \himodelprob model for all variables in the \gls{MEPS} data.
Note that these plots include the boundary area, which is not forecast, as a faded border in each plot.
All forecasts are for lead time \qty{57}{\hour}.
Rolling out a \gls{LAM} ensemble forecast from \himodelprob to \qty{57}{\hour} is even faster than generating the global ensemble.
Using batched sampling on a single GPU, we can produce 100 ensemble members in \qty{140}{\second} (\qty{1.4}{\second} per member).

\newcommand{\lamblfig}[3]{%
    \begin{subfigure}[t]{#3}%
        \centering
        \includegraphics[width=\textwidth]{graphics/baseline_forecasts/lam/gt/#1_57h.pdf}%
        \captionsetup{justification=centering}
        \caption{\wvar{\detokenize{#2}}\\Ground Truth}
    \end{subfigure}
    \begin{subfigure}[t]{#3}%
        \centering
        \includegraphics[width=\textwidth]{graphics/baseline_forecasts/lam/gc/#1_57h.pdf}%
        \captionsetup{justification=centering}
        \caption{\wvar{\detokenize{#2}}\\\gcour}
    \end{subfigure}
    \begin{subfigure}[t]{#3}%
        \centering
        \includegraphics[width=\textwidth]{graphics/baseline_forecasts/lam/fm/#1_57h.pdf}%
        \captionsetup{justification=centering}
        \caption{\wvar{\detokenize{#2}}\\\himodeldet}
    \end{subfigure}
    \begin{subfigure}[t]{#3}%
        \centering
        \includegraphics[width=\textwidth]{graphics/baseline_forecasts/lam/efm_ms/memb0_#1_57h.pdf}%
        \captionsetup{justification=centering}
        \caption{\wvar{\detokenize{#2}}\\\msmodelprob\\Ens. Member}
    \end{subfigure}
    \begin{subfigure}[t]{#3}%
        \centering
        \includegraphics[width=\textwidth]{graphics/baseline_forecasts/lam/efm/memb0_#1_57h.pdf}%
        \captionsetup{justification=centering}
        \caption{\wvar{\detokenize{#2}}\\\himodelprob\\Ens. Member}
    \end{subfigure}
}
\begin{figure}[tbp]%
    \centering
    \lamblfig{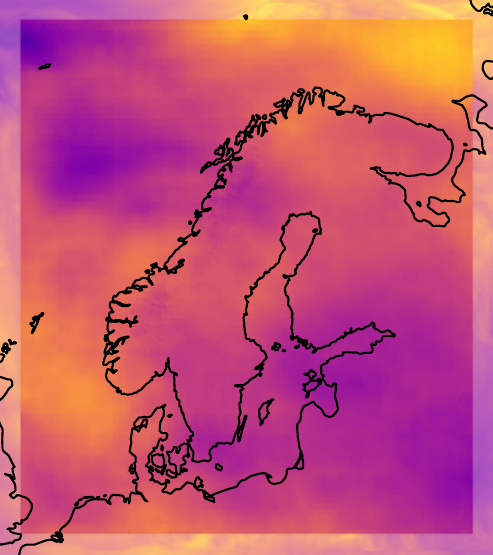}{u850}{0.27\textwidth}
    \\\vspace{1em}
    \lamblfig{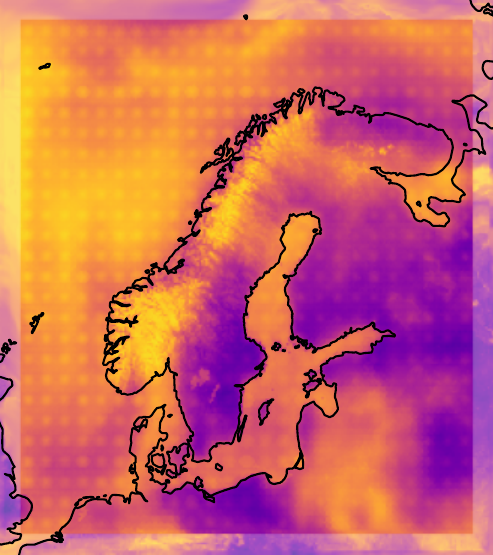}{2r}{0.27\textwidth}
    \caption{
    Comparison of \gls{LAM} forecasts for u-component of wind at \qty{850}{\hecto\pascal} (\wvar{u850}) and 2 m relative humidity (\wvar{2r}) at 57 h lead time. 
    For probabilistic models we show sampled ensemble members.
    }
    \label{fig:lam_baseline_fc}
\end{figure}
\newcommand{\lamfcfig}[2]{
    \begin{subfigure}[b]{\textwidth}%
        \centering
        \includegraphics[width=\textwidth]{graphics/example_forecasts/lam/prob/hi/\detokenize{#1}_57h.pdf}%
        \caption{\wvar{\detokenize{#2}}}
    \end{subfigure}
    \\\vspace{2em}
}
\begin{figure}[tbp]%
    \centering
    \lamfcfig{pres_0g}{pres0g}
    \lamfcfig{pres_0s}{pres0s}
\end{figure}
\begin{figure}[tbp]\ContinuedFloat%
    \centering
    \lamfcfig{nlwrs_0}{nlwrs}
    \lamfcfig{nswrs_0}{nswrs}
\end{figure}
\begin{figure}[tbp]\ContinuedFloat%
    \centering
    \lamfcfig{r_2}{2r}
    \lamfcfig{r_65}{r65}
\end{figure}
\begin{figure}[tbp]\ContinuedFloat%
    \centering
    \lamfcfig{t_2}{2t}
    \lamfcfig{t_65}{t65}
\end{figure}
\begin{figure}[tbp]\ContinuedFloat%
    \centering
    \lamfcfig{t_500}{t500}
    \lamfcfig{t_850}{t850}
\end{figure}
\begin{figure}[tbp]\ContinuedFloat%
    \centering
    \lamfcfig{u_65}{u65}
    \lamfcfig{u_850}{u850}
\end{figure}
\begin{figure}[tbp]\ContinuedFloat%
    \centering
    \lamfcfig{v_65}{v65}
    \lamfcfig{v_850}{v850}
\end{figure}
\begin{figure}[tbp]\ContinuedFloat%
    \centering
    \lamfcfig{wvint_0}{wvint}
    \lamfcfig{z_500}{z500}
\end{figure}
\begin{figure}[tbp]\ContinuedFloat%
    \centering
    \lamfcfig{z_1000}{z1000}
    \caption{Example \himodelprob \gls{LAM} ensemble forecasts at lead time \qty{57}{\hour}.}
    \label{fig:lam_example_fc_extra}
\end{figure}

\FloatBarrier
\section{Additional experiments}

\subsection{Comparing Interaction and Propagation Networks in \himodeldet}
\label{sec:propnet_exp}
Given the usefulness of Propagation Networks in \himodelprob it is reasonable to ask if these could be beneficial to use also in the deterministic \himodeldet model.
Also \himodeldet uses a hierarchical mesh graph, so the propagation of information between levels is important also for this model. 
We test this empirically, by training versions of \himodeldet using Interaction and Propagation networks.
As the retention of information is also important for some parts of the architecture, we do not replace all \glspl{GNN} with Propagation Networks (see \cref{sec:hi_det_details}).
We compare models on both global and \gls{LAM} forecasting, using the same experimental setups as described in \cref{sec:global_details,sec:lam_details}.

\paragraph{\globalsectionlower}
\Cref{fig:global_propnet_exp} shows \glspl{RMSE} for highlighted variables from ERA5 and \himodeldet using Interaction and Propagation Networks.
The models have very similar errors for almost all variables.
However, for variables in the upper atmosphere the Propagation Network model shows lower errors.

\begin{figure}[tbp]
    \centering
    \legendsubfig{propnet_global_exp}{rmse}{0.50}
    \varsubfig{propnet_global_exp}{rmse}{z500}{z500}%
    \varsubfig{propnet_global_exp}{rmse}{q700}{q700}%
    \varsubfig{propnet_global_exp}{rmse}{t850}{t850}
    \varsubfig{propnet_global_exp}{rmse}{2t}{2t}%
    \varsubfig{propnet_global_exp}{rmse}{t100}{t100}%
    \varsubfig{propnet_global_exp}{rmse}{w100}{w100}%
    \caption{\gls{RMSE} of \himodeldet models with Propagation and Interaction Networks, evaluated on the ERA5 test set}
    \label{fig:global_propnet_exp}
\end{figure}

\paragraph{\lamsectionlower}
In \cref{fig:lam_propnet_exp} we compare \glspl{RMSE} for \himodeldet models using the different \gls{GNN} layers on \gls{MEPS} data.
Here we see a greater advantage of the Propagation Networks. 
We hypothesize that this relates to the boundary forcing.
Using Propagation Networks the information from boundary nodes should easier reach the top graph level, where it can faster spread throughout the forecasting area.
Motivated by these results we use Propagation Networks in our final \himodeldet architecture, both for global and \gls{LAM} forecasting.

\begin{figure}[tbp]
    \centering
    \legendsubfig{propnet_lam_exp}{rmse}{0.50}
    \varsubfig{propnet_lam_exp}{rmse}{z_500}{z500}%
    \varsubfig{propnet_lam_exp}{rmse}{wvint_0}{wvint}%
    \varsubfig{propnet_lam_exp}{rmse}{nlwrs_0}{nlwrs}
    \varsubfig{propnet_lam_exp}{rmse}{t_2}{2t}%
    \varsubfig{propnet_lam_exp}{rmse}{u_65}{u65}%
    \varsubfig{propnet_lam_exp}{rmse}{v_65}{v65}
    \caption{\gls{RMSE} of \himodeldet models with Propagation and Interaction Networks, evaluated on the \gls{MEPS} test set.}
    \label{fig:lam_propnet_exp}
\end{figure}

\subsection{Importance of Latent Map}
\label{sec:latent_map_exp}
In \himodelprob we define the latent map $\priordist$ using a neural network with explicit dependence on $\arcond$.
Given that also the predictor takes $\arcond$ as inputs, it is not immediately clear that conditioning on this also in the latent map is necessary.
Indeed, due to the ability of deep neural networks to well approximate arbitrary functions, the predictor network should internally be able to transform a simple $\latent^\timei \sim \normal{0}{I}$ to introduce the dependence on $\arcond$.
This does however assume infinite flexibility in the predictor, which might be far from the situation in practice.

To investigate the importance of a learnable latent map we compare our \himodelprob model on the \gls{MEPS} dataset with one where $\latent^\timei$ is sampled from a static distribution $\normal{0}{I}$.
This experiment was carried out using an earlier version of \himodelprob, with the same architecture but a slightly different training schedule.
To save on computations we here only sample 16 ensemble members from each model.
\gls{RMSE}, \gls{CRPS} and \gls{SPSKR} are shown in \cref{fig:prior_exp_rmse,fig:prior_exp_crps,fig:prior_exp_spskr}.
In terms of \gls{RMSE} and \gls{CRPS} there is a clear benefit to letting the distribution over $\latent^\timei$ be a learnable mapping. 
The \gls{SPSKR} in \cref{fig:prior_exp_spskr} shows no clear trends between the models.
For practical network architectures the latent map does add flexibility, changing the mean of $\latent^\timei$ that enters the predictor.
The learnable latent map should also simplify the inference problem solved when optimizing our variational objective.
With the dependence on previous states we can expect a smaller discrepancy between $\priordist$ and $\postdist$, simplifying the optimization of the variational approximation $\vardist$.
We believe that this is an important aspect of the observed empirical benefit of a using the latent map.

\newcommand{\lmexpplot}[3]{\begin{figure}[tbp]
    \centering
    \legendsubfig{learn_prior_experiment}{#1}{#3}
    \varsubfig{learn_prior_experiment}{#1}{t_2}{2t}%
    \varsubfig{learn_prior_experiment}{#1}{wvint_0}{wvint}%
    \varsubfig{learn_prior_experiment}{#1}{u_850}{u850}%
    \caption{\gls{#2} for \himodelprob models with a static distribution for $\latent^\timei$ or a learnable latent map, evaluted on the \gls{MEPS} validation dataset.}
    \label{fig:prior_exp_#1}
\end{figure}}
\lmexpplot{rmse}{RMSE}{0.33}
\lmexpplot{crps}{CRPS}{0.33}
\lmexpplot{spskr}{SPSKR}{0.5}

\subsection{Impact of Ensemble Size}
\label{sec:ens_size_experiment}

We here study how the performance of the model, in terms of metric values, varies when sampling different numbers of ensemble members. To investigate this we ran the evaluation of \himodelprob with 5--80 members for the global model and 5--100 for the \gls{LAM} model.
Results for a selection of variables are shown in \cref{fig:ensemble_size_plots_global} (global, ERA5) and \cref{fig:ensemble_size_plots_lam} (\gls{LAM}, \gls{MEPS}). 
As expected the \gls{RMSE} of the ensemble mean decreases when sampling more members. 
However, already when sampling 20 or 25 members the results are fairly close to the full ensemble. For \gls{SPSKR} the differences are even smaller. 
As the \gls{CRPS} is a property of the distribution of the model forecast, its true value does not depend on the number of samples drawn. In practice we compute \gls{CRPS} using an unbiased estimator, and the variance of this estimator decreases with ensemble size. When averaged spatially and over the whole test set we do however not see any difference in \gls{CRPS} for different ensemble sizes. All these trends hold consistently for all variables in both the ERA5 and \gls{MEPS} datasets.

In our main experiments in \cref{sec:experiments} we use the full 80/100 member ensembles.
Given that any improvements to metrics saturate, we would not expect results to meaningfully change from sampling even more members than this.
It should however be noted that the motivation for sampling very large ensembles is mainly not to improve on metrics such as these. 
More important motivations for large ensembles include estimating probabilities of rare events or studying different possible scenarios of extreme weather. 

\newcommand{\enssizesf}[3]{%
\begin{subfigure}[b]{0.33\textwidth}%
        \includegraphics[width=\textwidth]{graphics/ens_size/#1}
        \caption{\wvar{#3}, \gls{#2}}
\end{subfigure}%
}

\begin{figure}[tbp]
    \centering
    \begin{subfigure}[b]{\textwidth}%
        \includegraphics[width=\textwidth]{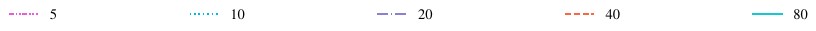}
    \end{subfigure}
    \enssizesf{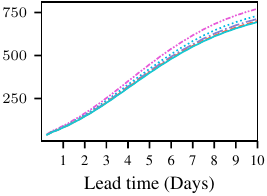}{RMSE}{z500}%
    \enssizesf{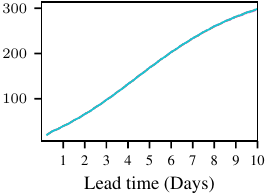}{CRPS}{z500}%
    \enssizesf{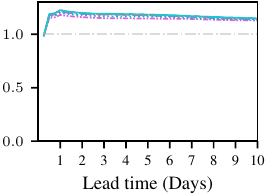}{SPSKR}{z500}
    \enssizesf{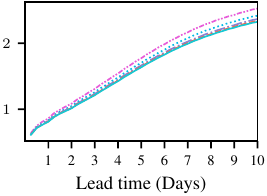}{RMSE}{2t}%
    \enssizesf{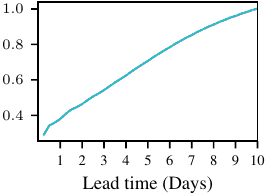}{CRPS}{2t}%
    \enssizesf{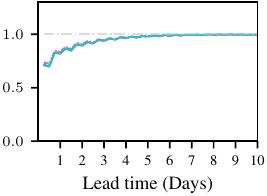}{SPSKR}{2t}%
    \caption{
    Metric values for \wvar{z500} and \wvar{2t} from the global experiment when sampling different numbers of ensemble members from \himodelprob.
    }
    \label{fig:ensemble_size_plots_global}
\end{figure}

\begin{figure}[tbp]
    \centering
    \begin{subfigure}[b]{\textwidth}%
        \includegraphics[width=\textwidth]{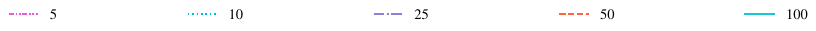}
    \end{subfigure}
    \enssizesf{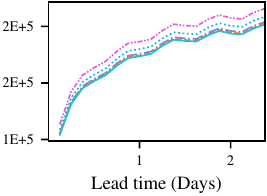}{RMSE}{nlwrs}%
    \enssizesf{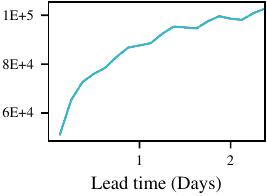}{CRPS}{nlwrs}%
    \enssizesf{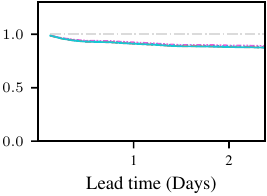}{SPSKR}{nlwrs}
    \enssizesf{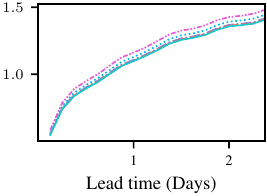}{RMSE}{2t}%
    \enssizesf{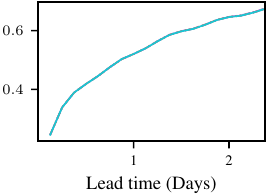}{CRPS}{2t}%
    \enssizesf{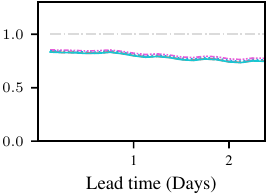}{SPSKR}{2t}%
    \caption{
    Metric values for \wvar{nlwrs} and \wvar{2t} from the \gls{LAM} experiment when sampling different numbers of ensemble members from \himodelprob.
    }
    \label{fig:ensemble_size_plots_lam}
\end{figure}
\FloatBarrier
\bibliographyapp{references}
\bibliographystyleapp{abbrvnat}

\ifnotpreprint
\newpage
    \section*{NeurIPS Paper Checklist}

\begin{enumerate}

\item {\bf Claims}
    \item[] Question: Do the main claims made in the abstract and introduction accurately reflect the paper's contributions and scope?
    \item[] Answer: \answerYes{} %
    \item[] Justification: We outline the scope and contributions of our work in the introduction. See specifically the last paragraph where we list our main contributions.

\item {\bf Limitations}
    \item[] Question: Does the paper discuss the limitations of the work performed by the authors?
    \item[] Answer: \answerYes{}
    \item[] Justification: Limitations are discussed in \cref{sec:discussion}.

\item {\bf Theory Assumptions and Proofs}
    \item[] Question: For each theoretical result, does the paper provide the full set of assumptions and a complete (and correct) proof?
    \item[] Answer: \answerNA{} %
    \item[] Justification: The paper does not present theoretical results.

\item {\bf Experimental Result Reproducibility}
    \item[] Question: Does the paper fully disclose all the information needed to reproduce the main experimental results of the paper to the extent that it affects the main claims and/or conclusions of the paper (regardless of whether the code and data are provided or not)?
    \item[] Answer: \answerYes{}
    \item[] Justification: We provide enough details about the model architecture and experimental setups to reproduce our results.
    Apart from the descriptions in the main paper, model details are given in \cref{sec:model_details} and experimental details in \cref{sec:lam_details,sec:global_details}.
    Our code is also openly available.

\item {\bf Open access to data and code}
    \item[] Question: Does the paper provide open access to the data and code, with sufficient instructions to faithfully reproduce the main experimental results, as described in supplemental material?
    \item[] Answer: \answerYes{} %
    \item[] Justification: Our code is openly available, together with instructions on how to run it. Along with details given in the appendices this allows for reproducing our results.
    The ERA5 dataset is openly available.
    The \gls{MEPS} data is available to other researchers upon request.

\item {\bf Experimental Setting/Details}
    \item[] Question: Does the paper specify all the training and test details (e.g., data splits, hyperparameters, how they were chosen, type of optimizer, etc.) necessary to understand the results?
    \item[] Answer: \answerYes{}
    \item[] Justification: See \cref{sec:lam_details,sec:global_details}.

\item {\bf Experiment Statistical Significance}
    \item[] Question: Does the paper report error bars suitably and correctly defined or other appropriate information about the statistical significance of the experiments?
    \item[] Answer: \answerNo{}
    \item[] Justification: The main sources of randomness in our experiments relate to the random seed used for parameter initialization and sampling.
    A proper statistical analysis would require training and evaluating multiple models.
    However, the computational cost associated makes it unfeasible to train enough models to draw any well founded conclusions.

\item {\bf Experiments Compute Resources}
    \item[] Question: For each experiment, does the paper provide sufficient information on the computer resources (type of compute workers, memory, time of execution) needed to reproduce the experiments?
    \item[] Answer: \answerYes{}
    \item[] Justification: See \cref{sec:experiments}, \cref{tab:global_model_details,tab:lam_model_details}.
    
\item {\bf Code Of Ethics}
    \item[] Question: Does the research conducted in the paper conform, in every respect, with the NeurIPS Code of Ethics \url{https://neurips.cc/public/EthicsGuidelines}?
    \item[] Answer: \answerYes{}
    \item[] Justification: 

\item {\bf Broader Impacts}
    \item[] Question: Does the paper discuss both potential positive societal impacts and negative societal impacts of the work performed?
    \item[] Answer: \answerYes{} %
    \item[] Justification: See \cref{sec:introduction} and \cref{sec:societal_impact}.
    
\item {\bf Safeguards}
    \item[] Question: Does the paper describe safeguards that have been put in place for responsible release of data or models that have a high risk for misuse (e.g., pretrained language models, image generators, or scraped datasets)?
    \item[] Answer: \answerNA{}
    \item[] Justification: We do not judge the shared models to have a high risk of misuse.

\item {\bf Licenses for existing assets}
    \item[] Question: Are the creators or original owners of assets (e.g., code, data, models), used in the paper, properly credited and are the license and terms of use explicitly mentioned and properly respected?
    \item[] Answer: \answerYes{}
    \item[] Justification: We explicitly refer to the ERA5 license. The MEPS dataset is used with permission from the the Swedish Meteorological and Hydrological Institute.

\item {\bf New Assets}
    \item[] Question: Are new assets introduced in the paper well documented and is the documentation provided alongside the assets?
    \item[] Answer: \answerYes{}
    \item[] Justification: We are releasing our code and models with the paper. The code comes with documentation to the degree that other researchers can reproduce and extend our work.

\item {\bf Crowdsourcing and Research with Human Subjects}
    \item[] Question: For crowdsourcing experiments and research with human subjects, does the paper include the full text of instructions given to participants and screenshots, if applicable, as well as details about compensation (if any)? 
    \item[] Answer: \answerNA{} 
    \item[] Justification: 

\item {\bf Institutional Review Board (IRB) Approvals or Equivalent for Research with Human Subjects}
    \item[] Question: Does the paper describe potential risks incurred by study participants, whether such risks were disclosed to the subjects, and whether Institutional Review Board (IRB) approvals (or an equivalent approval/review based on the requirements of your country or institution) were obtained?
    \item[] Answer: \answerNA{}
    \item[] Justification: 

\end{enumerate}
\fi
\end{document}